\documentclass{article}

     \usepackage[preprint]{neurips_2020}

\usepackage[utf8]{inputenc} % allow utf-8 input
\usepackage[T1]{fontenc}    % use 8-bit T1 fonts
\usepackage{hyperref}       % hyperlinks
\usepackage{url}            % simple URL typesetting
\usepackage{booktabs}       % professional-quality tables
\usepackage{amsfonts}       % blackboard math symbols
\usepackage{nicefrac}       % compact symbols for 1/2, etc.
\usepackage{microtype}      % microtypography
\usepackage{subfig}
\usepackage{natbib}
\usepackage{graphicx}
\usepackage{color}
\usepackage{amsfonts}
\usepackage{amsmath}
\usepackage{wrapfig}
\usepackage{amssymb}

\usepackage{multicol}
\usepackage{pgfplots}
\usepackage{filecontents}
\usepackage[linesnumbered,ruled,vlined]{algorithm2e}
\usepackage{algorithmic,algorithm2e,float}

\usepackage{pgfplotstable}

\newcommand\pgfplotscreatecolorbrewercyclelist{	\pgfmathtruncatemacro\maxindex{\pgfkeysvalueof{/pgfplots/colorbrewer values}-1}
	\pgfplotstabletypeset[
		begin table={},
		end table={},
		skip coltypes,
		string type,
		col sep=semicolon,
		write to macro=\listmacro,
		skip rows between index={\maxindex}{99},
		typeset=false,
		row sep=crcr
	]{
	colorbrewer1, thick, mark=*\\
	colorbrewer2, thick, mark=square*, mark size=1.75\\
	colorbrewer3, thick, mark=triangle*, mark size=2.5\\
	colorbrewer4, thick, mark=x, mark size=3, mark options={line width=2\pgflinewidth}\\
	colorbrewer5, thick, mark=*, mark options={fill=white}\\
	colorbrewer6, thick, mark=square*, mark size=1.75, mark options={fill=white}\\
	colorbrewer7, thick, mark=triangle*, mark size=2.5, mark options={fill=white}\\
	colorbrewer8, thick, mark=+, mark size=3, mark options={line width=2\pgflinewidth}\\
	}
	\begingroup
    \edef\A{\noexpand\global\noexpand\pgfplotscreateplotcyclelist{colorbrewer}{\unexpanded\expandafter{\listmacro}}}%
    \expandafter
    \endgroup\A
}

\pgfplotsset{
	colorbrewer values/.initial=8,
	colorbrewer scheme/.code={
		\pgfplotsset{colorbrewer scheme/#1}
		\pgfplotsset{colorbrewer values/\pgfkeysvalueof{/pgfplots/colorbrewer values}}
	},
	colorbrewer cycle list/.style={
		colorbrewer create cycle list,
		cycle list name=colorbrewer,
		/pgfplots/execute at begin axis={
			\pgfplotsset{
				colorbrewer scheme/#1,
				colorbrewer values/\pgfkeysvalueof{/pgfplots/colorbrewer values}
			}
		}
	},
	colorbrewer cycle list/.default=Dark2,
	colorbrewer create cycle list/.code={
		\pgfplotscreatecolorbrewercyclelist
	}
}

\pgfplotsset{
	colorbrewer scheme/.is choice, 
	colorbrewer scheme/Accent/.style={
		colorbrewer values/.is choice,
		colorbrewer values/3/.code={
			\definecolor{colorbrewer1}{RGB}{127,201,127}
			\definecolor{colorbrewer2}{RGB}{190,174,212}
			\definecolor{colorbrewer3}{RGB}{253,192,134}
		},
		colorbrewer values=3,
		colorbrewer values/4/.code={
			\definecolor{colorbrewer1}{RGB}{127,201,127}
			\definecolor{colorbrewer2}{RGB}{190,174,212}
			\definecolor{colorbrewer3}{RGB}{253,192,134}
			\definecolor{colorbrewer4}{RGB}{255,255,153}
		},
		colorbrewer values/5/.code={
			\definecolor{colorbrewer1}{RGB}{127,201,127}
			\definecolor{colorbrewer2}{RGB}{190,174,212}
			\definecolor{colorbrewer3}{RGB}{253,192,134}
			\definecolor{colorbrewer4}{RGB}{255,255,153}
			\definecolor{colorbrewer5}{RGB}{56,108,176}
		},
		colorbrewer values/6/.code={
			\definecolor{colorbrewer1}{RGB}{127,201,127}
			\definecolor{colorbrewer2}{RGB}{190,174,212}
			\definecolor{colorbrewer3}{RGB}{253,192,134}
			\definecolor{colorbrewer4}{RGB}{255,255,153}
			\definecolor{colorbrewer5}{RGB}{56,108,176}
			\definecolor{colorbrewer6}{RGB}{240,2,127}
		},
		colorbrewer values/7/.code={
			\definecolor{colorbrewer1}{RGB}{127,201,127}
			\definecolor{colorbrewer2}{RGB}{190,174,212}
			\definecolor{colorbrewer3}{RGB}{253,192,134}
			\definecolor{colorbrewer4}{RGB}{255,255,153}
			\definecolor{colorbrewer5}{RGB}{56,108,176}
			\definecolor{colorbrewer6}{RGB}{240,2,127}
			\definecolor{colorbrewer7}{RGB}{191,91,23}
		},
		colorbrewer values/8/.code={
			\definecolor{colorbrewer1}{RGB}{127,201,127}
			\definecolor{colorbrewer2}{RGB}{190,174,212}
			\definecolor{colorbrewer3}{RGB}{253,192,134}
			\definecolor{colorbrewer4}{RGB}{255,255,153}
			\definecolor{colorbrewer5}{RGB}{56,108,176}
			\definecolor{colorbrewer6}{RGB}{240,2,127}
			\definecolor{colorbrewer7}{RGB}{191,91,23}
			\definecolor{colorbrewer8}{RGB}{102,102,102}
		},
		colorbrewer values/.unknown/.style={/errors/unknown choice value=\pgfkeyscurrentkey\pgfkeyscurrentvalue}
	},
	colorbrewer scheme/Blues/.style={
		colorbrewer values/.is choice,	
		colorbrewer values/3/.code={
			\definecolor{colorbrewer1}{RGB}{222,235,247}
			\definecolor{colorbrewer2}{RGB}{158,202,225}
			\definecolor{colorbrewer3}{RGB}{49,130,189}
		},
		colorbrewer values=3,
		colorbrewer values/4/.code={
			\definecolor{colorbrewer1}{RGB}{239,243,255}
			\definecolor{colorbrewer2}{RGB}{189,215,231}
			\definecolor{colorbrewer3}{RGB}{107,174,214}
			\definecolor{colorbrewer4}{RGB}{33,113,181}
		},
		colorbrewer values/5/.code={
			\definecolor{colorbrewer1}{RGB}{239,243,255}
			\definecolor{colorbrewer2}{RGB}{189,215,231}
			\definecolor{colorbrewer3}{RGB}{107,174,214}
			\definecolor{colorbrewer4}{RGB}{49,130,189}
			\definecolor{colorbrewer5}{RGB}{8,81,156}
		},
		colorbrewer values/6/.code={
			\definecolor{colorbrewer1}{RGB}{239,243,255}
			\definecolor{colorbrewer2}{RGB}{198,219,239}
			\definecolor{colorbrewer3}{RGB}{158,202,225}
			\definecolor{colorbrewer4}{RGB}{107,174,214}
			\definecolor{colorbrewer5}{RGB}{49,130,189}
			\definecolor{colorbrewer6}{RGB}{8,81,156}
		},
		colorbrewer values/7/.code={
			\definecolor{colorbrewer1}{RGB}{239,243,255}
			\definecolor{colorbrewer2}{RGB}{198,219,239}
			\definecolor{colorbrewer3}{RGB}{158,202,225}
			\definecolor{colorbrewer4}{RGB}{107,174,214}
			\definecolor{colorbrewer5}{RGB}{66,146,198}
			\definecolor{colorbrewer6}{RGB}{33,113,181}
			\definecolor{colorbrewer7}{RGB}{8,69,148}
		},
		colorbrewer values/8/.code={
			\definecolor{colorbrewer1}{RGB}{247,251,255}
			\definecolor{colorbrewer2}{RGB}{222,235,247}
			\definecolor{colorbrewer3}{RGB}{198,219,239}
			\definecolor{colorbrewer4}{RGB}{158,202,225}
			\definecolor{colorbrewer5}{RGB}{107,174,214}
			\definecolor{colorbrewer6}{RGB}{66,146,198}
			\definecolor{colorbrewer7}{RGB}{33,113,181}
			\definecolor{colorbrewer8}{RGB}{8,69,148}
		},
		colorbrewer values/9/.code={
			\definecolor{colorbrewer1}{RGB}{247,251,255}
			\definecolor{colorbrewer2}{RGB}{222,235,247}
			\definecolor{colorbrewer3}{RGB}{198,219,239}
			\definecolor{colorbrewer4}{RGB}{158,202,225}
			\definecolor{colorbrewer5}{RGB}{107,174,214}
			\definecolor{colorbrewer6}{RGB}{66,146,198}
			\definecolor{colorbrewer7}{RGB}{33,113,181}
			\definecolor{colorbrewer8}{RGB}{8,81,156}
			\definecolor{colorbrewer9}{RGB}{8,48,107}
		}
	},
	colorbrewer scheme/BrBG/.style={
		colorbrewer values/.is choice,	
		colorbrewer values/3/.code={
			\definecolor{colorbrewer1}{RGB}{216,179,101}
			\definecolor{colorbrewer2}{RGB}{245,245,245}
			\definecolor{colorbrewer3}{RGB}{90,180,172}
		},
		colorbrewer values=3,
		colorbrewer values/4/.code={
			\definecolor{colorbrewer1}{RGB}{166,97,26}
			\definecolor{colorbrewer2}{RGB}{223,194,125}
			\definecolor{colorbrewer3}{RGB}{128,205,193}
			\definecolor{colorbrewer4}{RGB}{1,133,113}
		},
		colorbrewer values/5/.code={
			\definecolor{colorbrewer1}{RGB}{166,97,26}
			\definecolor{colorbrewer2}{RGB}{223,194,125}
			\definecolor{colorbrewer3}{RGB}{245,245,245}
			\definecolor{colorbrewer4}{RGB}{128,205,193}
			\definecolor{colorbrewer5}{RGB}{1,133,113}
		},
		colorbrewer values/6/.code={
			\definecolor{colorbrewer1}{RGB}{140,81,10}
			\definecolor{colorbrewer2}{RGB}{216,179,101}
			\definecolor{colorbrewer3}{RGB}{246,232,195}
			\definecolor{colorbrewer4}{RGB}{199,234,229}
			\definecolor{colorbrewer5}{RGB}{90,180,172}
			\definecolor{colorbrewer6}{RGB}{1,102,94}
		},
		colorbrewer values/7/.code={
			\definecolor{colorbrewer1}{RGB}{140,81,10}
			\definecolor{colorbrewer2}{RGB}{216,179,101}
			\definecolor{colorbrewer3}{RGB}{246,232,195}
			\definecolor{colorbrewer4}{RGB}{245,245,245}
			\definecolor{colorbrewer5}{RGB}{199,234,229}
			\definecolor{colorbrewer6}{RGB}{90,180,172}
			\definecolor{colorbrewer7}{RGB}{1,102,94}
		},
		colorbrewer values/8/.code={
			\definecolor{colorbrewer1}{RGB}{140,81,10}
			\definecolor{colorbrewer2}{RGB}{191,129,45}
			\definecolor{colorbrewer3}{RGB}{223,194,125}
			\definecolor{colorbrewer4}{RGB}{246,232,195}
			\definecolor{colorbrewer5}{RGB}{199,234,229}
			\definecolor{colorbrewer6}{RGB}{128,205,193}
			\definecolor{colorbrewer7}{RGB}{53,151,143}
			\definecolor{colorbrewer8}{RGB}{1,102,94}
		},
		colorbrewer values/9/.code={
			\definecolor{colorbrewer1}{RGB}{140,81,10}
			\definecolor{colorbrewer2}{RGB}{191,129,45}
			\definecolor{colorbrewer3}{RGB}{223,194,125}
			\definecolor{colorbrewer4}{RGB}{246,232,195}
			\definecolor{colorbrewer5}{RGB}{245,245,245}
			\definecolor{colorbrewer6}{RGB}{199,234,229}
			\definecolor{colorbrewer7}{RGB}{128,205,193}
			\definecolor{colorbrewer8}{RGB}{53,151,143}
			\definecolor{colorbrewer9}{RGB}{1,102,94}
		},
		colorbrewer values/10/.code={
			\definecolor{colorbrewer1}{RGB}{84,48,5}
			\definecolor{colorbrewer2}{RGB}{140,81,10}
			\definecolor{colorbrewer3}{RGB}{191,129,45}
			\definecolor{colorbrewer4}{RGB}{223,194,125}
			\definecolor{colorbrewer5}{RGB}{246,232,195}
			\definecolor{colorbrewer6}{RGB}{199,234,229}
			\definecolor{colorbrewer7}{RGB}{128,205,193}
			\definecolor{colorbrewer8}{RGB}{53,151,143}
			\definecolor{colorbrewer9}{RGB}{1,102,94}
			\definecolor{colorbrewer10}{RGB}{0,60,48}
		},
		colorbrewer values/11/.code={
			\definecolor{colorbrewer1}{RGB}{84,48,5}
			\definecolor{colorbrewer2}{RGB}{140,81,10}
			\definecolor{colorbrewer3}{RGB}{191,129,45}
			\definecolor{colorbrewer4}{RGB}{223,194,125}
			\definecolor{colorbrewer5}{RGB}{246,232,195}
			\definecolor{colorbrewer6}{RGB}{245,245,245}
			\definecolor{colorbrewer7}{RGB}{199,234,229}
			\definecolor{colorbrewer8}{RGB}{128,205,193}
			\definecolor{colorbrewer9}{RGB}{53,151,143}
			\definecolor{colorbrewer10}{RGB}{1,102,94}
			\definecolor{colorbrewer11}{RGB}{0,60,48}
		}
	},
	colorbrewer scheme/BuGn/.style={
		colorbrewer values/.is choice,	
		colorbrewer values/3/.code={
			\definecolor{colorbrewer1}{RGB}{229,245,249}
			\definecolor{colorbrewer2}{RGB}{153,216,201}
			\definecolor{colorbrewer3}{RGB}{44,162,95}
		},
		colorbrewer values=3,
		colorbrewer values/4/.code={
			\definecolor{colorbrewer1}{RGB}{237,248,251}
			\definecolor{colorbrewer2}{RGB}{178,226,226}
			\definecolor{colorbrewer3}{RGB}{102,194,164}
			\definecolor{colorbrewer4}{RGB}{35,139,69}
		},
		colorbrewer values/5/.code={
			\definecolor{colorbrewer1}{RGB}{237,248,251}
			\definecolor{colorbrewer2}{RGB}{178,226,226}
			\definecolor{colorbrewer3}{RGB}{102,194,164}
			\definecolor{colorbrewer4}{RGB}{44,162,95}
			\definecolor{colorbrewer5}{RGB}{0,109,44}
		},
		colorbrewer values/6/.code={
			\definecolor{colorbrewer1}{RGB}{237,248,251}
			\definecolor{colorbrewer2}{RGB}{204,236,230}
			\definecolor{colorbrewer3}{RGB}{153,216,201}
			\definecolor{colorbrewer4}{RGB}{102,194,164}
			\definecolor{colorbrewer5}{RGB}{44,162,95}
			\definecolor{colorbrewer6}{RGB}{0,109,44}
		},
		colorbrewer values/7/.code={
			\definecolor{colorbrewer1}{RGB}{237,248,251}
			\definecolor{colorbrewer2}{RGB}{204,236,230}
			\definecolor{colorbrewer3}{RGB}{153,216,201}
			\definecolor{colorbrewer4}{RGB}{102,194,164}
			\definecolor{colorbrewer5}{RGB}{65,174,118}
			\definecolor{colorbrewer6}{RGB}{35,139,69}
			\definecolor{colorbrewer7}{RGB}{0,88,36}
		},
		colorbrewer values/8/.code={
			\definecolor{colorbrewer1}{RGB}{247,252,253}
			\definecolor{colorbrewer2}{RGB}{229,245,249}
			\definecolor{colorbrewer3}{RGB}{204,236,230}
			\definecolor{colorbrewer4}{RGB}{153,216,201}
			\definecolor{colorbrewer5}{RGB}{102,194,164}
			\definecolor{colorbrewer6}{RGB}{65,174,118}
			\definecolor{colorbrewer7}{RGB}{35,139,69}
			\definecolor{colorbrewer8}{RGB}{0,88,36}
		},
		colorbrewer values/9/.code={
			\definecolor{colorbrewer1}{RGB}{247,252,253}
			\definecolor{colorbrewer2}{RGB}{229,245,249}
			\definecolor{colorbrewer3}{RGB}{204,236,230}
			\definecolor{colorbrewer4}{RGB}{153,216,201}
			\definecolor{colorbrewer5}{RGB}{102,194,164}
			\definecolor{colorbrewer6}{RGB}{65,174,118}
			\definecolor{colorbrewer7}{RGB}{35,139,69}
			\definecolor{colorbrewer8}{RGB}{0,109,44}
			\definecolor{colorbrewer9}{RGB}{0,68,27}
		}
	},
	colorbrewer scheme/BuPu/.style={
		colorbrewer values/.is choice,	
		colorbrewer values/3/.code={
			\definecolor{colorbrewer1}{RGB}{224,236,244}
			\definecolor{colorbrewer2}{RGB}{158,188,218}
			\definecolor{colorbrewer3}{RGB}{136,86,167}
		},
		colorbrewer values=3,
		colorbrewer values/4/.code={
			\definecolor{colorbrewer1}{RGB}{237,248,251}
			\definecolor{colorbrewer2}{RGB}{179,205,227}
			\definecolor{colorbrewer3}{RGB}{140,150,198}
			\definecolor{colorbrewer4}{RGB}{136,65,157}
		},
		colorbrewer values/5/.code={
			\definecolor{colorbrewer1}{RGB}{237,248,251}
			\definecolor{colorbrewer2}{RGB}{179,205,227}
			\definecolor{colorbrewer3}{RGB}{140,150,198}
			\definecolor{colorbrewer4}{RGB}{136,86,167}
			\definecolor{colorbrewer5}{RGB}{129,15,124}
		},
		colorbrewer values/6/.code={
			\definecolor{colorbrewer1}{RGB}{237,248,251}
			\definecolor{colorbrewer2}{RGB}{191,211,230}
			\definecolor{colorbrewer3}{RGB}{158,188,218}
			\definecolor{colorbrewer4}{RGB}{140,150,198}
			\definecolor{colorbrewer5}{RGB}{136,86,167}
			\definecolor{colorbrewer6}{RGB}{129,15,124}
		},
		colorbrewer values/7/.code={
			\definecolor{colorbrewer1}{RGB}{237,248,251}
			\definecolor{colorbrewer2}{RGB}{191,211,230}
			\definecolor{colorbrewer3}{RGB}{158,188,218}
			\definecolor{colorbrewer4}{RGB}{140,150,198}
			\definecolor{colorbrewer5}{RGB}{140,107,177}
			\definecolor{colorbrewer6}{RGB}{136,65,157}
			\definecolor{colorbrewer7}{RGB}{110,1,107}
		},
		colorbrewer values/8/.code={
			\definecolor{colorbrewer1}{RGB}{247,252,253}
			\definecolor{colorbrewer2}{RGB}{224,236,244}
			\definecolor{colorbrewer3}{RGB}{191,211,230}
			\definecolor{colorbrewer4}{RGB}{158,188,218}
			\definecolor{colorbrewer5}{RGB}{140,150,198}
			\definecolor{colorbrewer6}{RGB}{140,107,177}
			\definecolor{colorbrewer7}{RGB}{136,65,157}
			\definecolor{colorbrewer8}{RGB}{110,1,107}
		},
		colorbrewer values/9/.code={
			\definecolor{colorbrewer1}{RGB}{247,252,253}
			\definecolor{colorbrewer2}{RGB}{224,236,244}
			\definecolor{colorbrewer3}{RGB}{191,211,230}
			\definecolor{colorbrewer4}{RGB}{158,188,218}
			\definecolor{colorbrewer5}{RGB}{140,150,198}
			\definecolor{colorbrewer6}{RGB}{140,107,177}
			\definecolor{colorbrewer7}{RGB}{136,65,157}
			\definecolor{colorbrewer8}{RGB}{129,15,124}
			\definecolor{colorbrewer9}{RGB}{77,0,75}
		}
	},
	colorbrewer scheme/Dark2/.style={
		colorbrewer values/.is choice,	
		colorbrewer values/3/.code={
			\definecolor{colorbrewer1}{RGB}{27,158,119}
			\definecolor{colorbrewer2}{RGB}{217,95,2}
			\definecolor{colorbrewer3}{RGB}{117,112,179}
		},
		colorbrewer values=3,
		colorbrewer values/4/.code={
			\definecolor{colorbrewer1}{RGB}{27,158,119}
			\definecolor{colorbrewer2}{RGB}{217,95,2}
			\definecolor{colorbrewer3}{RGB}{117,112,179}
			\definecolor{colorbrewer4}{RGB}{231,41,138}
		},
		colorbrewer values/5/.code={
			\definecolor{colorbrewer1}{RGB}{27,158,119}
			\definecolor{colorbrewer2}{RGB}{217,95,2}
			\definecolor{colorbrewer3}{RGB}{117,112,179}
			\definecolor{colorbrewer4}{RGB}{231,41,138}
			\definecolor{colorbrewer5}{RGB}{102,166,30}
		},
		colorbrewer values/6/.code={
			\definecolor{colorbrewer1}{RGB}{27,158,119}
			\definecolor{colorbrewer2}{RGB}{217,95,2}
			\definecolor{colorbrewer3}{RGB}{117,112,179}
			\definecolor{colorbrewer4}{RGB}{231,41,138}
			\definecolor{colorbrewer5}{RGB}{102,166,30}
			\definecolor{colorbrewer6}{RGB}{230,171,2}
		},
		colorbrewer values/7/.code={
			\definecolor{colorbrewer1}{RGB}{27,158,119}
			\definecolor{colorbrewer2}{RGB}{217,95,2}
			\definecolor{colorbrewer3}{RGB}{117,112,179}
			\definecolor{colorbrewer4}{RGB}{231,41,138}
			\definecolor{colorbrewer5}{RGB}{102,166,30}
			\definecolor{colorbrewer6}{RGB}{230,171,2}
			\definecolor{colorbrewer7}{RGB}{166,118,29}
		},
		colorbrewer values/8/.code={
			\definecolor{colorbrewer1}{RGB}{27,158,119}
			\definecolor{colorbrewer2}{RGB}{217,95,2}
			\definecolor{colorbrewer3}{RGB}{117,112,179}
			\definecolor{colorbrewer4}{RGB}{231,41,138}
			\definecolor{colorbrewer5}{RGB}{102,166,30}
			\definecolor{colorbrewer6}{RGB}{230,171,2}
			\definecolor{colorbrewer7}{RGB}{166,118,29}
			\definecolor{colorbrewer8}{RGB}{102,102,102}
		}
	},
	colorbrewer scheme/GnBu/.style={
		colorbrewer values/.is choice,	
		colorbrewer values/3/.code={
			\definecolor{colorbrewer1}{RGB}{224,243,219}
			\definecolor{colorbrewer2}{RGB}{168,221,181}
			\definecolor{colorbrewer3}{RGB}{67,162,202}
		},
		colorbrewer values=3,
		colorbrewer values/4/.code={
			\definecolor{colorbrewer1}{RGB}{240,249,232}
			\definecolor{colorbrewer2}{RGB}{186,228,188}
			\definecolor{colorbrewer3}{RGB}{123,204,196}
			\definecolor{colorbrewer4}{RGB}{43,140,190}
		},
		colorbrewer values/5/.code={
			\definecolor{colorbrewer1}{RGB}{240,249,232}
			\definecolor{colorbrewer2}{RGB}{186,228,188}
			\definecolor{colorbrewer3}{RGB}{123,204,196}
			\definecolor{colorbrewer4}{RGB}{67,162,202}
			\definecolor{colorbrewer5}{RGB}{8,104,172}
		},
		colorbrewer values/6/.code={
			\definecolor{colorbrewer1}{RGB}{240,249,232}
			\definecolor{colorbrewer2}{RGB}{204,235,197}
			\definecolor{colorbrewer3}{RGB}{168,221,181}
			\definecolor{colorbrewer4}{RGB}{123,204,196}
			\definecolor{colorbrewer5}{RGB}{67,162,202}
			\definecolor{colorbrewer6}{RGB}{8,104,172}
		},
		colorbrewer values/7/.code={
			\definecolor{colorbrewer1}{RGB}{240,249,232}
			\definecolor{colorbrewer2}{RGB}{204,235,197}
			\definecolor{colorbrewer3}{RGB}{168,221,181}
			\definecolor{colorbrewer4}{RGB}{123,204,196}
			\definecolor{colorbrewer5}{RGB}{78,179,211}
			\definecolor{colorbrewer6}{RGB}{43,140,190}
			\definecolor{colorbrewer7}{RGB}{8,88,158}
		},
		colorbrewer values/8/.code={
			\definecolor{colorbrewer1}{RGB}{247,252,240}
			\definecolor{colorbrewer2}{RGB}{224,243,219}
			\definecolor{colorbrewer3}{RGB}{204,235,197}
			\definecolor{colorbrewer4}{RGB}{168,221,181}
			\definecolor{colorbrewer5}{RGB}{123,204,196}
			\definecolor{colorbrewer6}{RGB}{78,179,211}
			\definecolor{colorbrewer7}{RGB}{43,140,190}
			\definecolor{colorbrewer8}{RGB}{8,88,158}
		},
		colorbrewer values/9/.code={
			\definecolor{colorbrewer1}{RGB}{247,252,240}
			\definecolor{colorbrewer2}{RGB}{224,243,219}
			\definecolor{colorbrewer3}{RGB}{204,235,197}
			\definecolor{colorbrewer4}{RGB}{168,221,181}
			\definecolor{colorbrewer5}{RGB}{123,204,196}
			\definecolor{colorbrewer6}{RGB}{78,179,211}
			\definecolor{colorbrewer7}{RGB}{43,140,190}
			\definecolor{colorbrewer8}{RGB}{8,104,172}
			\definecolor{colorbrewer9}{RGB}{8,64,129}
		}
	},
	colorbrewer scheme/Greens/.style={
		colorbrewer values/.is choice,	
		colorbrewer values/3/.code={
			\definecolor{colorbrewer1}{RGB}{229,245,224}
			\definecolor{colorbrewer2}{RGB}{161,217,155}
			\definecolor{colorbrewer3}{RGB}{49,163,84}
		},
		colorbrewer values=3,
		colorbrewer values/4/.code={
			\definecolor{colorbrewer1}{RGB}{237,248,233}
			\definecolor{colorbrewer2}{RGB}{186,228,179}
			\definecolor{colorbrewer3}{RGB}{116,196,118}
			\definecolor{colorbrewer4}{RGB}{35,139,69}
		},
		colorbrewer values/5/.code={
			\definecolor{colorbrewer1}{RGB}{237,248,233}
			\definecolor{colorbrewer2}{RGB}{186,228,179}
			\definecolor{colorbrewer3}{RGB}{116,196,118}
			\definecolor{colorbrewer4}{RGB}{49,163,84}
			\definecolor{colorbrewer5}{RGB}{0,109,44}
		},
		colorbrewer values/6/.code={
			\definecolor{colorbrewer1}{RGB}{237,248,233}
			\definecolor{colorbrewer2}{RGB}{199,233,192}
			\definecolor{colorbrewer3}{RGB}{161,217,155}
			\definecolor{colorbrewer4}{RGB}{116,196,118}
			\definecolor{colorbrewer5}{RGB}{49,163,84}
			\definecolor{colorbrewer6}{RGB}{0,109,44}
		},
		colorbrewer values/7/.code={
			\definecolor{colorbrewer1}{RGB}{237,248,233}
			\definecolor{colorbrewer2}{RGB}{199,233,192}
			\definecolor{colorbrewer3}{RGB}{161,217,155}
			\definecolor{colorbrewer4}{RGB}{116,196,118}
			\definecolor{colorbrewer5}{RGB}{65,171,93}
			\definecolor{colorbrewer6}{RGB}{35,139,69}
			\definecolor{colorbrewer7}{RGB}{0,90,50}
		},
		colorbrewer values/8/.code={
			\definecolor{colorbrewer1}{RGB}{247,252,245}
			\definecolor{colorbrewer2}{RGB}{229,245,224}
			\definecolor{colorbrewer3}{RGB}{199,233,192}
			\definecolor{colorbrewer4}{RGB}{161,217,155}
			\definecolor{colorbrewer5}{RGB}{116,196,118}
			\definecolor{colorbrewer6}{RGB}{65,171,93}
			\definecolor{colorbrewer7}{RGB}{35,139,69}
			\definecolor{colorbrewer8}{RGB}{0,90,50}
		},
		colorbrewer values/9/.code={
			\definecolor{colorbrewer1}{RGB}{247,252,245}
			\definecolor{colorbrewer2}{RGB}{229,245,224}
			\definecolor{colorbrewer3}{RGB}{199,233,192}
			\definecolor{colorbrewer4}{RGB}{161,217,155}
			\definecolor{colorbrewer5}{RGB}{116,196,118}
			\definecolor{colorbrewer6}{RGB}{65,171,93}
			\definecolor{colorbrewer7}{RGB}{35,139,69}
			\definecolor{colorbrewer8}{RGB}{0,109,44}
			\definecolor{colorbrewer9}{RGB}{0,68,27}
		}
	},
	colorbrewer scheme/Greys/.style={
		colorbrewer values/.is choice,	
		colorbrewer values/3/.code={
			\definecolor{colorbrewer1}{RGB}{240,240,240}
			\definecolor{colorbrewer2}{RGB}{189,189,189}
			\definecolor{colorbrewer3}{RGB}{99,99,99}
		},
		colorbrewer values=3,
		colorbrewer values/4/.code={
			\definecolor{colorbrewer1}{RGB}{247,247,247}
			\definecolor{colorbrewer2}{RGB}{204,204,204}
			\definecolor{colorbrewer3}{RGB}{150,150,150}
			\definecolor{colorbrewer4}{RGB}{82,82,82}
		},
		colorbrewer values/5/.code={
			\definecolor{colorbrewer1}{RGB}{247,247,247}
			\definecolor{colorbrewer2}{RGB}{204,204,204}
			\definecolor{colorbrewer3}{RGB}{150,150,150}
			\definecolor{colorbrewer4}{RGB}{99,99,99}
			\definecolor{colorbrewer5}{RGB}{37,37,37}
		},
		colorbrewer values/6/.code={
			\definecolor{colorbrewer1}{RGB}{247,247,247}
			\definecolor{colorbrewer2}{RGB}{217,217,217}
			\definecolor{colorbrewer3}{RGB}{189,189,189}
			\definecolor{colorbrewer4}{RGB}{150,150,150}
			\definecolor{colorbrewer5}{RGB}{99,99,99}
			\definecolor{colorbrewer6}{RGB}{37,37,37}
		},
		colorbrewer values/7/.code={
			\definecolor{colorbrewer1}{RGB}{247,247,247}
			\definecolor{colorbrewer2}{RGB}{217,217,217}
			\definecolor{colorbrewer3}{RGB}{189,189,189}
			\definecolor{colorbrewer4}{RGB}{150,150,150}
			\definecolor{colorbrewer5}{RGB}{115,115,115}
			\definecolor{colorbrewer6}{RGB}{82,82,82}
			\definecolor{colorbrewer7}{RGB}{37,37,37}
		},
		colorbrewer values/8/.code={
			\definecolor{colorbrewer1}{RGB}{255,255,255}
			\definecolor{colorbrewer2}{RGB}{240,240,240}
			\definecolor{colorbrewer3}{RGB}{217,217,217}
			\definecolor{colorbrewer4}{RGB}{189,189,189}
			\definecolor{colorbrewer5}{RGB}{150,150,150}
			\definecolor{colorbrewer6}{RGB}{115,115,115}
			\definecolor{colorbrewer7}{RGB}{82,82,82}
			\definecolor{colorbrewer8}{RGB}{37,37,37}
		},
		colorbrewer values/9/.code={
			\definecolor{colorbrewer1}{RGB}{255,255,255}
			\definecolor{colorbrewer2}{RGB}{240,240,240}
			\definecolor{colorbrewer3}{RGB}{217,217,217}
			\definecolor{colorbrewer4}{RGB}{189,189,189}
			\definecolor{colorbrewer5}{RGB}{150,150,150}
			\definecolor{colorbrewer6}{RGB}{115,115,115}
			\definecolor{colorbrewer7}{RGB}{82,82,82}
			\definecolor{colorbrewer8}{RGB}{37,37,37}
			\definecolor{colorbrewer9}{RGB}{0,0,0}
		}
	},
	colorbrewer scheme/Oranges/.style={
		colorbrewer values/.is choice,	
		colorbrewer values/3/.code={
			\definecolor{colorbrewer1}{RGB}{254,230,206}
			\definecolor{colorbrewer2}{RGB}{253,174,107}
			\definecolor{colorbrewer3}{RGB}{230,85,13}
		},
		colorbrewer values=3,
		colorbrewer values/4/.code={
			\definecolor{colorbrewer1}{RGB}{254,237,222}
			\definecolor{colorbrewer2}{RGB}{253,190,133}
			\definecolor{colorbrewer3}{RGB}{253,141,60}
			\definecolor{colorbrewer4}{RGB}{217,71,1}
		},
		colorbrewer values/5/.code={
			\definecolor{colorbrewer1}{RGB}{254,237,222}
			\definecolor{colorbrewer2}{RGB}{253,190,133}
			\definecolor{colorbrewer3}{RGB}{253,141,60}
			\definecolor{colorbrewer4}{RGB}{230,85,13}
			\definecolor{colorbrewer5}{RGB}{166,54,3}
		},
		colorbrewer values/6/.code={
			\definecolor{colorbrewer1}{RGB}{254,237,222}
			\definecolor{colorbrewer2}{RGB}{253,208,162}
			\definecolor{colorbrewer3}{RGB}{253,174,107}
			\definecolor{colorbrewer4}{RGB}{253,141,60}
			\definecolor{colorbrewer5}{RGB}{230,85,13}
			\definecolor{colorbrewer6}{RGB}{166,54,3}
		},
		colorbrewer values/7/.code={
			\definecolor{colorbrewer1}{RGB}{254,237,222}
			\definecolor{colorbrewer2}{RGB}{253,208,162}
			\definecolor{colorbrewer3}{RGB}{253,174,107}
			\definecolor{colorbrewer4}{RGB}{253,141,60}
			\definecolor{colorbrewer5}{RGB}{241,105,19}
			\definecolor{colorbrewer6}{RGB}{217,72,1}
			\definecolor{colorbrewer7}{RGB}{140,45,4}
		},
		colorbrewer values/8/.code={
			\definecolor{colorbrewer1}{RGB}{255,245,235}
			\definecolor{colorbrewer2}{RGB}{254,230,206}
			\definecolor{colorbrewer3}{RGB}{253,208,162}
			\definecolor{colorbrewer4}{RGB}{253,174,107}
			\definecolor{colorbrewer5}{RGB}{253,141,60}
			\definecolor{colorbrewer6}{RGB}{241,105,19}
			\definecolor{colorbrewer7}{RGB}{217,72,1}
			\definecolor{colorbrewer8}{RGB}{140,45,4}
		},
		colorbrewer values/9/.code={
			\definecolor{colorbrewer1}{RGB}{255,245,235}
			\definecolor{colorbrewer2}{RGB}{254,230,206}
			\definecolor{colorbrewer3}{RGB}{253,208,162}
			\definecolor{colorbrewer4}{RGB}{253,174,107}
			\definecolor{colorbrewer5}{RGB}{253,141,60}
			\definecolor{colorbrewer6}{RGB}{241,105,19}
			\definecolor{colorbrewer7}{RGB}{217,72,1}
			\definecolor{colorbrewer8}{RGB}{166,54,3}
			\definecolor{colorbrewer9}{RGB}{127,39,4}
		}
	},
	colorbrewer scheme/OrRd/.style={
		colorbrewer values/.is choice,	
		colorbrewer values/3/.code={
			\definecolor{colorbrewer1}{RGB}{254,232,200}
			\definecolor{colorbrewer2}{RGB}{253,187,132}
			\definecolor{colorbrewer3}{RGB}{227,74,51}
		},
		colorbrewer values=3,
		colorbrewer values/4/.code={
			\definecolor{colorbrewer1}{RGB}{254,240,217}
			\definecolor{colorbrewer2}{RGB}{253,204,138}
			\definecolor{colorbrewer3}{RGB}{252,141,89}
			\definecolor{colorbrewer4}{RGB}{215,48,31}
		},
		colorbrewer values/5/.code={
			\definecolor{colorbrewer1}{RGB}{254,240,217}
			\definecolor{colorbrewer2}{RGB}{253,204,138}
			\definecolor{colorbrewer3}{RGB}{252,141,89}
			\definecolor{colorbrewer4}{RGB}{227,74,51}
			\definecolor{colorbrewer5}{RGB}{179,0,0}
		},
		colorbrewer values/6/.code={
			\definecolor{colorbrewer1}{RGB}{254,240,217}
			\definecolor{colorbrewer2}{RGB}{253,212,158}
			\definecolor{colorbrewer3}{RGB}{253,187,132}
			\definecolor{colorbrewer4}{RGB}{252,141,89}
			\definecolor{colorbrewer5}{RGB}{227,74,51}
			\definecolor{colorbrewer6}{RGB}{179,0,0}
		},
		colorbrewer values/7/.code={
			\definecolor{colorbrewer1}{RGB}{254,240,217}
			\definecolor{colorbrewer2}{RGB}{253,212,158}
			\definecolor{colorbrewer3}{RGB}{253,187,132}
			\definecolor{colorbrewer4}{RGB}{252,141,89}
			\definecolor{colorbrewer5}{RGB}{239,101,72}
			\definecolor{colorbrewer6}{RGB}{215,48,31}
			\definecolor{colorbrewer7}{RGB}{153,0,0}
		},
		colorbrewer values/8/.code={
			\definecolor{colorbrewer1}{RGB}{255,247,236}
			\definecolor{colorbrewer2}{RGB}{254,232,200}
			\definecolor{colorbrewer3}{RGB}{253,212,158}
			\definecolor{colorbrewer4}{RGB}{253,187,132}
			\definecolor{colorbrewer5}{RGB}{252,141,89}
			\definecolor{colorbrewer6}{RGB}{239,101,72}
			\definecolor{colorbrewer7}{RGB}{215,48,31}
			\definecolor{colorbrewer8}{RGB}{153,0,0}
		},
		colorbrewer values/9/.code={
			\definecolor{colorbrewer1}{RGB}{255,247,236}
			\definecolor{colorbrewer2}{RGB}{254,232,200}
			\definecolor{colorbrewer3}{RGB}{253,212,158}
			\definecolor{colorbrewer4}{RGB}{253,187,132}
			\definecolor{colorbrewer5}{RGB}{252,141,89}
			\definecolor{colorbrewer6}{RGB}{239,101,72}
			\definecolor{colorbrewer7}{RGB}{215,48,31}
			\definecolor{colorbrewer8}{RGB}{179,0,0}
			\definecolor{colorbrewer9}{RGB}{127,0,0}
		}
	},
	colorbrewer scheme/Paired/.style={
		colorbrewer values/.is choice,	
		colorbrewer values/3/.code={
			\definecolor{colorbrewer1}{RGB}{166,206,227}
			\definecolor{colorbrewer2}{RGB}{31,120,180}
			\definecolor{colorbrewer3}{RGB}{178,223,138}
		},
		colorbrewer values=3,
		colorbrewer values/4/.code={
			\definecolor{colorbrewer1}{RGB}{166,206,227}
			\definecolor{colorbrewer2}{RGB}{31,120,180}
			\definecolor{colorbrewer3}{RGB}{178,223,138}
			\definecolor{colorbrewer4}{RGB}{51,160,44}
		},
		colorbrewer values/5/.code={
			\definecolor{colorbrewer1}{RGB}{166,206,227}
			\definecolor{colorbrewer2}{RGB}{31,120,180}
			\definecolor{colorbrewer3}{RGB}{178,223,138}
			\definecolor{colorbrewer4}{RGB}{51,160,44}
			\definecolor{colorbrewer5}{RGB}{251,154,153}
		},
		colorbrewer values/6/.code={
			\definecolor{colorbrewer1}{RGB}{166,206,227}
			\definecolor{colorbrewer2}{RGB}{31,120,180}
			\definecolor{colorbrewer3}{RGB}{178,223,138}
			\definecolor{colorbrewer4}{RGB}{51,160,44}
			\definecolor{colorbrewer5}{RGB}{251,154,153}
			\definecolor{colorbrewer6}{RGB}{227,26,28}
		},
		colorbrewer values/7/.code={
			\definecolor{colorbrewer1}{RGB}{166,206,227}
			\definecolor{colorbrewer2}{RGB}{31,120,180}
			\definecolor{colorbrewer3}{RGB}{178,223,138}
			\definecolor{colorbrewer4}{RGB}{51,160,44}
			\definecolor{colorbrewer5}{RGB}{251,154,153}
			\definecolor{colorbrewer6}{RGB}{227,26,28}
			\definecolor{colorbrewer7}{RGB}{253,191,111}
		},
		colorbrewer values/8/.code={
			\definecolor{colorbrewer1}{RGB}{166,206,227}
			\definecolor{colorbrewer2}{RGB}{31,120,180}
			\definecolor{colorbrewer3}{RGB}{178,223,138}
			\definecolor{colorbrewer4}{RGB}{51,160,44}
			\definecolor{colorbrewer5}{RGB}{251,154,153}
			\definecolor{colorbrewer6}{RGB}{227,26,28}
			\definecolor{colorbrewer7}{RGB}{253,191,111}
			\definecolor{colorbrewer8}{RGB}{255,127,0}
		},
		colorbrewer values/9/.code={
			\definecolor{colorbrewer1}{RGB}{166,206,227}
			\definecolor{colorbrewer2}{RGB}{31,120,180}
			\definecolor{colorbrewer3}{RGB}{178,223,138}
			\definecolor{colorbrewer4}{RGB}{51,160,44}
			\definecolor{colorbrewer5}{RGB}{251,154,153}
			\definecolor{colorbrewer6}{RGB}{227,26,28}
			\definecolor{colorbrewer7}{RGB}{253,191,111}
			\definecolor{colorbrewer8}{RGB}{255,127,0}
			\definecolor{colorbrewer9}{RGB}{202,178,214}
		},
		colorbrewer values/10/.code={
			\definecolor{colorbrewer1}{RGB}{166,206,227}
			\definecolor{colorbrewer2}{RGB}{31,120,180}
			\definecolor{colorbrewer3}{RGB}{178,223,138}
			\definecolor{colorbrewer4}{RGB}{51,160,44}
			\definecolor{colorbrewer5}{RGB}{251,154,153}
			\definecolor{colorbrewer6}{RGB}{227,26,28}
			\definecolor{colorbrewer7}{RGB}{253,191,111}
			\definecolor{colorbrewer8}{RGB}{255,127,0}
			\definecolor{colorbrewer9}{RGB}{202,178,214}
			\definecolor{colorbrewer10}{RGB}{106,61,154}
		},
		colorbrewer values/11/.code={
			\definecolor{colorbrewer1}{RGB}{166,206,227}
			\definecolor{colorbrewer2}{RGB}{31,120,180}
			\definecolor{colorbrewer3}{RGB}{178,223,138}
			\definecolor{colorbrewer4}{RGB}{51,160,44}
			\definecolor{colorbrewer5}{RGB}{251,154,153}
			\definecolor{colorbrewer6}{RGB}{227,26,28}
			\definecolor{colorbrewer7}{RGB}{253,191,111}
			\definecolor{colorbrewer8}{RGB}{255,127,0}
			\definecolor{colorbrewer9}{RGB}{202,178,214}
			\definecolor{colorbrewer10}{RGB}{106,61,154}
			\definecolor{colorbrewer11}{RGB}{255,255,153}
		},
		colorbrewer values/12/.code={
			\definecolor{colorbrewer1}{RGB}{166,206,227}
			\definecolor{colorbrewer2}{RGB}{31,120,180}
			\definecolor{colorbrewer3}{RGB}{178,223,138}
			\definecolor{colorbrewer4}{RGB}{51,160,44}
			\definecolor{colorbrewer5}{RGB}{251,154,153}
			\definecolor{colorbrewer6}{RGB}{227,26,28}
			\definecolor{colorbrewer7}{RGB}{253,191,111}
			\definecolor{colorbrewer8}{RGB}{255,127,0}
			\definecolor{colorbrewer9}{RGB}{202,178,214}
			\definecolor{colorbrewer10}{RGB}{106,61,154}
			\definecolor{colorbrewer11}{RGB}{255,255,153}
			\definecolor{colorbrewer12}{RGB}{177,89,40}
		}
	},
	colorbrewer scheme/Pastel1/.style={
		colorbrewer values/.is choice,	
		colorbrewer values/3/.code={
			\definecolor{colorbrewer1}{RGB}{251,180,174}
			\definecolor{colorbrewer2}{RGB}{179,205,227}
			\definecolor{colorbrewer3}{RGB}{204,235,197}
		},
		colorbrewer values=3,
		colorbrewer values/4/.code={
			\definecolor{colorbrewer1}{RGB}{251,180,174}
			\definecolor{colorbrewer2}{RGB}{179,205,227}
			\definecolor{colorbrewer3}{RGB}{204,235,197}
			\definecolor{colorbrewer4}{RGB}{222,203,228}
		},
		colorbrewer values/5/.code={
			\definecolor{colorbrewer1}{RGB}{251,180,174}
			\definecolor{colorbrewer2}{RGB}{179,205,227}
			\definecolor{colorbrewer3}{RGB}{204,235,197}
			\definecolor{colorbrewer4}{RGB}{222,203,228}
			\definecolor{colorbrewer5}{RGB}{254,217,166}
		},
		colorbrewer values/6/.code={
			\definecolor{colorbrewer1}{RGB}{251,180,174}
			\definecolor{colorbrewer2}{RGB}{179,205,227}
			\definecolor{colorbrewer3}{RGB}{204,235,197}
			\definecolor{colorbrewer4}{RGB}{222,203,228}
			\definecolor{colorbrewer5}{RGB}{254,217,166}
			\definecolor{colorbrewer6}{RGB}{255,255,204}
		},
		colorbrewer values/7/.code={
			\definecolor{colorbrewer1}{RGB}{251,180,174}
			\definecolor{colorbrewer2}{RGB}{179,205,227}
			\definecolor{colorbrewer3}{RGB}{204,235,197}
			\definecolor{colorbrewer4}{RGB}{222,203,228}
			\definecolor{colorbrewer5}{RGB}{254,217,166}
			\definecolor{colorbrewer6}{RGB}{255,255,204}
			\definecolor{colorbrewer7}{RGB}{229,216,189}
		},
		colorbrewer values/8/.code={
			\definecolor{colorbrewer1}{RGB}{251,180,174}
			\definecolor{colorbrewer2}{RGB}{179,205,227}
			\definecolor{colorbrewer3}{RGB}{204,235,197}
			\definecolor{colorbrewer4}{RGB}{222,203,228}
			\definecolor{colorbrewer5}{RGB}{254,217,166}
			\definecolor{colorbrewer6}{RGB}{255,255,204}
			\definecolor{colorbrewer7}{RGB}{229,216,189}
			\definecolor{colorbrewer8}{RGB}{253,218,236}
		},
		colorbrewer values/9/.code={
			\definecolor{colorbrewer1}{RGB}{251,180,174}
			\definecolor{colorbrewer2}{RGB}{179,205,227}
			\definecolor{colorbrewer3}{RGB}{204,235,197}
			\definecolor{colorbrewer4}{RGB}{222,203,228}
			\definecolor{colorbrewer5}{RGB}{254,217,166}
			\definecolor{colorbrewer6}{RGB}{255,255,204}
			\definecolor{colorbrewer7}{RGB}{229,216,189}
			\definecolor{colorbrewer8}{RGB}{253,218,236}
			\definecolor{colorbrewer9}{RGB}{242,242,242}
		}
	},
	colorbrewer scheme/Pastel2/.style={
		colorbrewer values/.is choice,	
		colorbrewer values/3/.code={
			\definecolor{colorbrewer1}{RGB}{179,226,205}
			\definecolor{colorbrewer2}{RGB}{253,205,172}
			\definecolor{colorbrewer3}{RGB}{203,213,232}
		},
		colorbrewer values=3,
		colorbrewer values/4/.code={
			\definecolor{colorbrewer1}{RGB}{179,226,205}
			\definecolor{colorbrewer2}{RGB}{253,205,172}
			\definecolor{colorbrewer3}{RGB}{203,213,232}
			\definecolor{colorbrewer4}{RGB}{244,202,228}
		},
		colorbrewer values/5/.code={
			\definecolor{colorbrewer1}{RGB}{179,226,205}
			\definecolor{colorbrewer2}{RGB}{253,205,172}
			\definecolor{colorbrewer3}{RGB}{203,213,232}
			\definecolor{colorbrewer4}{RGB}{244,202,228}
			\definecolor{colorbrewer5}{RGB}{230,245,201}
		},
		colorbrewer values/6/.code={
			\definecolor{colorbrewer1}{RGB}{179,226,205}
			\definecolor{colorbrewer2}{RGB}{253,205,172}
			\definecolor{colorbrewer3}{RGB}{203,213,232}
			\definecolor{colorbrewer4}{RGB}{244,202,228}
			\definecolor{colorbrewer5}{RGB}{230,245,201}
			\definecolor{colorbrewer6}{RGB}{255,242,174}
		},
		colorbrewer values/7/.code={
			\definecolor{colorbrewer1}{RGB}{179,226,205}
			\definecolor{colorbrewer2}{RGB}{253,205,172}
			\definecolor{colorbrewer3}{RGB}{203,213,232}
			\definecolor{colorbrewer4}{RGB}{244,202,228}
			\definecolor{colorbrewer5}{RGB}{230,245,201}
			\definecolor{colorbrewer6}{RGB}{255,242,174}
			\definecolor{colorbrewer7}{RGB}{241,226,204}
		},
		colorbrewer values/8/.code={
			\definecolor{colorbrewer1}{RGB}{179,226,205}
			\definecolor{colorbrewer2}{RGB}{253,205,172}
			\definecolor{colorbrewer3}{RGB}{203,213,232}
			\definecolor{colorbrewer4}{RGB}{244,202,228}
			\definecolor{colorbrewer5}{RGB}{230,245,201}
			\definecolor{colorbrewer6}{RGB}{255,242,174}
			\definecolor{colorbrewer7}{RGB}{241,226,204}
			\definecolor{colorbrewer8}{RGB}{204,204,204}
		}
	},
	colorbrewer scheme/PiYG/.style={
		colorbrewer values/.is choice,	
		colorbrewer values/3/.code={
			\definecolor{colorbrewer1}{RGB}{233,163,201}
			\definecolor{colorbrewer2}{RGB}{247,247,247}
			\definecolor{colorbrewer3}{RGB}{161,215,106}
		},
		colorbrewer values=3,
		colorbrewer values/4/.code={
			\definecolor{colorbrewer1}{RGB}{208,28,139}
			\definecolor{colorbrewer2}{RGB}{241,182,218}
			\definecolor{colorbrewer3}{RGB}{184,225,134}
			\definecolor{colorbrewer4}{RGB}{77,172,38}
		},
		colorbrewer values/5/.code={
			\definecolor{colorbrewer1}{RGB}{208,28,139}
			\definecolor{colorbrewer2}{RGB}{241,182,218}
			\definecolor{colorbrewer3}{RGB}{247,247,247}
			\definecolor{colorbrewer4}{RGB}{184,225,134}
			\definecolor{colorbrewer5}{RGB}{77,172,38}
		},
		colorbrewer values/6/.code={
			\definecolor{colorbrewer1}{RGB}{197,27,125}
			\definecolor{colorbrewer2}{RGB}{233,163,201}
			\definecolor{colorbrewer3}{RGB}{253,224,239}
			\definecolor{colorbrewer4}{RGB}{230,245,208}
			\definecolor{colorbrewer5}{RGB}{161,215,106}
			\definecolor{colorbrewer6}{RGB}{77,146,33}
		},
		colorbrewer values/7/.code={
			\definecolor{colorbrewer1}{RGB}{197,27,125}
			\definecolor{colorbrewer2}{RGB}{233,163,201}
			\definecolor{colorbrewer3}{RGB}{253,224,239}
			\definecolor{colorbrewer4}{RGB}{247,247,247}
			\definecolor{colorbrewer5}{RGB}{230,245,208}
			\definecolor{colorbrewer6}{RGB}{161,215,106}
			\definecolor{colorbrewer7}{RGB}{77,146,33}
		},
		colorbrewer values/8/.code={
			\definecolor{colorbrewer1}{RGB}{197,27,125}
			\definecolor{colorbrewer2}{RGB}{222,119,174}
			\definecolor{colorbrewer3}{RGB}{241,182,218}
			\definecolor{colorbrewer4}{RGB}{253,224,239}
			\definecolor{colorbrewer5}{RGB}{230,245,208}
			\definecolor{colorbrewer6}{RGB}{184,225,134}
			\definecolor{colorbrewer7}{RGB}{127,188,65}
			\definecolor{colorbrewer8}{RGB}{77,146,33}
		},
		colorbrewer values/9/.code={
			\definecolor{colorbrewer1}{RGB}{197,27,125}
			\definecolor{colorbrewer2}{RGB}{222,119,174}
			\definecolor{colorbrewer3}{RGB}{241,182,218}
			\definecolor{colorbrewer4}{RGB}{253,224,239}
			\definecolor{colorbrewer5}{RGB}{247,247,247}
			\definecolor{colorbrewer6}{RGB}{230,245,208}
			\definecolor{colorbrewer7}{RGB}{184,225,134}
			\definecolor{colorbrewer8}{RGB}{127,188,65}
			\definecolor{colorbrewer9}{RGB}{77,146,33}
		},
		colorbrewer values/10/.code={
			\definecolor{colorbrewer1}{RGB}{142,1,82}
			\definecolor{colorbrewer2}{RGB}{197,27,125}
			\definecolor{colorbrewer3}{RGB}{222,119,174}
			\definecolor{colorbrewer4}{RGB}{241,182,218}
			\definecolor{colorbrewer5}{RGB}{253,224,239}
			\definecolor{colorbrewer6}{RGB}{230,245,208}
			\definecolor{colorbrewer7}{RGB}{184,225,134}
			\definecolor{colorbrewer8}{RGB}{127,188,65}
			\definecolor{colorbrewer9}{RGB}{77,146,33}
			\definecolor{colorbrewer10}{RGB}{39,100,25}
		},
		colorbrewer values/11/.code={
			\definecolor{colorbrewer1}{RGB}{142,1,82}
			\definecolor{colorbrewer2}{RGB}{197,27,125}
			\definecolor{colorbrewer3}{RGB}{222,119,174}
			\definecolor{colorbrewer4}{RGB}{241,182,218}
			\definecolor{colorbrewer5}{RGB}{253,224,239}
			\definecolor{colorbrewer6}{RGB}{247,247,247}
			\definecolor{colorbrewer7}{RGB}{230,245,208}
			\definecolor{colorbrewer8}{RGB}{184,225,134}
			\definecolor{colorbrewer9}{RGB}{127,188,65}
			\definecolor{colorbrewer10}{RGB}{77,146,33}
			\definecolor{colorbrewer11}{RGB}{39,100,25}
		}
	},
	colorbrewer scheme/PRGn/.style={
		colorbrewer values/.is choice,	
		colorbrewer values/3/.code={
			\definecolor{colorbrewer1}{RGB}{175,141,195}
			\definecolor{colorbrewer2}{RGB}{247,247,247}
			\definecolor{colorbrewer3}{RGB}{127,191,123}
		},
		colorbrewer values=3,
		colorbrewer values/4/.code={
			\definecolor{colorbrewer1}{RGB}{123,50,148}
			\definecolor{colorbrewer2}{RGB}{194,165,207}
			\definecolor{colorbrewer3}{RGB}{166,219,160}
			\definecolor{colorbrewer4}{RGB}{0,136,55}
		},
		colorbrewer values/5/.code={
			\definecolor{colorbrewer1}{RGB}{123,50,148}
			\definecolor{colorbrewer2}{RGB}{194,165,207}
			\definecolor{colorbrewer3}{RGB}{247,247,247}
			\definecolor{colorbrewer4}{RGB}{166,219,160}
			\definecolor{colorbrewer5}{RGB}{0,136,55}
		},
		colorbrewer values/6/.code={
			\definecolor{colorbrewer1}{RGB}{118,42,131}
			\definecolor{colorbrewer2}{RGB}{175,141,195}
			\definecolor{colorbrewer3}{RGB}{231,212,232}
			\definecolor{colorbrewer4}{RGB}{217,240,211}
			\definecolor{colorbrewer5}{RGB}{127,191,123}
			\definecolor{colorbrewer6}{RGB}{27,120,55}
		},
		colorbrewer values/7/.code={
			\definecolor{colorbrewer1}{RGB}{118,42,131}
			\definecolor{colorbrewer2}{RGB}{175,141,195}
			\definecolor{colorbrewer3}{RGB}{231,212,232}
			\definecolor{colorbrewer4}{RGB}{247,247,247}
			\definecolor{colorbrewer5}{RGB}{217,240,211}
			\definecolor{colorbrewer6}{RGB}{127,191,123}
			\definecolor{colorbrewer7}{RGB}{27,120,55}
		},
		colorbrewer values/8/.code={
			\definecolor{colorbrewer1}{RGB}{118,42,131}
			\definecolor{colorbrewer2}{RGB}{153,112,171}
			\definecolor{colorbrewer3}{RGB}{194,165,207}
			\definecolor{colorbrewer4}{RGB}{231,212,232}
			\definecolor{colorbrewer5}{RGB}{217,240,211}
			\definecolor{colorbrewer6}{RGB}{166,219,160}
			\definecolor{colorbrewer7}{RGB}{90,174,97}
			\definecolor{colorbrewer8}{RGB}{27,120,55}
		},
		colorbrewer values/9/.code={
			\definecolor{colorbrewer1}{RGB}{118,42,131}
			\definecolor{colorbrewer2}{RGB}{153,112,171}
			\definecolor{colorbrewer3}{RGB}{194,165,207}
			\definecolor{colorbrewer4}{RGB}{231,212,232}
			\definecolor{colorbrewer5}{RGB}{247,247,247}
			\definecolor{colorbrewer6}{RGB}{217,240,211}
			\definecolor{colorbrewer7}{RGB}{166,219,160}
			\definecolor{colorbrewer8}{RGB}{90,174,97}
			\definecolor{colorbrewer9}{RGB}{27,120,55}
		},
		colorbrewer values/10/.code={
			\definecolor{colorbrewer1}{RGB}{64,0,75}
			\definecolor{colorbrewer2}{RGB}{118,42,131}
			\definecolor{colorbrewer3}{RGB}{153,112,171}
			\definecolor{colorbrewer4}{RGB}{194,165,207}
			\definecolor{colorbrewer5}{RGB}{231,212,232}
			\definecolor{colorbrewer6}{RGB}{217,240,211}
			\definecolor{colorbrewer7}{RGB}{166,219,160}
			\definecolor{colorbrewer8}{RGB}{90,174,97}
			\definecolor{colorbrewer9}{RGB}{27,120,55}
			\definecolor{colorbrewer10}{RGB}{0,68,27}
		},
		colorbrewer values/11/.code={
			\definecolor{colorbrewer1}{RGB}{64,0,75}
			\definecolor{colorbrewer2}{RGB}{118,42,131}
			\definecolor{colorbrewer3}{RGB}{153,112,171}
			\definecolor{colorbrewer4}{RGB}{194,165,207}
			\definecolor{colorbrewer5}{RGB}{231,212,232}
			\definecolor{colorbrewer6}{RGB}{247,247,247}
			\definecolor{colorbrewer7}{RGB}{217,240,211}
			\definecolor{colorbrewer8}{RGB}{166,219,160}
			\definecolor{colorbrewer9}{RGB}{90,174,97}
			\definecolor{colorbrewer10}{RGB}{27,120,55}
			\definecolor{colorbrewer11}{RGB}{0,68,27}
		}
	},
	colorbrewer scheme/PuBu/.style={
		colorbrewer values/.is choice,	
		colorbrewer values/3/.code={
			\definecolor{colorbrewer1}{RGB}{236,231,242}
			\definecolor{colorbrewer2}{RGB}{166,189,219}
			\definecolor{colorbrewer3}{RGB}{43,140,190}
		},
		colorbrewer values=3,
		colorbrewer values/4/.code={
			\definecolor{colorbrewer1}{RGB}{241,238,246}
			\definecolor{colorbrewer2}{RGB}{189,201,225}
			\definecolor{colorbrewer3}{RGB}{116,169,207}
			\definecolor{colorbrewer4}{RGB}{5,112,176}
		},
		colorbrewer values/5/.code={
			\definecolor{colorbrewer1}{RGB}{241,238,246}
			\definecolor{colorbrewer2}{RGB}{189,201,225}
			\definecolor{colorbrewer3}{RGB}{116,169,207}
			\definecolor{colorbrewer4}{RGB}{43,140,190}
			\definecolor{colorbrewer5}{RGB}{4,90,141}
		},
		colorbrewer values/6/.code={
			\definecolor{colorbrewer1}{RGB}{241,238,246}
			\definecolor{colorbrewer2}{RGB}{208,209,230}
			\definecolor{colorbrewer3}{RGB}{166,189,219}
			\definecolor{colorbrewer4}{RGB}{116,169,207}
			\definecolor{colorbrewer5}{RGB}{43,140,190}
			\definecolor{colorbrewer6}{RGB}{4,90,141}
		},
		colorbrewer values/7/.code={
			\definecolor{colorbrewer1}{RGB}{241,238,246}
			\definecolor{colorbrewer2}{RGB}{208,209,230}
			\definecolor{colorbrewer3}{RGB}{166,189,219}
			\definecolor{colorbrewer4}{RGB}{116,169,207}
			\definecolor{colorbrewer5}{RGB}{54,144,192}
			\definecolor{colorbrewer6}{RGB}{5,112,176}
			\definecolor{colorbrewer7}{RGB}{3,78,123}
		},
		colorbrewer values/8/.code={
			\definecolor{colorbrewer1}{RGB}{255,247,251}
			\definecolor{colorbrewer2}{RGB}{236,231,242}
			\definecolor{colorbrewer3}{RGB}{208,209,230}
			\definecolor{colorbrewer4}{RGB}{166,189,219}
			\definecolor{colorbrewer5}{RGB}{116,169,207}
			\definecolor{colorbrewer6}{RGB}{54,144,192}
			\definecolor{colorbrewer7}{RGB}{5,112,176}
			\definecolor{colorbrewer8}{RGB}{3,78,123}
		},
		colorbrewer values/9/.code={
			\definecolor{colorbrewer1}{RGB}{255,247,251}
			\definecolor{colorbrewer2}{RGB}{236,231,242}
			\definecolor{colorbrewer3}{RGB}{208,209,230}
			\definecolor{colorbrewer4}{RGB}{166,189,219}
			\definecolor{colorbrewer5}{RGB}{116,169,207}
			\definecolor{colorbrewer6}{RGB}{54,144,192}
			\definecolor{colorbrewer7}{RGB}{5,112,176}
			\definecolor{colorbrewer8}{RGB}{4,90,141}
			\definecolor{colorbrewer9}{RGB}{2,56,88}
		}
	},
	colorbrewer scheme/PuBuGn/.style={
		colorbrewer values/.is choice,	
		colorbrewer values/3/.code={
			\definecolor{colorbrewer1}{RGB}{236,226,240}
			\definecolor{colorbrewer2}{RGB}{166,189,219}
			\definecolor{colorbrewer3}{RGB}{28,144,153}
		},
		colorbrewer values=3,
		colorbrewer values/4/.code={
			\definecolor{colorbrewer1}{RGB}{246,239,247}
			\definecolor{colorbrewer2}{RGB}{189,201,225}
			\definecolor{colorbrewer3}{RGB}{103,169,207}
			\definecolor{colorbrewer4}{RGB}{2,129,138}
		},
		colorbrewer values/5/.code={
			\definecolor{colorbrewer1}{RGB}{246,239,247}
			\definecolor{colorbrewer2}{RGB}{189,201,225}
			\definecolor{colorbrewer3}{RGB}{103,169,207}
			\definecolor{colorbrewer4}{RGB}{28,144,153}
			\definecolor{colorbrewer5}{RGB}{1,108,89}
		},
		colorbrewer values/6/.code={
			\definecolor{colorbrewer1}{RGB}{246,239,247}
			\definecolor{colorbrewer2}{RGB}{208,209,230}
			\definecolor{colorbrewer3}{RGB}{166,189,219}
			\definecolor{colorbrewer4}{RGB}{103,169,207}
			\definecolor{colorbrewer5}{RGB}{28,144,153}
			\definecolor{colorbrewer6}{RGB}{1,108,89}
		},
		colorbrewer values/7/.code={
			\definecolor{colorbrewer1}{RGB}{246,239,247}
			\definecolor{colorbrewer2}{RGB}{208,209,230}
			\definecolor{colorbrewer3}{RGB}{166,189,219}
			\definecolor{colorbrewer4}{RGB}{103,169,207}
			\definecolor{colorbrewer5}{RGB}{54,144,192}
			\definecolor{colorbrewer6}{RGB}{2,129,138}
			\definecolor{colorbrewer7}{RGB}{1,100,80}
		},
		colorbrewer values/8/.code={
			\definecolor{colorbrewer1}{RGB}{255,247,251}
			\definecolor{colorbrewer2}{RGB}{236,226,240}
			\definecolor{colorbrewer3}{RGB}{208,209,230}
			\definecolor{colorbrewer4}{RGB}{166,189,219}
			\definecolor{colorbrewer5}{RGB}{103,169,207}
			\definecolor{colorbrewer6}{RGB}{54,144,192}
			\definecolor{colorbrewer7}{RGB}{2,129,138}
			\definecolor{colorbrewer8}{RGB}{1,100,80}
		},
		colorbrewer values/9/.code={
			\definecolor{colorbrewer1}{RGB}{255,247,251}
			\definecolor{colorbrewer2}{RGB}{236,226,240}
			\definecolor{colorbrewer3}{RGB}{208,209,230}
			\definecolor{colorbrewer4}{RGB}{166,189,219}
			\definecolor{colorbrewer5}{RGB}{103,169,207}
			\definecolor{colorbrewer6}{RGB}{54,144,192}
			\definecolor{colorbrewer7}{RGB}{2,129,138}
			\definecolor{colorbrewer8}{RGB}{1,108,89}
			\definecolor{colorbrewer9}{RGB}{1,70,54}
		}
	},
	colorbrewer scheme/PuOr/.style={
		colorbrewer values/.is choice,	
		colorbrewer values/3/.code={
			\definecolor{colorbrewer1}{RGB}{241,163,64}
			\definecolor{colorbrewer2}{RGB}{247,247,247}
			\definecolor{colorbrewer3}{RGB}{153,142,195}
		},
		colorbrewer values=3,
		colorbrewer values/4/.code={
			\definecolor{colorbrewer1}{RGB}{230,97,1}
			\definecolor{colorbrewer2}{RGB}{253,184,99}
			\definecolor{colorbrewer3}{RGB}{178,171,210}
			\definecolor{colorbrewer4}{RGB}{94,60,153}
		},
		colorbrewer values/5/.code={
			\definecolor{colorbrewer1}{RGB}{230,97,1}
			\definecolor{colorbrewer2}{RGB}{253,184,99}
			\definecolor{colorbrewer3}{RGB}{247,247,247}
			\definecolor{colorbrewer4}{RGB}{178,171,210}
			\definecolor{colorbrewer5}{RGB}{94,60,153}
		},
		colorbrewer values/6/.code={
			\definecolor{colorbrewer1}{RGB}{179,88,6}
			\definecolor{colorbrewer2}{RGB}{241,163,64}
			\definecolor{colorbrewer3}{RGB}{254,224,182}
			\definecolor{colorbrewer4}{RGB}{216,218,235}
			\definecolor{colorbrewer5}{RGB}{153,142,195}
			\definecolor{colorbrewer6}{RGB}{84,39,136}
		},
		colorbrewer values/7/.code={
			\definecolor{colorbrewer1}{RGB}{179,88,6}
			\definecolor{colorbrewer2}{RGB}{241,163,64}
			\definecolor{colorbrewer3}{RGB}{254,224,182}
			\definecolor{colorbrewer4}{RGB}{247,247,247}
			\definecolor{colorbrewer5}{RGB}{216,218,235}
			\definecolor{colorbrewer6}{RGB}{153,142,195}
			\definecolor{colorbrewer7}{RGB}{84,39,136}
		},
		colorbrewer values/8/.code={
			\definecolor{colorbrewer1}{RGB}{179,88,6}
			\definecolor{colorbrewer2}{RGB}{224,130,20}
			\definecolor{colorbrewer3}{RGB}{253,184,99}
			\definecolor{colorbrewer4}{RGB}{254,224,182}
			\definecolor{colorbrewer5}{RGB}{216,218,235}
			\definecolor{colorbrewer6}{RGB}{178,171,210}
			\definecolor{colorbrewer7}{RGB}{128,115,172}
			\definecolor{colorbrewer8}{RGB}{84,39,136}
		},
		colorbrewer values/9/.code={
			\definecolor{colorbrewer1}{RGB}{179,88,6}
			\definecolor{colorbrewer2}{RGB}{224,130,20}
			\definecolor{colorbrewer3}{RGB}{253,184,99}
			\definecolor{colorbrewer4}{RGB}{254,224,182}
			\definecolor{colorbrewer5}{RGB}{247,247,247}
			\definecolor{colorbrewer6}{RGB}{216,218,235}
			\definecolor{colorbrewer7}{RGB}{178,171,210}
			\definecolor{colorbrewer8}{RGB}{128,115,172}
			\definecolor{colorbrewer9}{RGB}{84,39,136}
		},
		colorbrewer values/10/.code={
			\definecolor{colorbrewer1}{RGB}{127,59,8}
			\definecolor{colorbrewer2}{RGB}{179,88,6}
			\definecolor{colorbrewer3}{RGB}{224,130,20}
			\definecolor{colorbrewer4}{RGB}{253,184,99}
			\definecolor{colorbrewer5}{RGB}{254,224,182}
			\definecolor{colorbrewer6}{RGB}{216,218,235}
			\definecolor{colorbrewer7}{RGB}{178,171,210}
			\definecolor{colorbrewer8}{RGB}{128,115,172}
			\definecolor{colorbrewer9}{RGB}{84,39,136}
			\definecolor{colorbrewer10}{RGB}{45,0,75}
		},
		colorbrewer values/11/.code={
			\definecolor{colorbrewer1}{RGB}{127,59,8}
			\definecolor{colorbrewer2}{RGB}{179,88,6}
			\definecolor{colorbrewer3}{RGB}{224,130,20}
			\definecolor{colorbrewer4}{RGB}{253,184,99}
			\definecolor{colorbrewer5}{RGB}{254,224,182}
			\definecolor{colorbrewer6}{RGB}{247,247,247}
			\definecolor{colorbrewer7}{RGB}{216,218,235}
			\definecolor{colorbrewer8}{RGB}{178,171,210}
			\definecolor{colorbrewer9}{RGB}{128,115,172}
			\definecolor{colorbrewer10}{RGB}{84,39,136}
			\definecolor{colorbrewer11}{RGB}{45,0,75}
		}
	},
	colorbrewer scheme/PuRd/.style={
		colorbrewer values/.is choice,	
		colorbrewer values/3/.code={
			\definecolor{colorbrewer1}{RGB}{231,225,239}
			\definecolor{colorbrewer2}{RGB}{201,148,199}
			\definecolor{colorbrewer3}{RGB}{221,28,119}
		},
		colorbrewer values=3,
		colorbrewer values/4/.code={
			\definecolor{colorbrewer1}{RGB}{241,238,246}
			\definecolor{colorbrewer2}{RGB}{215,181,216}
			\definecolor{colorbrewer3}{RGB}{223,101,176}
			\definecolor{colorbrewer4}{RGB}{206,18,86}
		},
		colorbrewer values/5/.code={
			\definecolor{colorbrewer1}{RGB}{241,238,246}
			\definecolor{colorbrewer2}{RGB}{215,181,216}
			\definecolor{colorbrewer3}{RGB}{223,101,176}
			\definecolor{colorbrewer4}{RGB}{221,28,119}
			\definecolor{colorbrewer5}{RGB}{152,0,67}
		},
		colorbrewer values/6/.code={
			\definecolor{colorbrewer1}{RGB}{241,238,246}
			\definecolor{colorbrewer2}{RGB}{212,185,218}
			\definecolor{colorbrewer3}{RGB}{201,148,199}
			\definecolor{colorbrewer4}{RGB}{223,101,176}
			\definecolor{colorbrewer5}{RGB}{221,28,119}
			\definecolor{colorbrewer6}{RGB}{152,0,67}
		},
		colorbrewer values/7/.code={
			\definecolor{colorbrewer1}{RGB}{241,238,246}
			\definecolor{colorbrewer2}{RGB}{212,185,218}
			\definecolor{colorbrewer3}{RGB}{201,148,199}
			\definecolor{colorbrewer4}{RGB}{223,101,176}
			\definecolor{colorbrewer5}{RGB}{231,41,138}
			\definecolor{colorbrewer6}{RGB}{206,18,86}
			\definecolor{colorbrewer7}{RGB}{145,0,63}
		},
		colorbrewer values/8/.code={
			\definecolor{colorbrewer1}{RGB}{247,244,249}
			\definecolor{colorbrewer2}{RGB}{231,225,239}
			\definecolor{colorbrewer3}{RGB}{212,185,218}
			\definecolor{colorbrewer4}{RGB}{201,148,199}
			\definecolor{colorbrewer5}{RGB}{223,101,176}
			\definecolor{colorbrewer6}{RGB}{231,41,138}
			\definecolor{colorbrewer7}{RGB}{206,18,86}
			\definecolor{colorbrewer8}{RGB}{145,0,63}
		},
		colorbrewer values/9/.code={
			\definecolor{colorbrewer1}{RGB}{247,244,249}
			\definecolor{colorbrewer2}{RGB}{231,225,239}
			\definecolor{colorbrewer3}{RGB}{212,185,218}
			\definecolor{colorbrewer4}{RGB}{201,148,199}
			\definecolor{colorbrewer5}{RGB}{223,101,176}
			\definecolor{colorbrewer6}{RGB}{231,41,138}
			\definecolor{colorbrewer7}{RGB}{206,18,86}
			\definecolor{colorbrewer8}{RGB}{152,0,67}
			\definecolor{colorbrewer9}{RGB}{103,0,31}
		}
	},
	colorbrewer scheme/Purples/.style={
		colorbrewer values/.is choice,	
		colorbrewer values/3/.code={
			\definecolor{colorbrewer1}{RGB}{239,237,245}
			\definecolor{colorbrewer2}{RGB}{188,189,220}
			\definecolor{colorbrewer3}{RGB}{117,107,177}
		},
		colorbrewer values=3,
		colorbrewer values/4/.code={
			\definecolor{colorbrewer1}{RGB}{242,240,247}
			\definecolor{colorbrewer2}{RGB}{203,201,226}
			\definecolor{colorbrewer3}{RGB}{158,154,200}
			\definecolor{colorbrewer4}{RGB}{106,81,163}
		},
		colorbrewer values/5/.code={
			\definecolor{colorbrewer1}{RGB}{242,240,247}
			\definecolor{colorbrewer2}{RGB}{203,201,226}
			\definecolor{colorbrewer3}{RGB}{158,154,200}
			\definecolor{colorbrewer4}{RGB}{117,107,177}
			\definecolor{colorbrewer5}{RGB}{84,39,143}
		},
		colorbrewer values/6/.code={
			\definecolor{colorbrewer1}{RGB}{242,240,247}
			\definecolor{colorbrewer2}{RGB}{218,218,235}
			\definecolor{colorbrewer3}{RGB}{188,189,220}
			\definecolor{colorbrewer4}{RGB}{158,154,200}
			\definecolor{colorbrewer5}{RGB}{117,107,177}
			\definecolor{colorbrewer6}{RGB}{84,39,143}
		},
		colorbrewer values/7/.code={
			\definecolor{colorbrewer1}{RGB}{242,240,247}
			\definecolor{colorbrewer2}{RGB}{218,218,235}
			\definecolor{colorbrewer3}{RGB}{188,189,220}
			\definecolor{colorbrewer4}{RGB}{158,154,200}
			\definecolor{colorbrewer5}{RGB}{128,125,186}
			\definecolor{colorbrewer6}{RGB}{106,81,163}
			\definecolor{colorbrewer7}{RGB}{74,20,134}
		},
		colorbrewer values/8/.code={
			\definecolor{colorbrewer1}{RGB}{252,251,253}
			\definecolor{colorbrewer2}{RGB}{239,237,245}
			\definecolor{colorbrewer3}{RGB}{218,218,235}
			\definecolor{colorbrewer4}{RGB}{188,189,220}
			\definecolor{colorbrewer5}{RGB}{158,154,200}
			\definecolor{colorbrewer6}{RGB}{128,125,186}
			\definecolor{colorbrewer7}{RGB}{106,81,163}
			\definecolor{colorbrewer8}{RGB}{74,20,134}
		},
		colorbrewer values/9/.code={
			\definecolor{colorbrewer1}{RGB}{252,251,253}
			\definecolor{colorbrewer2}{RGB}{239,237,245}
			\definecolor{colorbrewer3}{RGB}{218,218,235}
			\definecolor{colorbrewer4}{RGB}{188,189,220}
			\definecolor{colorbrewer5}{RGB}{158,154,200}
			\definecolor{colorbrewer6}{RGB}{128,125,186}
			\definecolor{colorbrewer7}{RGB}{106,81,163}
			\definecolor{colorbrewer8}{RGB}{84,39,143}
			\definecolor{colorbrewer9}{RGB}{63,0,125}
		}
	},
	colorbrewer scheme/RdBu/.style={
		colorbrewer values/.is choice,	
		colorbrewer values/3/.code={
			\definecolor{colorbrewer1}{RGB}{239,138,98}
			\definecolor{colorbrewer2}{RGB}{247,247,247}
			\definecolor{colorbrewer3}{RGB}{103,169,207}
		},
		colorbrewer values=3,
		colorbrewer values/4/.code={
			\definecolor{colorbrewer1}{RGB}{202,0,32}
			\definecolor{colorbrewer2}{RGB}{244,165,130}
			\definecolor{colorbrewer3}{RGB}{146,197,222}
			\definecolor{colorbrewer4}{RGB}{5,113,176}
		},
		colorbrewer values/5/.code={
			\definecolor{colorbrewer1}{RGB}{202,0,32}
			\definecolor{colorbrewer2}{RGB}{244,165,130}
			\definecolor{colorbrewer3}{RGB}{247,247,247}
			\definecolor{colorbrewer4}{RGB}{146,197,222}
			\definecolor{colorbrewer5}{RGB}{5,113,176}
		},
		colorbrewer values/6/.code={
			\definecolor{colorbrewer1}{RGB}{178,24,43}
			\definecolor{colorbrewer2}{RGB}{239,138,98}
			\definecolor{colorbrewer3}{RGB}{253,219,199}
			\definecolor{colorbrewer4}{RGB}{209,229,240}
			\definecolor{colorbrewer5}{RGB}{103,169,207}
			\definecolor{colorbrewer6}{RGB}{33,102,172}
		},
		colorbrewer values/7/.code={
			\definecolor{colorbrewer1}{RGB}{178,24,43}
			\definecolor{colorbrewer2}{RGB}{239,138,98}
			\definecolor{colorbrewer3}{RGB}{253,219,199}
			\definecolor{colorbrewer4}{RGB}{247,247,247}
			\definecolor{colorbrewer5}{RGB}{209,229,240}
			\definecolor{colorbrewer6}{RGB}{103,169,207}
			\definecolor{colorbrewer7}{RGB}{33,102,172}
		},
		colorbrewer values/8/.code={
			\definecolor{colorbrewer1}{RGB}{178,24,43}
			\definecolor{colorbrewer2}{RGB}{214,96,77}
			\definecolor{colorbrewer3}{RGB}{244,165,130}
			\definecolor{colorbrewer4}{RGB}{253,219,199}
			\definecolor{colorbrewer5}{RGB}{209,229,240}
			\definecolor{colorbrewer6}{RGB}{146,197,222}
			\definecolor{colorbrewer7}{RGB}{67,147,195}
			\definecolor{colorbrewer8}{RGB}{33,102,172}
		},
		colorbrewer values/9/.code={
			\definecolor{colorbrewer1}{RGB}{178,24,43}
			\definecolor{colorbrewer2}{RGB}{214,96,77}
			\definecolor{colorbrewer3}{RGB}{244,165,130}
			\definecolor{colorbrewer4}{RGB}{253,219,199}
			\definecolor{colorbrewer5}{RGB}{247,247,247}
			\definecolor{colorbrewer6}{RGB}{209,229,240}
			\definecolor{colorbrewer7}{RGB}{146,197,222}
			\definecolor{colorbrewer8}{RGB}{67,147,195}
			\definecolor{colorbrewer9}{RGB}{33,102,172}
		},
		colorbrewer values/10/.code={
			\definecolor{colorbrewer1}{RGB}{103,0,31}
			\definecolor{colorbrewer2}{RGB}{178,24,43}
			\definecolor{colorbrewer3}{RGB}{214,96,77}
			\definecolor{colorbrewer4}{RGB}{244,165,130}
			\definecolor{colorbrewer5}{RGB}{253,219,199}
			\definecolor{colorbrewer6}{RGB}{209,229,240}
			\definecolor{colorbrewer7}{RGB}{146,197,222}
			\definecolor{colorbrewer8}{RGB}{67,147,195}
			\definecolor{colorbrewer9}{RGB}{33,102,172}
			\definecolor{colorbrewer10}{RGB}{5,48,97}
		},
		colorbrewer values/11/.code={
			\definecolor{colorbrewer1}{RGB}{103,0,31}
			\definecolor{colorbrewer2}{RGB}{178,24,43}
			\definecolor{colorbrewer3}{RGB}{214,96,77}
			\definecolor{colorbrewer4}{RGB}{244,165,130}
			\definecolor{colorbrewer5}{RGB}{253,219,199}
			\definecolor{colorbrewer6}{RGB}{247,247,247}
			\definecolor{colorbrewer7}{RGB}{209,229,240}
			\definecolor{colorbrewer8}{RGB}{146,197,222}
			\definecolor{colorbrewer9}{RGB}{67,147,195}
			\definecolor{colorbrewer10}{RGB}{33,102,172}
			\definecolor{colorbrewer11}{RGB}{5,48,97}
		}
	},
	colorbrewer scheme/RdGy/.style={
		colorbrewer values/.is choice,	
		colorbrewer values/3/.code={
			\definecolor{colorbrewer1}{RGB}{239,138,98}
			\definecolor{colorbrewer2}{RGB}{255,255,255}
			\definecolor{colorbrewer3}{RGB}{153,153,153}
		},
		colorbrewer values=3,
		colorbrewer values/4/.code={
			\definecolor{colorbrewer1}{RGB}{202,0,32}
			\definecolor{colorbrewer2}{RGB}{244,165,130}
			\definecolor{colorbrewer3}{RGB}{186,186,186}
			\definecolor{colorbrewer4}{RGB}{64,64,64}
		},
		colorbrewer values/5/.code={
			\definecolor{colorbrewer1}{RGB}{202,0,32}
			\definecolor{colorbrewer2}{RGB}{244,165,130}
			\definecolor{colorbrewer3}{RGB}{255,255,255}
			\definecolor{colorbrewer4}{RGB}{186,186,186}
			\definecolor{colorbrewer5}{RGB}{64,64,64}
		},
		colorbrewer values/6/.code={
			\definecolor{colorbrewer1}{RGB}{178,24,43}
			\definecolor{colorbrewer2}{RGB}{239,138,98}
			\definecolor{colorbrewer3}{RGB}{253,219,199}
			\definecolor{colorbrewer4}{RGB}{224,224,224}
			\definecolor{colorbrewer5}{RGB}{153,153,153}
			\definecolor{colorbrewer6}{RGB}{77,77,77}
		},
		colorbrewer values/7/.code={
			\definecolor{colorbrewer1}{RGB}{178,24,43}
			\definecolor{colorbrewer2}{RGB}{239,138,98}
			\definecolor{colorbrewer3}{RGB}{253,219,199}
			\definecolor{colorbrewer4}{RGB}{255,255,255}
			\definecolor{colorbrewer5}{RGB}{224,224,224}
			\definecolor{colorbrewer6}{RGB}{153,153,153}
			\definecolor{colorbrewer7}{RGB}{77,77,77}
		},
		colorbrewer values/8/.code={
			\definecolor{colorbrewer1}{RGB}{178,24,43}
			\definecolor{colorbrewer2}{RGB}{214,96,77}
			\definecolor{colorbrewer3}{RGB}{244,165,130}
			\definecolor{colorbrewer4}{RGB}{253,219,199}
			\definecolor{colorbrewer5}{RGB}{224,224,224}
			\definecolor{colorbrewer6}{RGB}{186,186,186}
			\definecolor{colorbrewer7}{RGB}{135,135,135}
			\definecolor{colorbrewer8}{RGB}{77,77,77}
		},
		colorbrewer values/9/.code={
			\definecolor{colorbrewer1}{RGB}{178,24,43}
			\definecolor{colorbrewer2}{RGB}{214,96,77}
			\definecolor{colorbrewer3}{RGB}{244,165,130}
			\definecolor{colorbrewer4}{RGB}{253,219,199}
			\definecolor{colorbrewer5}{RGB}{255,255,255}
			\definecolor{colorbrewer6}{RGB}{224,224,224}
			\definecolor{colorbrewer7}{RGB}{186,186,186}
			\definecolor{colorbrewer8}{RGB}{135,135,135}
			\definecolor{colorbrewer9}{RGB}{77,77,77}
		},
		colorbrewer values/10/.code={
			\definecolor{colorbrewer1}{RGB}{103,0,31}
			\definecolor{colorbrewer2}{RGB}{178,24,43}
			\definecolor{colorbrewer3}{RGB}{214,96,77}
			\definecolor{colorbrewer4}{RGB}{244,165,130}
			\definecolor{colorbrewer5}{RGB}{253,219,199}
			\definecolor{colorbrewer6}{RGB}{224,224,224}
			\definecolor{colorbrewer7}{RGB}{186,186,186}
			\definecolor{colorbrewer8}{RGB}{135,135,135}
			\definecolor{colorbrewer9}{RGB}{77,77,77}
			\definecolor{colorbrewer10}{RGB}{26,26,26}
		},
		colorbrewer values/11/.code={
			\definecolor{colorbrewer1}{RGB}{103,0,31}
			\definecolor{colorbrewer2}{RGB}{178,24,43}
			\definecolor{colorbrewer3}{RGB}{214,96,77}
			\definecolor{colorbrewer4}{RGB}{244,165,130}
			\definecolor{colorbrewer5}{RGB}{253,219,199}
			\definecolor{colorbrewer6}{RGB}{255,255,255}
			\definecolor{colorbrewer7}{RGB}{224,224,224}
			\definecolor{colorbrewer8}{RGB}{186,186,186}
			\definecolor{colorbrewer9}{RGB}{135,135,135}
			\definecolor{colorbrewer10}{RGB}{77,77,77}
			\definecolor{colorbrewer11}{RGB}{26,26,26}
		}
	},
	colorbrewer scheme/RdPu/.style={
		colorbrewer values/.is choice,	
		colorbrewer values/3/.code={
			\definecolor{colorbrewer1}{RGB}{253,224,221}
			\definecolor{colorbrewer2}{RGB}{250,159,181}
			\definecolor{colorbrewer3}{RGB}{197,27,138}
		},
		colorbrewer values=3,
		colorbrewer values/4/.code={
			\definecolor{colorbrewer1}{RGB}{254,235,226}
			\definecolor{colorbrewer2}{RGB}{251,180,185}
			\definecolor{colorbrewer3}{RGB}{247,104,161}
			\definecolor{colorbrewer4}{RGB}{174,1,126}
		},
		colorbrewer values/5/.code={
			\definecolor{colorbrewer1}{RGB}{254,235,226}
			\definecolor{colorbrewer2}{RGB}{251,180,185}
			\definecolor{colorbrewer3}{RGB}{247,104,161}
			\definecolor{colorbrewer4}{RGB}{197,27,138}
			\definecolor{colorbrewer5}{RGB}{122,1,119}
		},
		colorbrewer values/6/.code={
			\definecolor{colorbrewer1}{RGB}{254,235,226}
			\definecolor{colorbrewer2}{RGB}{252,197,192}
			\definecolor{colorbrewer3}{RGB}{250,159,181}
			\definecolor{colorbrewer4}{RGB}{247,104,161}
			\definecolor{colorbrewer5}{RGB}{197,27,138}
			\definecolor{colorbrewer6}{RGB}{122,1,119}
		},
		colorbrewer values/7/.code={
			\definecolor{colorbrewer1}{RGB}{254,235,226}
			\definecolor{colorbrewer2}{RGB}{252,197,192}
			\definecolor{colorbrewer3}{RGB}{250,159,181}
			\definecolor{colorbrewer4}{RGB}{247,104,161}
			\definecolor{colorbrewer5}{RGB}{221,52,151}
			\definecolor{colorbrewer6}{RGB}{174,1,126}
			\definecolor{colorbrewer7}{RGB}{122,1,119}
		},
		colorbrewer values/8/.code={
			\definecolor{colorbrewer1}{RGB}{255,247,243}
			\definecolor{colorbrewer2}{RGB}{253,224,221}
			\definecolor{colorbrewer3}{RGB}{252,197,192}
			\definecolor{colorbrewer4}{RGB}{250,159,181}
			\definecolor{colorbrewer5}{RGB}{247,104,161}
			\definecolor{colorbrewer6}{RGB}{221,52,151}
			\definecolor{colorbrewer7}{RGB}{174,1,126}
			\definecolor{colorbrewer8}{RGB}{122,1,119}
		},
		colorbrewer values/9/.code={
			\definecolor{colorbrewer1}{RGB}{255,247,243}
			\definecolor{colorbrewer2}{RGB}{253,224,221}
			\definecolor{colorbrewer3}{RGB}{252,197,192}
			\definecolor{colorbrewer4}{RGB}{250,159,181}
			\definecolor{colorbrewer5}{RGB}{247,104,161}
			\definecolor{colorbrewer6}{RGB}{221,52,151}
			\definecolor{colorbrewer7}{RGB}{174,1,126}
			\definecolor{colorbrewer8}{RGB}{122,1,119}
			\definecolor{colorbrewer9}{RGB}{73,0,106}
		}
	},
	colorbrewer scheme/Reds/.style={
		colorbrewer values/.is choice,	
		colorbrewer values/3/.code={
			\definecolor{colorbrewer1}{RGB}{254,224,210}
			\definecolor{colorbrewer2}{RGB}{252,146,114}
			\definecolor{colorbrewer3}{RGB}{222,45,38}
		},
		colorbrewer values=3,
		colorbrewer values/4/.code={
			\definecolor{colorbrewer1}{RGB}{254,229,217}
			\definecolor{colorbrewer2}{RGB}{252,174,145}
			\definecolor{colorbrewer3}{RGB}{251,106,74}
			\definecolor{colorbrewer4}{RGB}{203,24,29}
		},
		colorbrewer values/5/.code={
			\definecolor{colorbrewer1}{RGB}{254,229,217}
			\definecolor{colorbrewer2}{RGB}{252,174,145}
			\definecolor{colorbrewer3}{RGB}{251,106,74}
			\definecolor{colorbrewer4}{RGB}{222,45,38}
			\definecolor{colorbrewer5}{RGB}{165,15,21}
		},
		colorbrewer values/6/.code={
			\definecolor{colorbrewer1}{RGB}{254,229,217}
			\definecolor{colorbrewer2}{RGB}{252,187,161}
			\definecolor{colorbrewer3}{RGB}{252,146,114}
			\definecolor{colorbrewer4}{RGB}{251,106,74}
			\definecolor{colorbrewer5}{RGB}{222,45,38}
			\definecolor{colorbrewer6}{RGB}{165,15,21}
		},
		colorbrewer values/7/.code={
			\definecolor{colorbrewer1}{RGB}{254,229,217}
			\definecolor{colorbrewer2}{RGB}{252,187,161}
			\definecolor{colorbrewer3}{RGB}{252,146,114}
			\definecolor{colorbrewer4}{RGB}{251,106,74}
			\definecolor{colorbrewer5}{RGB}{239,59,44}
			\definecolor{colorbrewer6}{RGB}{203,24,29}
			\definecolor{colorbrewer7}{RGB}{153,0,13}
		},
		colorbrewer values/8/.code={
			\definecolor{colorbrewer1}{RGB}{255,245,240}
			\definecolor{colorbrewer2}{RGB}{254,224,210}
			\definecolor{colorbrewer3}{RGB}{252,187,161}
			\definecolor{colorbrewer4}{RGB}{252,146,114}
			\definecolor{colorbrewer5}{RGB}{251,106,74}
			\definecolor{colorbrewer6}{RGB}{239,59,44}
			\definecolor{colorbrewer7}{RGB}{203,24,29}
			\definecolor{colorbrewer8}{RGB}{153,0,13}
		},
		colorbrewer values/9/.code={
			\definecolor{colorbrewer1}{RGB}{255,245,240}
			\definecolor{colorbrewer2}{RGB}{254,224,210}
			\definecolor{colorbrewer3}{RGB}{252,187,161}
			\definecolor{colorbrewer4}{RGB}{252,146,114}
			\definecolor{colorbrewer5}{RGB}{251,106,74}
			\definecolor{colorbrewer6}{RGB}{239,59,44}
			\definecolor{colorbrewer7}{RGB}{203,24,29}
			\definecolor{colorbrewer8}{RGB}{165,15,21}
			\definecolor{colorbrewer9}{RGB}{103,0,13}
		}
	},
	colorbrewer scheme/RdYlBu/.style={
		colorbrewer values/.is choice,	
		colorbrewer values/3/.code={
			\definecolor{colorbrewer1}{RGB}{252,141,89}
			\definecolor{colorbrewer2}{RGB}{255,255,191}
			\definecolor{colorbrewer3}{RGB}{145,191,219}
		},
		colorbrewer values=3,
		colorbrewer values/4/.code={
			\definecolor{colorbrewer1}{RGB}{215,25,28}
			\definecolor{colorbrewer2}{RGB}{253,174,97}
			\definecolor{colorbrewer3}{RGB}{171,217,233}
			\definecolor{colorbrewer4}{RGB}{44,123,182}
		},
		colorbrewer values/5/.code={
			\definecolor{colorbrewer1}{RGB}{215,25,28}
			\definecolor{colorbrewer2}{RGB}{253,174,97}
			\definecolor{colorbrewer3}{RGB}{255,255,191}
			\definecolor{colorbrewer4}{RGB}{171,217,233}
			\definecolor{colorbrewer5}{RGB}{44,123,182}
		},
		colorbrewer values/6/.code={
			\definecolor{colorbrewer1}{RGB}{215,48,39}
			\definecolor{colorbrewer2}{RGB}{252,141,89}
			\definecolor{colorbrewer3}{RGB}{254,224,144}
			\definecolor{colorbrewer4}{RGB}{224,243,248}
			\definecolor{colorbrewer5}{RGB}{145,191,219}
			\definecolor{colorbrewer6}{RGB}{69,117,180}
		},
		colorbrewer values/7/.code={
			\definecolor{colorbrewer1}{RGB}{215,48,39}
			\definecolor{colorbrewer2}{RGB}{252,141,89}
			\definecolor{colorbrewer3}{RGB}{254,224,144}
			\definecolor{colorbrewer4}{RGB}{255,255,191}
			\definecolor{colorbrewer5}{RGB}{224,243,248}
			\definecolor{colorbrewer6}{RGB}{145,191,219}
			\definecolor{colorbrewer7}{RGB}{69,117,180}
		},
		colorbrewer values/8/.code={
			\definecolor{colorbrewer1}{RGB}{215,48,39}
			\definecolor{colorbrewer2}{RGB}{244,109,67}
			\definecolor{colorbrewer3}{RGB}{253,174,97}
			\definecolor{colorbrewer4}{RGB}{254,224,144}
			\definecolor{colorbrewer5}{RGB}{224,243,248}
			\definecolor{colorbrewer6}{RGB}{171,217,233}
			\definecolor{colorbrewer7}{RGB}{116,173,209}
			\definecolor{colorbrewer8}{RGB}{69,117,180}
		},
		colorbrewer values/9/.code={
			\definecolor{colorbrewer1}{RGB}{215,48,39}
			\definecolor{colorbrewer2}{RGB}{244,109,67}
			\definecolor{colorbrewer3}{RGB}{253,174,97}
			\definecolor{colorbrewer4}{RGB}{254,224,144}
			\definecolor{colorbrewer5}{RGB}{255,255,191}
			\definecolor{colorbrewer6}{RGB}{224,243,248}
			\definecolor{colorbrewer7}{RGB}{171,217,233}
			\definecolor{colorbrewer8}{RGB}{116,173,209}
			\definecolor{colorbrewer9}{RGB}{69,117,180}
		},
		colorbrewer values/10/.code={
			\definecolor{colorbrewer1}{RGB}{165,0,38}
			\definecolor{colorbrewer2}{RGB}{215,48,39}
			\definecolor{colorbrewer3}{RGB}{244,109,67}
			\definecolor{colorbrewer4}{RGB}{253,174,97}
			\definecolor{colorbrewer5}{RGB}{254,224,144}
			\definecolor{colorbrewer6}{RGB}{224,243,248}
			\definecolor{colorbrewer7}{RGB}{171,217,233}
			\definecolor{colorbrewer8}{RGB}{116,173,209}
			\definecolor{colorbrewer9}{RGB}{69,117,180}
			\definecolor{colorbrewer10}{RGB}{49,54,149}
		},
		colorbrewer values/11/.code={
			\definecolor{colorbrewer1}{RGB}{165,0,38}
			\definecolor{colorbrewer2}{RGB}{215,48,39}
			\definecolor{colorbrewer3}{RGB}{244,109,67}
			\definecolor{colorbrewer4}{RGB}{253,174,97}
			\definecolor{colorbrewer5}{RGB}{254,224,144}
			\definecolor{colorbrewer6}{RGB}{255,255,191}
			\definecolor{colorbrewer7}{RGB}{224,243,248}
			\definecolor{colorbrewer8}{RGB}{171,217,233}
			\definecolor{colorbrewer9}{RGB}{116,173,209}
			\definecolor{colorbrewer10}{RGB}{69,117,180}
			\definecolor{colorbrewer11}{RGB}{49,54,149}
		}
	},
	colorbrewer scheme/RdYlGn/.style={
		colorbrewer values/.is choice,	
		colorbrewer values/3/.code={
			\definecolor{colorbrewer1}{RGB}{252,141,89}
			\definecolor{colorbrewer2}{RGB}{255,255,191}
			\definecolor{colorbrewer3}{RGB}{145,207,96}
		},
		colorbrewer values=3,
		colorbrewer values/4/.code={
			\definecolor{colorbrewer1}{RGB}{215,25,28}
			\definecolor{colorbrewer2}{RGB}{253,174,97}
			\definecolor{colorbrewer3}{RGB}{166,217,106}
			\definecolor{colorbrewer4}{RGB}{26,150,65}
		},
		colorbrewer values/5/.code={
			\definecolor{colorbrewer1}{RGB}{215,25,28}
			\definecolor{colorbrewer2}{RGB}{253,174,97}
			\definecolor{colorbrewer3}{RGB}{255,255,191}
			\definecolor{colorbrewer4}{RGB}{166,217,106}
			\definecolor{colorbrewer5}{RGB}{26,150,65}
		},
		colorbrewer values/6/.code={
			\definecolor{colorbrewer1}{RGB}{215,48,39}
			\definecolor{colorbrewer2}{RGB}{252,141,89}
			\definecolor{colorbrewer3}{RGB}{254,224,139}
			\definecolor{colorbrewer4}{RGB}{217,239,139}
			\definecolor{colorbrewer5}{RGB}{145,207,96}
			\definecolor{colorbrewer6}{RGB}{26,152,80}
		},
		colorbrewer values/7/.code={
			\definecolor{colorbrewer1}{RGB}{215,48,39}
			\definecolor{colorbrewer2}{RGB}{252,141,89}
			\definecolor{colorbrewer3}{RGB}{254,224,139}
			\definecolor{colorbrewer4}{RGB}{255,255,191}
			\definecolor{colorbrewer5}{RGB}{217,239,139}
			\definecolor{colorbrewer6}{RGB}{145,207,96}
			\definecolor{colorbrewer7}{RGB}{26,152,80}
		},
		colorbrewer values/8/.code={
			\definecolor{colorbrewer1}{RGB}{215,48,39}
			\definecolor{colorbrewer2}{RGB}{244,109,67}
			\definecolor{colorbrewer3}{RGB}{253,174,97}
			\definecolor{colorbrewer4}{RGB}{254,224,139}
			\definecolor{colorbrewer5}{RGB}{217,239,139}
			\definecolor{colorbrewer6}{RGB}{166,217,106}
			\definecolor{colorbrewer7}{RGB}{102,189,99}
			\definecolor{colorbrewer8}{RGB}{26,152,80}
		},
		colorbrewer values/9/.code={
			\definecolor{colorbrewer1}{RGB}{215,48,39}
			\definecolor{colorbrewer2}{RGB}{244,109,67}
			\definecolor{colorbrewer3}{RGB}{253,174,97}
			\definecolor{colorbrewer4}{RGB}{254,224,139}
			\definecolor{colorbrewer5}{RGB}{255,255,191}
			\definecolor{colorbrewer6}{RGB}{217,239,139}
			\definecolor{colorbrewer7}{RGB}{166,217,106}
			\definecolor{colorbrewer8}{RGB}{102,189,99}
			\definecolor{colorbrewer9}{RGB}{26,152,80}
		},
		colorbrewer values/10/.code={
			\definecolor{colorbrewer1}{RGB}{165,0,38}
			\definecolor{colorbrewer2}{RGB}{215,48,39}
			\definecolor{colorbrewer3}{RGB}{244,109,67}
			\definecolor{colorbrewer4}{RGB}{253,174,97}
			\definecolor{colorbrewer5}{RGB}{254,224,139}
			\definecolor{colorbrewer6}{RGB}{217,239,139}
			\definecolor{colorbrewer7}{RGB}{166,217,106}
			\definecolor{colorbrewer8}{RGB}{102,189,99}
			\definecolor{colorbrewer9}{RGB}{26,152,80}
			\definecolor{colorbrewer10}{RGB}{0,104,55}
		},
		colorbrewer values/11/.code={
			\definecolor{colorbrewer1}{RGB}{165,0,38}
			\definecolor{colorbrewer2}{RGB}{215,48,39}
			\definecolor{colorbrewer3}{RGB}{244,109,67}
			\definecolor{colorbrewer4}{RGB}{253,174,97}
			\definecolor{colorbrewer5}{RGB}{254,224,139}
			\definecolor{colorbrewer6}{RGB}{255,255,191}
			\definecolor{colorbrewer7}{RGB}{217,239,139}
			\definecolor{colorbrewer8}{RGB}{166,217,106}
			\definecolor{colorbrewer9}{RGB}{102,189,99}
			\definecolor{colorbrewer10}{RGB}{26,152,80}
			\definecolor{colorbrewer11}{RGB}{0,104,55}
		}
	},
	colorbrewer scheme/Set1/.style={
		colorbrewer values/.is choice,	
		colorbrewer values/3/.code={
			\definecolor{colorbrewer1}{RGB}{228,26,28}
			\definecolor{colorbrewer2}{RGB}{55,126,184}
			\definecolor{colorbrewer3}{RGB}{77,175,74}
		},
		colorbrewer values=3,
		colorbrewer values/4/.code={
			\definecolor{colorbrewer1}{RGB}{228,26,28}
			\definecolor{colorbrewer2}{RGB}{55,126,184}
			\definecolor{colorbrewer3}{RGB}{77,175,74}
			\definecolor{colorbrewer4}{RGB}{152,78,163}
		},
		colorbrewer values/5/.code={
			\definecolor{colorbrewer1}{RGB}{228,26,28}
			\definecolor{colorbrewer2}{RGB}{55,126,184}
			\definecolor{colorbrewer3}{RGB}{77,175,74}
			\definecolor{colorbrewer4}{RGB}{152,78,163}
			\definecolor{colorbrewer5}{RGB}{255,127,0}
		},
		colorbrewer values/6/.code={
			\definecolor{colorbrewer1}{RGB}{228,26,28}
			\definecolor{colorbrewer2}{RGB}{55,126,184}
			\definecolor{colorbrewer3}{RGB}{77,175,74}
			\definecolor{colorbrewer4}{RGB}{152,78,163}
			\definecolor{colorbrewer5}{RGB}{255,127,0}
			\definecolor{colorbrewer6}{RGB}{255,255,51}
		},
		colorbrewer values/7/.code={
			\definecolor{colorbrewer1}{RGB}{228,26,28}
			\definecolor{colorbrewer2}{RGB}{55,126,184}
			\definecolor{colorbrewer3}{RGB}{77,175,74}
			\definecolor{colorbrewer4}{RGB}{152,78,163}
			\definecolor{colorbrewer5}{RGB}{255,127,0}
			\definecolor{colorbrewer6}{RGB}{255,255,51}
			\definecolor{colorbrewer7}{RGB}{166,86,40}
		},
		colorbrewer values/8/.code={
			\definecolor{colorbrewer1}{RGB}{228,26,28}
			\definecolor{colorbrewer2}{RGB}{55,126,184}
			\definecolor{colorbrewer3}{RGB}{77,175,74}
			\definecolor{colorbrewer4}{RGB}{152,78,163}
			\definecolor{colorbrewer5}{RGB}{255,127,0}
			\definecolor{colorbrewer6}{RGB}{255,255,51}
			\definecolor{colorbrewer7}{RGB}{166,86,40}
			\definecolor{colorbrewer8}{RGB}{247,129,191}
		},
		colorbrewer values/9/.code={
			\definecolor{colorbrewer1}{RGB}{228,26,28}
			\definecolor{colorbrewer2}{RGB}{55,126,184}
			\definecolor{colorbrewer3}{RGB}{77,175,74}
			\definecolor{colorbrewer4}{RGB}{152,78,163}
			\definecolor{colorbrewer5}{RGB}{255,127,0}
			\definecolor{colorbrewer6}{RGB}{255,255,51}
			\definecolor{colorbrewer7}{RGB}{166,86,40}
			\definecolor{colorbrewer8}{RGB}{247,129,191}
			\definecolor{colorbrewer9}{RGB}{153,153,153}
		}
	},
	colorbrewer scheme/Set2/.style={
		colorbrewer values/.is choice,	
		colorbrewer values/3/.code={
			\definecolor{colorbrewer1}{RGB}{102,194,165}
			\definecolor{colorbrewer2}{RGB}{252,141,98}
			\definecolor{colorbrewer3}{RGB}{141,160,203}
		},
		colorbrewer values=3,
		colorbrewer values/4/.code={
			\definecolor{colorbrewer1}{RGB}{102,194,165}
			\definecolor{colorbrewer2}{RGB}{252,141,98}
			\definecolor{colorbrewer3}{RGB}{141,160,203}
			\definecolor{colorbrewer4}{RGB}{231,138,195}
		},
		colorbrewer values/5/.code={
			\definecolor{colorbrewer1}{RGB}{102,194,165}
			\definecolor{colorbrewer2}{RGB}{252,141,98}
			\definecolor{colorbrewer3}{RGB}{141,160,203}
			\definecolor{colorbrewer4}{RGB}{231,138,195}
			\definecolor{colorbrewer5}{RGB}{166,216,84}
		},
		colorbrewer values/6/.code={
			\definecolor{colorbrewer1}{RGB}{102,194,165}
			\definecolor{colorbrewer2}{RGB}{252,141,98}
			\definecolor{colorbrewer3}{RGB}{141,160,203}
			\definecolor{colorbrewer4}{RGB}{231,138,195}
			\definecolor{colorbrewer5}{RGB}{166,216,84}
			\definecolor{colorbrewer6}{RGB}{255,217,47}
		},
		colorbrewer values/7/.code={
			\definecolor{colorbrewer1}{RGB}{102,194,165}
			\definecolor{colorbrewer2}{RGB}{252,141,98}
			\definecolor{colorbrewer3}{RGB}{141,160,203}
			\definecolor{colorbrewer4}{RGB}{231,138,195}
			\definecolor{colorbrewer5}{RGB}{166,216,84}
			\definecolor{colorbrewer6}{RGB}{255,217,47}
			\definecolor{colorbrewer7}{RGB}{229,196,148}
		},
		colorbrewer values/8/.code={
			\definecolor{colorbrewer1}{RGB}{102,194,165}
			\definecolor{colorbrewer2}{RGB}{252,141,98}
			\definecolor{colorbrewer3}{RGB}{141,160,203}
			\definecolor{colorbrewer4}{RGB}{231,138,195}
			\definecolor{colorbrewer5}{RGB}{166,216,84}
			\definecolor{colorbrewer6}{RGB}{255,217,47}
			\definecolor{colorbrewer7}{RGB}{229,196,148}
			\definecolor{colorbrewer8}{RGB}{179,179,179}
		}
	},
	colorbrewer scheme/Set3/.style={
		colorbrewer values/.is choice,	
		colorbrewer values/3/.code={
			\definecolor{colorbrewer1}{RGB}{141,211,199}
			\definecolor{colorbrewer2}{RGB}{255,255,179}
			\definecolor{colorbrewer3}{RGB}{190,186,218}
		},
		colorbrewer values=3,
		colorbrewer values/4/.code={
			\definecolor{colorbrewer1}{RGB}{141,211,199}
			\definecolor{colorbrewer2}{RGB}{255,255,179}
			\definecolor{colorbrewer3}{RGB}{190,186,218}
			\definecolor{colorbrewer4}{RGB}{251,128,114}
		},
		colorbrewer values/5/.code={
			\definecolor{colorbrewer1}{RGB}{141,211,199}
			\definecolor{colorbrewer2}{RGB}{255,255,179}
			\definecolor{colorbrewer3}{RGB}{190,186,218}
			\definecolor{colorbrewer4}{RGB}{251,128,114}
			\definecolor{colorbrewer5}{RGB}{128,177,211}
		},
		colorbrewer values/6/.code={
			\definecolor{colorbrewer1}{RGB}{141,211,199}
			\definecolor{colorbrewer2}{RGB}{255,255,179}
			\definecolor{colorbrewer3}{RGB}{190,186,218}
			\definecolor{colorbrewer4}{RGB}{251,128,114}
			\definecolor{colorbrewer5}{RGB}{128,177,211}
			\definecolor{colorbrewer6}{RGB}{253,180,98}
		},
		colorbrewer values/7/.code={
			\definecolor{colorbrewer1}{RGB}{141,211,199}
			\definecolor{colorbrewer2}{RGB}{255,255,179}
			\definecolor{colorbrewer3}{RGB}{190,186,218}
			\definecolor{colorbrewer4}{RGB}{251,128,114}
			\definecolor{colorbrewer5}{RGB}{128,177,211}
			\definecolor{colorbrewer6}{RGB}{253,180,98}
			\definecolor{colorbrewer7}{RGB}{179,222,105}
		},
		colorbrewer values/8/.code={
			\definecolor{colorbrewer1}{RGB}{141,211,199}
			\definecolor{colorbrewer2}{RGB}{255,255,179}
			\definecolor{colorbrewer3}{RGB}{190,186,218}
			\definecolor{colorbrewer4}{RGB}{251,128,114}
			\definecolor{colorbrewer5}{RGB}{128,177,211}
			\definecolor{colorbrewer6}{RGB}{253,180,98}
			\definecolor{colorbrewer7}{RGB}{179,222,105}
			\definecolor{colorbrewer8}{RGB}{252,205,229}
		},
		colorbrewer values/9/.code={
			\definecolor{colorbrewer1}{RGB}{141,211,199}
			\definecolor{colorbrewer2}{RGB}{255,255,179}
			\definecolor{colorbrewer3}{RGB}{190,186,218}
			\definecolor{colorbrewer4}{RGB}{251,128,114}
			\definecolor{colorbrewer5}{RGB}{128,177,211}
			\definecolor{colorbrewer6}{RGB}{253,180,98}
			\definecolor{colorbrewer7}{RGB}{179,222,105}
			\definecolor{colorbrewer8}{RGB}{252,205,229}
			\definecolor{colorbrewer9}{RGB}{217,217,217}
		},
		colorbrewer values/10/.code={
			\definecolor{colorbrewer1}{RGB}{141,211,199}
			\definecolor{colorbrewer2}{RGB}{255,255,179}
			\definecolor{colorbrewer3}{RGB}{190,186,218}
			\definecolor{colorbrewer4}{RGB}{251,128,114}
			\definecolor{colorbrewer5}{RGB}{128,177,211}
			\definecolor{colorbrewer6}{RGB}{253,180,98}
			\definecolor{colorbrewer7}{RGB}{179,222,105}
			\definecolor{colorbrewer8}{RGB}{252,205,229}
			\definecolor{colorbrewer9}{RGB}{217,217,217}
			\definecolor{colorbrewer10}{RGB}{188,128,189}
		},
		colorbrewer values/11/.code={
			\definecolor{colorbrewer1}{RGB}{141,211,199}
			\definecolor{colorbrewer2}{RGB}{255,255,179}
			\definecolor{colorbrewer3}{RGB}{190,186,218}
			\definecolor{colorbrewer4}{RGB}{251,128,114}
			\definecolor{colorbrewer5}{RGB}{128,177,211}
			\definecolor{colorbrewer6}{RGB}{253,180,98}
			\definecolor{colorbrewer7}{RGB}{179,222,105}
			\definecolor{colorbrewer8}{RGB}{252,205,229}
			\definecolor{colorbrewer9}{RGB}{217,217,217}
			\definecolor{colorbrewer10}{RGB}{188,128,189}
			\definecolor{colorbrewer11}{RGB}{204,235,197}
		},
		colorbrewer values/12/.code={
			\definecolor{colorbrewer1}{RGB}{141,211,199}
			\definecolor{colorbrewer2}{RGB}{255,255,179}
			\definecolor{colorbrewer3}{RGB}{190,186,218}
			\definecolor{colorbrewer4}{RGB}{251,128,114}
			\definecolor{colorbrewer5}{RGB}{128,177,211}
			\definecolor{colorbrewer6}{RGB}{253,180,98}
			\definecolor{colorbrewer7}{RGB}{179,222,105}
			\definecolor{colorbrewer8}{RGB}{252,205,229}
			\definecolor{colorbrewer9}{RGB}{217,217,217}
			\definecolor{colorbrewer10}{RGB}{188,128,189}
			\definecolor{colorbrewer11}{RGB}{204,235,197}
			\definecolor{colorbrewer12}{RGB}{255,237,111}
		}
	},
	colorbrewer scheme/Spectral/.style={
		colorbrewer values/.is choice,	
		colorbrewer values/3/.code={
			\definecolor{colorbrewer1}{RGB}{252,141,89}
			\definecolor{colorbrewer2}{RGB}{255,255,191}
			\definecolor{colorbrewer3}{RGB}{153,213,148}
		},
		colorbrewer values=3,
		colorbrewer values/4/.code={
			\definecolor{colorbrewer1}{RGB}{215,25,28}
			\definecolor{colorbrewer2}{RGB}{253,174,97}
			\definecolor{colorbrewer3}{RGB}{171,221,164}
			\definecolor{colorbrewer4}{RGB}{43,131,186}
		},
		colorbrewer values/5/.code={
			\definecolor{colorbrewer1}{RGB}{215,25,28}
			\definecolor{colorbrewer2}{RGB}{253,174,97}
			\definecolor{colorbrewer3}{RGB}{255,255,191}
			\definecolor{colorbrewer4}{RGB}{171,221,164}
			\definecolor{colorbrewer5}{RGB}{43,131,186}
		},
		colorbrewer values/6/.code={
			\definecolor{colorbrewer1}{RGB}{213,62,79}
			\definecolor{colorbrewer2}{RGB}{252,141,89}
			\definecolor{colorbrewer3}{RGB}{254,224,139}
			\definecolor{colorbrewer4}{RGB}{230,245,152}
			\definecolor{colorbrewer5}{RGB}{153,213,148}
			\definecolor{colorbrewer6}{RGB}{50,136,189}
		},
		colorbrewer values/7/.code={
			\definecolor{colorbrewer1}{RGB}{213,62,79}
			\definecolor{colorbrewer2}{RGB}{252,141,89}
			\definecolor{colorbrewer3}{RGB}{254,224,139}
			\definecolor{colorbrewer4}{RGB}{255,255,191}
			\definecolor{colorbrewer5}{RGB}{230,245,152}
			\definecolor{colorbrewer6}{RGB}{153,213,148}
			\definecolor{colorbrewer7}{RGB}{50,136,189}
		},
		colorbrewer values/8/.code={
			\definecolor{colorbrewer1}{RGB}{213,62,79}
			\definecolor{colorbrewer2}{RGB}{244,109,67}
			\definecolor{colorbrewer3}{RGB}{253,174,97}
			\definecolor{colorbrewer4}{RGB}{254,224,139}
			\definecolor{colorbrewer5}{RGB}{230,245,152}
			\definecolor{colorbrewer6}{RGB}{171,221,164}
			\definecolor{colorbrewer7}{RGB}{102,194,165}
			\definecolor{colorbrewer8}{RGB}{50,136,189}
		},
		colorbrewer values/9/.code={
			\definecolor{colorbrewer1}{RGB}{213,62,79}
			\definecolor{colorbrewer2}{RGB}{244,109,67}
			\definecolor{colorbrewer3}{RGB}{253,174,97}
			\definecolor{colorbrewer4}{RGB}{254,224,139}
			\definecolor{colorbrewer5}{RGB}{255,255,191}
			\definecolor{colorbrewer6}{RGB}{230,245,152}
			\definecolor{colorbrewer7}{RGB}{171,221,164}
			\definecolor{colorbrewer8}{RGB}{102,194,165}
			\definecolor{colorbrewer9}{RGB}{50,136,189}
		},
		colorbrewer values/10/.code={
			\definecolor{colorbrewer1}{RGB}{158,1,66}
			\definecolor{colorbrewer2}{RGB}{213,62,79}
			\definecolor{colorbrewer3}{RGB}{244,109,67}
			\definecolor{colorbrewer4}{RGB}{253,174,97}
			\definecolor{colorbrewer5}{RGB}{254,224,139}
			\definecolor{colorbrewer6}{RGB}{230,245,152}
			\definecolor{colorbrewer7}{RGB}{171,221,164}
			\definecolor{colorbrewer8}{RGB}{102,194,165}
			\definecolor{colorbrewer9}{RGB}{50,136,189}
			\definecolor{colorbrewer10}{RGB}{94,79,162}
		},
		colorbrewer values/11/.code={
			\definecolor{colorbrewer1}{RGB}{158,1,66}
			\definecolor{colorbrewer2}{RGB}{213,62,79}
			\definecolor{colorbrewer3}{RGB}{244,109,67}
			\definecolor{colorbrewer4}{RGB}{253,174,97}
			\definecolor{colorbrewer5}{RGB}{254,224,139}
			\definecolor{colorbrewer6}{RGB}{255,255,191}
			\definecolor{colorbrewer7}{RGB}{230,245,152}
			\definecolor{colorbrewer8}{RGB}{171,221,164}
			\definecolor{colorbrewer9}{RGB}{102,194,165}
			\definecolor{colorbrewer10}{RGB}{50,136,189}
			\definecolor{colorbrewer11}{RGB}{94,79,162}
		}
	},
	colorbrewer scheme/YlGn/.style={
		colorbrewer values/.is choice,	
		colorbrewer values/3/.code={
			\definecolor{colorbrewer1}{RGB}{247,252,185}
			\definecolor{colorbrewer2}{RGB}{173,221,142}
			\definecolor{colorbrewer3}{RGB}{49,163,84}
		},
		colorbrewer values=3,
		colorbrewer values/4/.code={
			\definecolor{colorbrewer1}{RGB}{255,255,204}
			\definecolor{colorbrewer2}{RGB}{194,230,153}
			\definecolor{colorbrewer3}{RGB}{120,198,121}
			\definecolor{colorbrewer4}{RGB}{35,132,67}
		},
		colorbrewer values/5/.code={
			\definecolor{colorbrewer1}{RGB}{255,255,204}
			\definecolor{colorbrewer2}{RGB}{194,230,153}
			\definecolor{colorbrewer3}{RGB}{120,198,121}
			\definecolor{colorbrewer4}{RGB}{49,163,84}
			\definecolor{colorbrewer5}{RGB}{0,104,55}
		},
		colorbrewer values/6/.code={
			\definecolor{colorbrewer1}{RGB}{255,255,204}
			\definecolor{colorbrewer2}{RGB}{217,240,163}
			\definecolor{colorbrewer3}{RGB}{173,221,142}
			\definecolor{colorbrewer4}{RGB}{120,198,121}
			\definecolor{colorbrewer5}{RGB}{49,163,84}
			\definecolor{colorbrewer6}{RGB}{0,104,55}
		},
		colorbrewer values/7/.code={
			\definecolor{colorbrewer1}{RGB}{255,255,204}
			\definecolor{colorbrewer2}{RGB}{217,240,163}
			\definecolor{colorbrewer3}{RGB}{173,221,142}
			\definecolor{colorbrewer4}{RGB}{120,198,121}
			\definecolor{colorbrewer5}{RGB}{65,171,93}
			\definecolor{colorbrewer6}{RGB}{35,132,67}
			\definecolor{colorbrewer7}{RGB}{0,90,50}
		},
		colorbrewer values/8/.code={
			\definecolor{colorbrewer1}{RGB}{255,255,229}
			\definecolor{colorbrewer2}{RGB}{247,252,185}
			\definecolor{colorbrewer3}{RGB}{217,240,163}
			\definecolor{colorbrewer4}{RGB}{173,221,142}
			\definecolor{colorbrewer5}{RGB}{120,198,121}
			\definecolor{colorbrewer6}{RGB}{65,171,93}
			\definecolor{colorbrewer7}{RGB}{35,132,67}
			\definecolor{colorbrewer8}{RGB}{0,90,50}
		},
		colorbrewer values/9/.code={
			\definecolor{colorbrewer1}{RGB}{255,255,229}
			\definecolor{colorbrewer2}{RGB}{247,252,185}
			\definecolor{colorbrewer3}{RGB}{217,240,163}
			\definecolor{colorbrewer4}{RGB}{173,221,142}
			\definecolor{colorbrewer5}{RGB}{120,198,121}
			\definecolor{colorbrewer6}{RGB}{65,171,93}
			\definecolor{colorbrewer7}{RGB}{35,132,67}
			\definecolor{colorbrewer8}{RGB}{0,104,55}
			\definecolor{colorbrewer9}{RGB}{0,69,41}
		}
	},
	colorbrewer scheme/YlGnBu/.style={
		colorbrewer values/.is choice,	
		colorbrewer values/3/.code={
			\definecolor{colorbrewer1}{RGB}{237,248,177}
			\definecolor{colorbrewer2}{RGB}{127,205,187}
			\definecolor{colorbrewer3}{RGB}{44,127,184}
		},
		colorbrewer values=3,
		colorbrewer values/4/.code={
			\definecolor{colorbrewer1}{RGB}{255,255,204}
			\definecolor{colorbrewer2}{RGB}{161,218,180}
			\definecolor{colorbrewer3}{RGB}{65,182,196}
			\definecolor{colorbrewer4}{RGB}{34,94,168}
		},
		colorbrewer values/5/.code={
			\definecolor{colorbrewer1}{RGB}{255,255,204}
			\definecolor{colorbrewer2}{RGB}{161,218,180}
			\definecolor{colorbrewer3}{RGB}{65,182,196}
			\definecolor{colorbrewer4}{RGB}{44,127,184}
			\definecolor{colorbrewer5}{RGB}{37,52,148}
		},
		colorbrewer values/6/.code={
			\definecolor{colorbrewer1}{RGB}{255,255,204}
			\definecolor{colorbrewer2}{RGB}{199,233,180}
			\definecolor{colorbrewer3}{RGB}{127,205,187}
			\definecolor{colorbrewer4}{RGB}{65,182,196}
			\definecolor{colorbrewer5}{RGB}{44,127,184}
			\definecolor{colorbrewer6}{RGB}{37,52,148}
		},
		colorbrewer values/7/.code={
			\definecolor{colorbrewer1}{RGB}{255,255,204}
			\definecolor{colorbrewer2}{RGB}{199,233,180}
			\definecolor{colorbrewer3}{RGB}{127,205,187}
			\definecolor{colorbrewer4}{RGB}{65,182,196}
			\definecolor{colorbrewer5}{RGB}{29,145,192}
			\definecolor{colorbrewer6}{RGB}{34,94,168}
			\definecolor{colorbrewer7}{RGB}{12,44,132}
		},
		colorbrewer values/8/.code={
			\definecolor{colorbrewer1}{RGB}{255,255,217}
			\definecolor{colorbrewer2}{RGB}{237,248,177}
			\definecolor{colorbrewer3}{RGB}{199,233,180}
			\definecolor{colorbrewer4}{RGB}{127,205,187}
			\definecolor{colorbrewer5}{RGB}{65,182,196}
			\definecolor{colorbrewer6}{RGB}{29,145,192}
			\definecolor{colorbrewer7}{RGB}{34,94,168}
			\definecolor{colorbrewer8}{RGB}{12,44,132}
		},
		colorbrewer values/9/.code={
			\definecolor{colorbrewer1}{RGB}{255,255,217}
			\definecolor{colorbrewer2}{RGB}{237,248,177}
			\definecolor{colorbrewer3}{RGB}{199,233,180}
			\definecolor{colorbrewer4}{RGB}{127,205,187}
			\definecolor{colorbrewer5}{RGB}{65,182,196}
			\definecolor{colorbrewer6}{RGB}{29,145,192}
			\definecolor{colorbrewer7}{RGB}{34,94,168}
			\definecolor{colorbrewer8}{RGB}{37,52,148}
			\definecolor{colorbrewer9}{RGB}{8,29,88}
		}
	},
	colorbrewer scheme/YlOrBr/.style={
		colorbrewer values/.is choice,	
		colorbrewer values/3/.code={
			\definecolor{colorbrewer1}{RGB}{255,247,188}
			\definecolor{colorbrewer2}{RGB}{254,196,79}
			\definecolor{colorbrewer3}{RGB}{217,95,14}
		},
		colorbrewer values=3,
		colorbrewer values/4/.code={
			\definecolor{colorbrewer1}{RGB}{255,255,212}
			\definecolor{colorbrewer2}{RGB}{254,217,142}
			\definecolor{colorbrewer3}{RGB}{254,153,41}
			\definecolor{colorbrewer4}{RGB}{204,76,2}
		},
		colorbrewer values/5/.code={
			\definecolor{colorbrewer1}{RGB}{255,255,212}
			\definecolor{colorbrewer2}{RGB}{254,217,142}
			\definecolor{colorbrewer3}{RGB}{254,153,41}
			\definecolor{colorbrewer4}{RGB}{217,95,14}
			\definecolor{colorbrewer5}{RGB}{153,52,4}
		},
		colorbrewer values/6/.code={
			\definecolor{colorbrewer1}{RGB}{255,255,212}
			\definecolor{colorbrewer2}{RGB}{254,227,145}
			\definecolor{colorbrewer3}{RGB}{254,196,79}
			\definecolor{colorbrewer4}{RGB}{254,153,41}
			\definecolor{colorbrewer5}{RGB}{217,95,14}
			\definecolor{colorbrewer6}{RGB}{153,52,4}
		},
		colorbrewer values/7/.code={
			\definecolor{colorbrewer1}{RGB}{255,255,212}
			\definecolor{colorbrewer2}{RGB}{254,227,145}
			\definecolor{colorbrewer3}{RGB}{254,196,79}
			\definecolor{colorbrewer4}{RGB}{254,153,41}
			\definecolor{colorbrewer5}{RGB}{236,112,20}
			\definecolor{colorbrewer6}{RGB}{204,76,2}
			\definecolor{colorbrewer7}{RGB}{140,45,4}
		},
		colorbrewer values/8/.code={
			\definecolor{colorbrewer1}{RGB}{255,255,229}
			\definecolor{colorbrewer2}{RGB}{255,247,188}
			\definecolor{colorbrewer3}{RGB}{254,227,145}
			\definecolor{colorbrewer4}{RGB}{254,196,79}
			\definecolor{colorbrewer5}{RGB}{254,153,41}
			\definecolor{colorbrewer6}{RGB}{236,112,20}
			\definecolor{colorbrewer7}{RGB}{204,76,2}
			\definecolor{colorbrewer8}{RGB}{140,45,4}
		},
		colorbrewer values/9/.code={
			\definecolor{colorbrewer1}{RGB}{255,255,229}
			\definecolor{colorbrewer2}{RGB}{255,247,188}
			\definecolor{colorbrewer3}{RGB}{254,227,145}
			\definecolor{colorbrewer4}{RGB}{254,196,79}
			\definecolor{colorbrewer5}{RGB}{254,153,41}
			\definecolor{colorbrewer6}{RGB}{236,112,20}
			\definecolor{colorbrewer7}{RGB}{204,76,2}
			\definecolor{colorbrewer8}{RGB}{153,52,4}
			\definecolor{colorbrewer9}{RGB}{102,37,6}
		}
	},
	colorbrewer scheme/YlOrRd/.style={
		colorbrewer values/.is choice,	
		colorbrewer values/3/.code={
			\definecolor{colorbrewer1}{RGB}{255,237,160}
			\definecolor{colorbrewer2}{RGB}{254,178,76}
			\definecolor{colorbrewer3}{RGB}{240,59,32}
		},
		colorbrewer values=3,
		colorbrewer values/4/.code={
			\definecolor{colorbrewer1}{RGB}{255,255,178}
			\definecolor{colorbrewer2}{RGB}{254,204,92}
			\definecolor{colorbrewer3}{RGB}{253,141,60}
			\definecolor{colorbrewer4}{RGB}{227,26,28}
		},
		colorbrewer values/5/.code={
			\definecolor{colorbrewer1}{RGB}{255,255,178}
			\definecolor{colorbrewer2}{RGB}{254,204,92}
			\definecolor{colorbrewer3}{RGB}{253,141,60}
			\definecolor{colorbrewer4}{RGB}{240,59,32}
			\definecolor{colorbrewer5}{RGB}{189,0,38}
		},
		colorbrewer values/6/.code={
			\definecolor{colorbrewer1}{RGB}{255,255,178}
			\definecolor{colorbrewer2}{RGB}{254,217,118}
			\definecolor{colorbrewer3}{RGB}{254,178,76}
			\definecolor{colorbrewer4}{RGB}{253,141,60}
			\definecolor{colorbrewer5}{RGB}{240,59,32}
			\definecolor{colorbrewer6}{RGB}{189,0,38}
		},
		colorbrewer values/7/.code={
			\definecolor{colorbrewer1}{RGB}{255,255,178}
			\definecolor{colorbrewer2}{RGB}{254,217,118}
			\definecolor{colorbrewer3}{RGB}{254,178,76}
			\definecolor{colorbrewer4}{RGB}{253,141,60}
			\definecolor{colorbrewer5}{RGB}{252,78,42}
			\definecolor{colorbrewer6}{RGB}{227,26,28}
			\definecolor{colorbrewer7}{RGB}{177,0,38}
		},
		colorbrewer values/8/.code={
			\definecolor{colorbrewer1}{RGB}{255,255,204}
			\definecolor{colorbrewer2}{RGB}{255,237,160}
			\definecolor{colorbrewer3}{RGB}{254,217,118}
			\definecolor{colorbrewer4}{RGB}{254,178,76}
			\definecolor{colorbrewer5}{RGB}{253,141,60}
			\definecolor{colorbrewer6}{RGB}{252,78,42}
			\definecolor{colorbrewer7}{RGB}{227,26,28}
			\definecolor{colorbrewer8}{RGB}{177,0,38}
		},
		colorbrewer values/9/.code={
			\definecolor{colorbrewer1}{RGB}{255,255,204}
			\definecolor{colorbrewer2}{RGB}{255,237,160}
			\definecolor{colorbrewer3}{RGB}{254,217,118}
			\definecolor{colorbrewer4}{RGB}{254,178,76}
			\definecolor{colorbrewer5}{RGB}{253,141,60}
			\definecolor{colorbrewer6}{RGB}{252,78,42}
			\definecolor{colorbrewer7}{RGB}{227,26,28}
			\definecolor{colorbrewer8}{RGB}{189,0,38}
			\definecolor{colorbrewer9}{RGB}{128,0,38}
		}
	}
}
\usetikzlibrary{decorations.text}

\title{Lossy Image Compression with Normalizing Flows}

\author{%
  Leonhard Helminger$^1$\quad Abdelaziz Djelouah$^2$\quad Markus Gross$^1$\quad Christopher Schroers$^2$\\
  \\
  $^1$Department of Computer Science\quad \quad \quad \quad  $^2$DisneyResearch|Studios\\
  ~~~~~ETH Zurich, Switzerland\quad \quad \quad \quad \quad \quad \quad Zurich, Switzerland\\
}

\begin{document}

\maketitle

\begin{abstract}
Deep learning based image compression has recently witnessed exciting progress
and in some cases even managed to surpass transform coding based approaches 
that have been established and refined over many decades. 
However, state-of-the-art solutions for deep image compression typically 
employ autoencoders which map the input to a lower dimensional latent space
and thus irreversibly discard information already before quantization. 
Due to that, they inherently limit the range of quality levels that can 
be covered. In contrast, traditional approaches in image compression allow 
for a larger range of quality levels. Interestingly, they employ an invertible 
transformation before performing the quantization step which explicitly discards 
information. Inspired by this, we propose a deep image compression method 
that is able to go from low bit-rates to near lossless quality 
by leveraging normalizing flows to learn a bijective mapping from 
the image space to a latent representation. In addition to this, we demonstrate 
further advantages unique to our solution, such as the ability to maintain 
constant quality results through re-encoding, even when performed multiple times. 
To the best of our knowledge, this is the first work to explore the opportunities 
for leveraging normalizing flows for lossy image compression.
\end{abstract}

% show how we can use normalizing flow for image compression
% additional properties
\section{Introduction}
% 1) Which problem do we address and why is it important.
Given the extremely large share of visual data in today's internet traffic, 
improvements in lossy image compression are likely to have an important impact.
In the recent years, several 
works~\citep{toderici2015variable,waveone2017,mentzer2018conditional, minnen2018full}
have explored the usage of deep learning for lossy image compression. 
These methods quickly started to challenge decades of work
in transform coding and they are now at the core of a number of learning 
based video compression techniques such
as~\citep{lu2018dvc,rippel2018learned,djelouah2019vidcodec}.

% 2) What other people do.
Contrary to traditional image compression 
codecs that use handcrafted features, deep learning based methods 
are able to directly minimise the rate-distortion objective to obtain optimal nonlinear encoding and decoding transforms along with
the probability models required for entropy coding.
A common approach in deep lossy image compression is to map the images 
into a lower dimensional latent space in which a probability distribution is learned to allow for entropy coding. 
For this mapping, \citet{balle2017endtoend}~proposed to leverage specialized nonlinear transformations (GDN), 
while \citet{waveone2017} use a pyramidal encoder to extract features 
of various scales. 
Following these initial efforts, related works have focused
on building better probability 
models~\cite{mentzer2019practical,balle2018variational,minnen2018full},
addressing multiple quality levels~\cite{DBLP:conf/iccv/ChoiEL19} 
or the very low bitrate regime~\cite{agustsson2019extreme}.	
%not the case
%necessarily leads to
%rate–distortion optimality (and ).
%few direction have been explored such
%as taking into account context information fmethods have
%After quantizing the encoding, they use adaptive arithmetic 
%coding and adaptive code length regularization to reduce the length of the bitstream.

%which implement a form of local gain control. Through the introduced continuous proxy 
%for the discontinuous loss function, they are able to optimize the model end-to-end. 
%The used entropy model is fully factorized. They also compare their model to optimizing a variational autoencoder.

% 3) What do we do.
%Generative models are at the core of learned image compression and a 
However, the existing compression methods can typically be interpreted as variational 
autoencoders (VAEs), where the entropy model corresponds to the prior on the latent representation.
Since autoencoders map images to a lower dimensional latent space, they impose an implicit limit on the reconstruction quality. As such, they cannot address high quality levels well. % and may even be surpassed by JPEG in this regime. 
Using a recurrent architecture as proposed by ~\cite{toderici2015variable}
to iteratively achieve multiple compression rates can partially mitigate
such problems, but this requires multiple passes and is not competitive in rate-distortion performance in general when compared to more recent learning based methods.
%but the empirical results suggest that this may not be the best solution. 

\begin{figure*}
	\centering
	\includegraphics[width=0.99\textwidth]{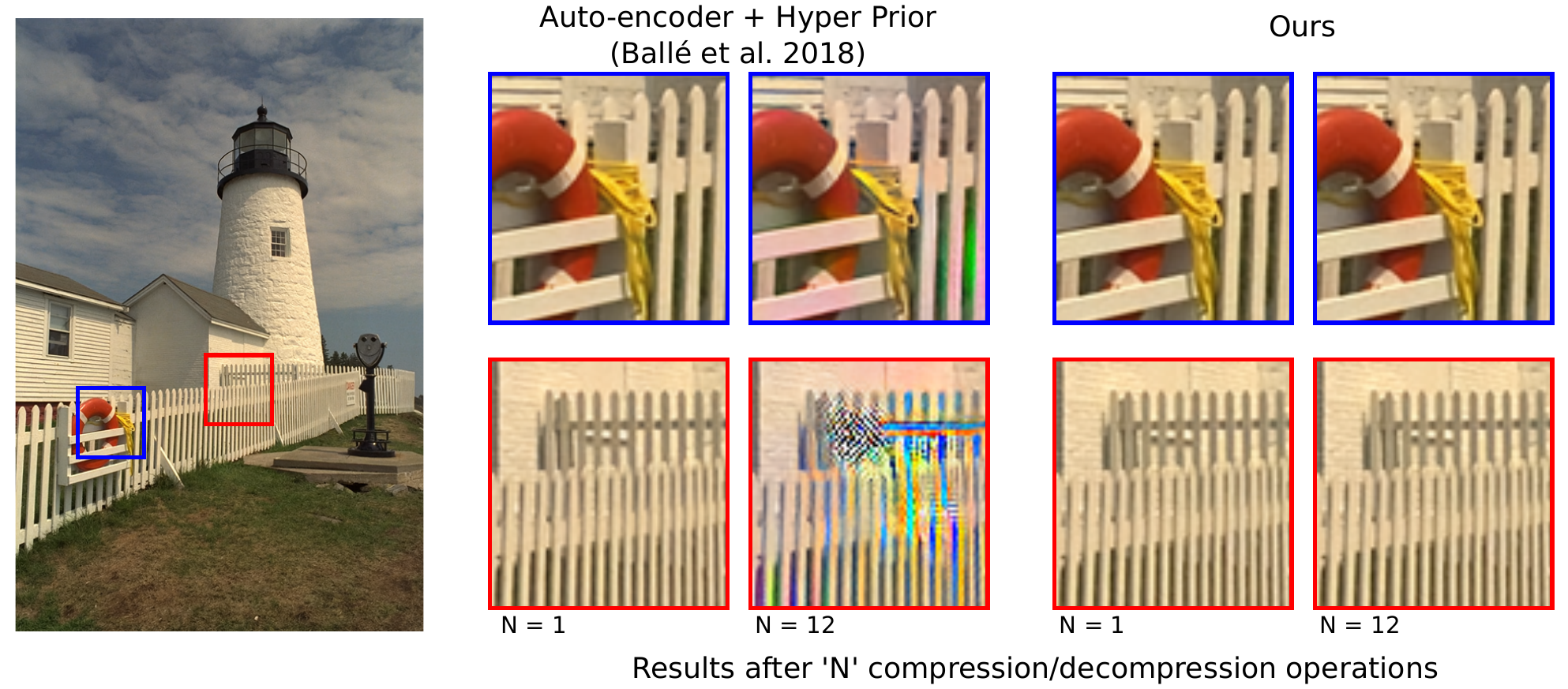} 
	\caption{\label{fig:teaser}Using normalizing flows,
		image quality and bit-rate are not affected by multiple
		compression/decompression operations. With auto-encoder based
		solutions~\citep{balle2018variational}, we notice color shifts
		and strong artifacts while the bit-rate increases.}
	\vspace{-0.5cm}
\end{figure*}

In this work, 
%normalizing flows use bijective functions to map images 
%to an alternative space (or base distribution). 
we propose to leverage normalizing flows as generative models
in lossy image compression instead of autoencoders. By learning bijective transforms from image space to latent space, we can cover a very wide range of quality levels and effectively enable to go from low bit-rates to near lossless quality. 
%This is possible thanks to the learned bijective mapping from image space 
%to latent space. 
Interestingly, this strategy is related
to traditional compression approaches that use lossless 
invertible transformations, such as the Discrete Cosine Transform (DCT),
before a quantization step that explicitly discards information. 
Another interesting property of this bijective mapping is that 
a compressed image is always mapped to the same point. Therefore consecutive compression steps can retain the same rate-distortion performance.
Figure~\ref{fig:teaser} shows an example for this where multiple compression and decompression
operations are run. The reconstruction quality
of autoencoders quickly drops and noticeable artifacts appear in the image,
while the reconstruction quality as well as the bit rate remain 
constant for our model.
This property can be particularly interesting for media production pipelines
where images are sent between multiple entities and may be compressed multiple times. Another beneficial property of our approach is progressive reconstruction, i.e. the ability to gradually increase reconstruction quality as more information is sent. 
%Our model supports this through its hierarchical design. 
Even though we do not yet outperform 
existing autoencoder based approaches over the full range of bit rates, 
we believe the results obtained in this work are very encouraging nevertheless, especially as this marks a first step with a lot of potential for future work. 
%and would push new research works to explore new bijective layers,
%architecture designs and refined training procedures. 

Summing up, the contribution of this paper is three fold: (1)~To the best of our knowledge, our work is the first to explore the potential of normalizing flows for lossy image compression. (2)~Our model achieves the widest range of quality levels among learned image compression methods and can outperform existing autoencoder solutions in the high quality regime.
(3)~We demonstrate additional advantages to our solution such 
as the capacity to maintain rate-distortion performance during multiple reencodings, as well as the ability to perform progressive reconstruction. 

Before explaining our method in detail in Chapter \ref{chap:method}, we cover the most important background on normalizing flow in Chapter \ref{chap:background}. We then set our method into context with related work (Chapter \ref{chap:related}) and show a comprehensive experimental evaluation (Chapter \ref{chap:experiments}) before finally concluding (Chapter~\ref{chap:conclusion}).

%the multi-level encoding 
%that allows a transition from low to high quality
%by only transferring the required additional data.

%\begin{itemize}
%	\item concept of AE cannot handle the full range (because of destroying information)
%	\item AE don't have guarantee of invertibility
%	\item NF: bijection leads partially to lossless compression
%	\item could bridge gap from lossy to lossless compression
%	\item artifacts removal by optimizing the latent space
%	\item being able to compress images multiple times without loss of quality with the same bit length
%\end{itemize}

% 4) Evaluation and results.

% 5) Summarize contributions. 

%------------------------------------------------
% Figure : Teaser
%------------------------------------------------
%------------------------------------------------
\section{Background on Normalizing Flow}
\label{chap:background}
In normalizing flow~\citep{DBLP:conf/icml/RezendeM15} the objective
is to map an arbitrary distribution to a base distribution,
which is done through a change of variable.
Let's consider two random variables $X$ and $Z$ that are related
through the invertible transformation $f : \mathbb{R}^d ~\rightarrow~\mathbb{R}^d$, $f\left(\mathbf{z}\right) = \mathbf{x}$.
In this case, the distributions of the two variables are related through
%--------------------------------------------------
% Equation : change of variable NF
%--------------------------------------------------
\begin{equation}
p_{X}\left(\mathbf{x}\right) = p_{Z}\left(\mathbf{z}\right) \left| \det\left( J_{f}\left(\mathbf{z}\right)\right) \right|^{-1} \;\;,
\end{equation}
where $J_{f}\left(\mathbf{z}\right)$ is the Jacobian matrix of $f$.
%-------------------------------------------------
%Here, the determinant preserves total probability
%and can be understood as the \emph{amount} of squeezing
%and stretching of the space induced by the function $f$.
In normalizing flow, a series $f_K, \ldots, f_1$ of such mappings is applied
to transform a simple probability distribution into a more complex
multi-modal distribution:
% old
%%--------------------------------------------------
%% Equation : Normalizing flow composition
%%--------------------------------------------------
%\begin{equation}
%\mathbf{x} = f_K \circ \ldots \circ f_1(\mathbf{z})
%\end{equation}
%%---------------------
%\begin{equation}
%\label{NF}
%p_{X}(\mathbf{x}) = p_{Z}(\mathbf{z}) \prod_{k=1}^{K} \left| \det\left(\frac{\partial \mathbf{h}_{k-1}}{\partial \mathbf{h}_{k}} \right) \right|^{-1}
%\end{equation}
%where we define $\mathbf{h}_k = f_k\left(\mathbf{h}_{k-1}\right)$ to be the intermediate outputs with $\mathbf{h}_k \triangleq \mathbf{x}$ and $\mathbf{h}_0 \triangleq \mathbf{z}$.
%--------------------------------------------------
% Equation : Normalizing flow composition
%--------------------------------------------------
\begin{equation}
\label{NF}
p_{X}(\mathbf{x}) = p_{Z}(\mathbf{z}) \prod_{k=1}^{K} \left| \det\left(\frac{\partial \mathbf{h}_{k-1}}{\partial \mathbf{h}_{k}} \right) \right|^{-1}  ,
\end{equation}
where $\mathbf{x} = (f_K \circ \ldots \circ f_1)(\mathbf{z})$ is the bijective relationship between $\mathbf{z}$ and $\mathbf{x}$,
%---------------------
and $\mathbf{h}_k = f_k\left(\mathbf{h}_{k-1}\right)$ 
the intermediate outputs with $\mathbf{h}_K \triangleq \mathbf{x}$ and $\mathbf{h}_0 \triangleq \mathbf{z}$.

%---------------------
%It is clear that computing the determinant of these Jacobian matrices
%as well as the function inverses must remain easy to allow their integration
%as part of a neural network.
%This is not the case for arbitrary Jacobians and
%Recent developments and successes in normalizing flow are due to the proposition
%of invertible transformations with easy to compute determinants,
%which allowed their integration as part of a neural network.
%########################################################
% Subsection : NF as generative models
%########################################################
%\subsection{Normalizing Flows as Generative model}
%\paragraph{Normalizing flows as generative model.}
Recent works~\citep{glow_KingmaD18,realNVP_DinhSB17}
have shown the great potential of using normalizing flow
as a generative model.
%As discussed by \citet{glow_KingmaD18},
We can consider an image as a high-dimensional random
vector $\mathbf{x}$ that has an unknown distribution $p\left(\mathbf{x}\right)$.
In flow based generative models the observation $\mathbf{x}$
is generated from a latent representation $\mathbf{z}$:
%--------------------------------------------------
% Equation : generative model
%--------------------------------------------------
\begin{equation}
\mathbf{x} = f_\theta\left(\mathbf{z}\right) \qquad \text{with} \quad \mathbf{z} \sim p_\theta\left(\mathbf{z}\right),
\end{equation}
%--------------------------------------------------
where $f_\theta$ is a sequence of bijective transformations
and $p_\theta\left(\mathbf{z}\right)$ a tractable distribution 
with $\theta$ being the parameters of the model.
Following equation~\ref{NF}, we can learn the parameterized distribution
$p_\theta\left(\mathbf{x}\right)$ from a discrete set of $N$ images, by minimizing
the following negative log-likelihood ($nll$) objective
%--------------------------------------------------
\begin{equation}
\label{eq:glow}
\mathcal{L}_{nll} \simeq \frac{1}{N} \sum_{i=1}^{N}
- \log p_{\theta}\left(\bar{\mathbf{x}}^{(i)}\right).
\end{equation}
%--------------------------------------------------
Here $\bar{\mathbf{x}}^{(i)} = \mathbf{x}^{(i)} + \mathbf{\xi} $ with $\mathbf{\xi} \sim \mathcal{U}\left(0, a\right)$
and $a$ is determined by the discretization level
of the data. %For brevity, we drop the subscript from $f$ and $f^{-1}$.
%Recent developments and successes in normalizing flows are
%due to the proposition of invertible transformations with
%easy to compute determinants,
%which allowed their integration as part of neural networks.
%Next we present such bijective mappings which we will use as coupling layers in our model.
Next we present the bijective mappings used in our model.
%###################################################
% Subsection : Coupling and Factor-out layers
%###################################################
%\subsection{Coupling and Factor-out Layers}

%-------------------------
% coupling layer figure
%------------------------
\begin{wrapfigure}{r}{0.4\textwidth}
	\centering
	\vspace{-0.7cm}
	\includegraphics[width=0.35\textwidth]{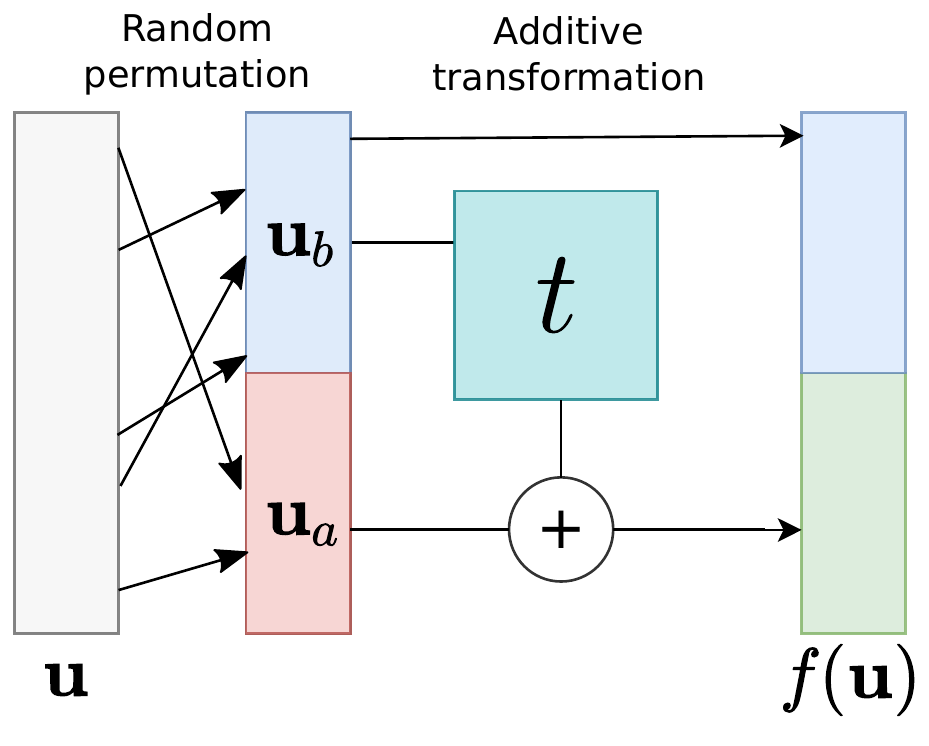}
	\caption{\label{fig:coupling} Additive coupling layer}
\end{wrapfigure}
%----------------------
\paragraph{Coupling layers}
%are a type of invertible bijective functions that
%are tractable and flexible.
%There exist several variations of coupling layers but
%The core idea is to 
split the input into two partitions, where one conditions 
a neural network to modify the remaining channels.
% an upper or lower diagonal
%This leads to a Jacobian matrix with a determinant that is easy to compute.
%%------------------------------------------------
%% Figure : Coupling layer
%%------------------------------------------------
%\begin{figure}
%	\begin{center}
%		\includegraphics[width=0.5\columnwidth]{figures/NF_level.pdf}
%		\caption{\label{fig:coupling}Coupling layer}
%	\end{center}
%\end{figure}
%%------------------------------------------------
The \emph{additive} coupling layer is the one used %\leo{should we cite these?}
as building block for our models (Figure.~\ref{fig:coupling}).
Given an input tensor $\mathbf{u}$ with partitions $\left(\mathbf{u}_a, \mathbf{u}_b\right)$,
the additive layer is defined as the following mapping
%--------------------------------------------------
% Equation : Additive Coupling layer
%--------------------------------------------------
\begin{equation}
f\left(\mathbf{u}\right) = \left( \mathbf{u}_a, \mathbf{u}_a + t\left(\mathbf{u}_b\right)\right),
\end{equation}
%--------------------------------------------------
where $t$ is modeled as a neural network.
The partitioning is random and defined
during the initialization of the model.
The inverse is easy to compute and the determinant is equal to $1$.

%########################################################
% Subsection : Factor-out layer
%########################################################
%\subsection{Factor-out Layers}
\paragraph{Factor-out layers}
%\textcolor{red}{name clashes, we use $f(x)$ and $f(z)$}
allow a coarse to fine modeling and a simplification of the representation
by only further processing a part of the input features. 
The concept was proposed in the \textit{RealNVP} \citep{realNVP_DinhSB17}.
%In Figure~\ref{fig:overview}, \chris{somehow it feels to early to reference to Fig~\ref{fig:overview} from the method section here?} we can see multiple factor-out layers
%Given a latent representation $U$ where the input features are split into two parts $\mathbf{h}_k$ and $\mathbf{z}_k$,
Formally this is expressed as
%%--------------------------------------------------
%% Equation : Factor-out layer
%%--------------------------------------------------
\begin{equation}
\left(\mathbf{z_1}, \mathbf{h_1} \right) = g_1(\textbf{x}) \qquad \text{and} \qquad \mathbf{z_0} = g_0(\mathbf{h_1}) \;,
\end{equation}
%%--------------------------------------------------
where the input vector $\textbf{x}$ is first mapped to
$\left(\mathbf{z_1}, \mathbf{h_1} \right)$, then only a part of the latents, $\mathbf{h_1}$,
are further processed and mapped to $\mathbf{z_0}$.
The resulting latent representation for $\mathbf{x}$ is 
$\mathbf{z} = (\mathbf{z}_1, \mathbf{z}_0)$.
In addition to computation efficiency, this defines
a conditional dependency between the latents.

\section{Lossy Image Compression with Normalizing Flows}
\label{chap:method}
%------------------------------------------------
% Figure : Overview
%------------------------------------------------
\begin{figure*}
%	\hspace{-0.75cm}
	\centering
	\includegraphics[width=\columnwidth]{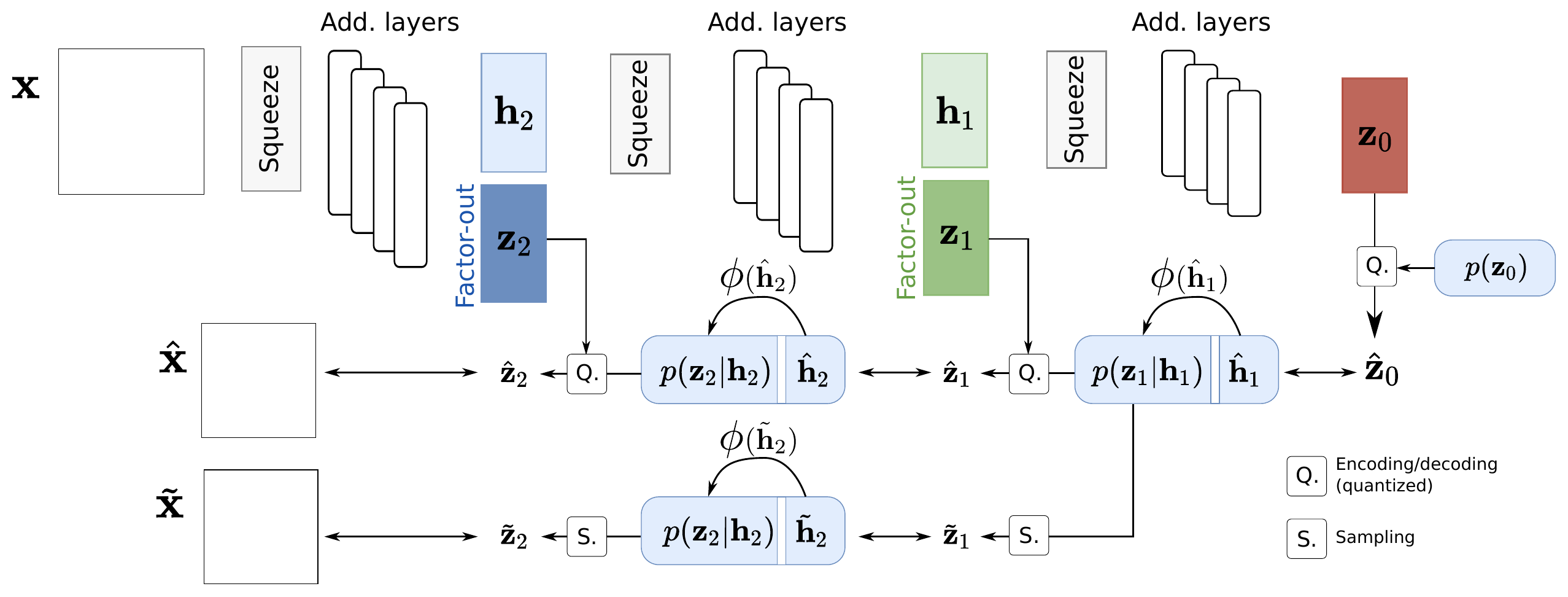}
	\caption{\label{fig:overview}Overview of the flow based image compression architecture. 
		The input image $\textbf{x}$ is processed through a $3$ level network, where each 
		level consists of a squeeze operation followed by a series of additive 
		layers before a factor-out. 
		The input image $\textbf{x}$ is mapped to its latent representation 
		$(\textbf{z}_0,\textbf{z}_1,\textbf{z}_2)$. 
		For compression, these latents are quantized and entropy coded according to
		the learned distributions $p(\mathbf{z}_2 | \mathbf{h}_2)$,
		$p\left(\mathbf{z}_1 | \mathbf{h}_1\right)$ and $p(\mathbf{z}_0)$. }
%		When using (transmitting) only the deepest level $\textbf{z}_0$,
%		the other latents $\textbf{z}_2$ and $\textbf{z}_1$
%		are sampled according to the learned distribution. 
%		Both paths, respectively decoding back to  $\hat{\textbf{x}}$ and $\tilde{\textbf{x}}$,
%		are used during training. \chris{describe both paths verbally? or remove from caption and only in text?}}
\end{figure*}
%------------------------------------------------

\label{sec:methods}
We propose using the normalizing flows for lossy image compression,
and in particular we design an architecture based on the bijective
layers described in the previous section. 
The key properties are the coarse-to-fine representation obtained with
the multi-level factor-out layers and the usage of series of simple additive couplings.
The model is illustrated in Figure~\ref{fig:overview}
%it is possible to train a generative model given a training dataset
%as described in Equation~\ref{eq:glow}.
and consists of $3$ different levels bijectively mapping any 
input image $\mathbf{x}$ to its latent representation 
$\mathbf{z} = (\mathbf{z}_0, \mathbf{z}_1, \mathbf{z}_2)$.
We express lossy image compression as the quantization
and the entropy coding of this latent representation.

%Each level starting
%with a \emph{squeeze} operation
%%that transforms spatial dimension into depth,
%followed by a set of coupling layers. Intermediate levels end with
%a factor-out layer.
We can obtain an estimate of the data distribution by optimizing 
Equation~\ref{eq:glow}. Adapted to our model this equation becomes
%\leo{Simone: she said, she would expect the data distribution. 
%	But its the objective to get an estimation of the data distribution}
%--------------------------------------------------
% Equation : Generative model for our model
%--------------------------------------------------
\begin{align}
\mathcal{L}_{nll}
%	& \simeq \frac{1}{N} \sum_{i=1}^{N} - \log p_{\theta}\left(\mathbf{x}^{(i)} + u\right)
%	 = - \frac{1}{N} \sum_{i=1}^{N}
%	\log\bigg(p\left(\mathbf{z}_0\right)
%				p\left(\mathbf{z}_1 | \mathbf{z}_0\right)
%				p\left(\mathbf{z}_2 | \mathbf{z}_1\right)\bigg)\\
	 %& \simeq \frac{1}{N} \sum_{i=1}^{N} - \log p_{\theta}\left(\tilde{\mathbf{x}}^{(i)}\right) 
	  = - \frac{1}{N} \sum_{i=1}^{N}
	 \log p_\theta\left(\mathbf{z}_0\right) +
	 \log p_\theta\left(\mathbf{z}_1 | \mathbf{z}_0\right) +
	 \log p_\theta\left(\mathbf{z}_2 | \mathbf{z}_1\right)
\label{lbl:equ_wo_quant}
\end{align}
%--------------------------------------------------
%with $\tilde{x}^{(i)} = x^{(i)} + a $.
%Training this generative model we obtain an estimate of data distribution,
%through estimating the distributions $p\left(z_0\right)$,
%$p\left(z_1\,|\,z_0\right)$ and $p\left(z_2\,| z_1\, \right)$.
%as well as a generative process from which it is possible to draw new samples\textcolor{red}{(in general yes, in our case no)}.
%In the case of image compression, one can consider encoding
%a quantized version of the latents $z_0$, $z_1$ and $z_2$.
with all the log-determinant terms equal to $0$ given that we only use
additive layers. %not appear as inverse is easy to compute and the determinant is equal to $1$.
%It is possible to further develop this equation to account for the factor-out layers:
Although the distributions $p_\theta\left(\mathbf{z}_0\right)$,
$p_\theta\left(\mathbf{z}_1 | \mathbf{z}_0\right)$ and
$p_\theta\left(\mathbf{z}_2 | \mathbf{z}_1\right)$
are computed, this model is however not adapted to image compression.
% (see figure \ref{fig:curves_logpx_mse}),
%despite estimating the distributions $p\left(z_0\right)$, $p\left(z_1\,|\,z_0\right)$ and $p\left(z_2\,| z_1\, \right)$.
The reason for this is that the quantized latent values are not observed
during the training as part of the dataset.
%and hence are less likely under the learned probability models.
This leads to our proposed image compression objective.
Considering the decompressed image $\hat{\mathbf{x}}$ obtained
from the rounded latents $\hat{\mathbf{z}}_l = \textit{r}\left(\mathbf{z}_l\right)$
%--------------------------------------------------
% Equation : Decompressed image from rounded latent
%--------------------------------------------------
%\begin{equation}
%\label{eq:eq_gen_quantized}
%\begin{split}
%&\hat{x} = f_\theta(\left[\hat{z}_0, \hat{z}_1, \hat{z}_2\right]) \\
%&\text{with} \quad \hat{z}_0 = \operatorname{round}_{\Delta}\left(z_0\right) ~ , ~
%\hat{z}_1 = \operatorname{round}_{\Delta}\left(z_1\right) ~ , ~
%\hat{z}_2 = \operatorname{round}_{\Delta}\left(z_2\right) \\
%\end{split}
%\end{equation}
\begin{equation}
\label{eq:eq_gen_quantized}
\hat{\mathbf{x}} = f_\theta\left(\hat{\mathbf{z}}_0, \hat{\mathbf{z}}_1, \hat{\mathbf{z}}_2\right), \;\;\; %\hat{\mathbf{z}}_l = \textit{r}\left(\mathbf{z}_l\right)
\end{equation}
%--------------------------------------------------
the objective of image compression is to both minimize 
the entropy and the deviation from the input data point.
This is expressed in the rate-distortion loss ($rdl$)
%--------------------------------------------------
% Equation : Generic loss for Image compression
%--------------------------------------------------
\begin{equation}
\label{eq:rate_dist_optim}
\mathcal{L}_{rdl} = - \log p_\theta \left(\hat{\mathbf{z}}\right) + \lambda~d\left(\mathbf{x}, \hat{\mathbf{x}}\right)
\end{equation}
with $d\left(\mathbf{x}, \hat{\mathbf{x}}\right)$ a term penalizing the deviation 
of $\hat{\mathbf{x}}$ from the original image $\mathbf{x}$. 

In order to further reduce the bit-rate, we need to exploit the hierarchical
architecture and the properties of the factor out distributions of our model.
This is achieved by only transmitting a part of the latent space 
and sampling the missing values.
This is illustrated in Figure~\ref{fig:overview}, where on the sampling path,
only latent values $\mathbf{z}_0$ are entropy coded while 
$\mathbf{z}_1$ and $\mathbf{z}_2$ 
are respectively sampled as the most likely values in the distributions
$p(\mathbf{z}_1 | \mathbf{z}_0)$ and $p(\mathbf{z}_2 | \mathbf{z}_1)$.
This new sampling path is easily included in our model through 
%To force the model to store as much as possible in the prior distribution, we add 
an additional reconstruction loss on the decoded sample 
$\mathbf{\tilde{x}} = f_\theta(\hat{\mathbf{z}}_0, \tilde{\mathbf{z}}_1, \tilde{\mathbf{z}}_2)$. 
The final objective is then defined as follows:
%---------------
% Final loss
%---------------
\begin{equation}
\label{eq:final_optim}
\mathcal{L}\left(\theta\right) = - \log p_\theta (\mathbf{\hat{z}}) + \lambda \left(d(\mathbf{x}, \mathbf{\hat{x}}) + d (\mathbf{x}, \mathbf{\tilde{x}})\right)
\end{equation}
%---------------

\paragraph{Quantization.} Since the rounding operation is a step function,
it is not differentiable and an approximation is necessary to allow gradient 
based optimization.
%In this work, we follow the approach proposed by \cite{DBLP:journals/corr/BalleLS16a} and replace this quantization with an i.i.d. uniform noise source $u$, with $u \sim \mathcal{U}(-\frac{\Delta}{2}, \frac{\Delta}{2})$ and bin size $\Delta$.
In this work, we employ universal quantization proposed
by~\cite{DBLP:conf/iccv/ChoiEL19} to relax this problem.
%Instead of having i.i.d. uniform noise for every element of $z$
%(\cite{DBLP:journals/corr/BalleLS16a})
Universal quantization shifts every element of the latents
by a common random sample
$u \sim \mathcal{U}\left(-\frac{\Delta}{2}, \frac{\Delta}{2}\right)$:
%-----------------------------------------
% Equation : Quantization
%-----------------------------------------
\begin{equation}
\hat{\mathbf{z}} = \operatorname{round}\left(\mathbf{z} + \mathbf{u}\right) - \mathbf{u}, \;\;\; \mathbf{u} = \left[u, u, \dots, u\right].
\end{equation}
%-----------------------------------------
This allows the model to learn correlation of the elements in the latent space. 
For the $\texttt{round}\left(.\right)$ operation, we use the straight-through 
estimator~\citep{DBLP:journals/corr/BengioLC13} and set the gradients to the identity,
$\nabla_\mathbf{x}\texttt{round}\left(\mathbf{x}\right) = \mathbf{I}$.

\paragraph{Probability models.}
To model the base distribution $p_\theta\left(\mathbf{z}_0\right)$ we use the
fully factorized distribution by~\cite{balle2018variational}.
%where a flexible distribution is learned 
For each channel dimension $c$, a neural network is used to learn 
the cumulative density function $F_c$. %As a result,
Considering probability independence between the latent elements $\hat{z}^{j,c}_0$,
we have
%---------------
% Final loss
%---------------
\begin{equation}
\label{eq:ff_model}
p_\theta\left(\hat{\mathbf{z}}_0\right) = 
	\prod_{c} \prod_{j} \left(F_c(\hat{z}^{j,c}_0+\frac{\Delta}{2}) 
	- F_c(\hat{z} ^{j,c}_0-\frac{\Delta}{2}) \right).
\end{equation}
%---------------

% and further the possibility to encode full images. However, this introduces the trade off of losing the ability to sample from the prior distribution.
%The factor out distributions $p(z_1\,|\,z_0)$ and $p(z_2\,|\,z_1)$ are conditioned discretized logistic distributions. %\cite{PixelCNN?}

% For each position in the base distribution we learn a factorized discrete logistic distribution.

For the factor-out layers, we use a conditional distribution.
Let $\hat{z}^{j}$ be an element in the latent encoding $\hat{\mathbf{z}}_l$.
%You can think of $j$ as an iterator over all elements in the tensor $\mathbf{z}_l$. 
%and let $\mu_j$ and $\sigma_j$ be the parameters of the distribution at position $j$, computed by the two parameterized neural networks $\mathtt{NN}_\mu(\mathbf{h}_l)$ and $\mathtt{NN}_\sigma(\mathbf{h}_l)$. 
We model its probability with a discrete logistic distribution 
$\operatorname{DLogistic}\left(\hat{z}_j\,|\, \mu_j, \sigma_j\right)$ defined as:
\begin{align}
\operatorname{DLogistic}\left(\hat{z}_j\,|\,\mu_j, \sigma_j \right) &= \int_{\hat{z}_j -\frac{\Delta}{2}}^{\hat{z}_j + \frac{\Delta}{2}} \operatorname{Logistic}\left(z'\,|\, \mu_j, \sigma_j\right)dz'.
\end{align}
%&\vspace{5pt}\hspace{-15pt} 
%&=  \operatorname{sig}\left(\frac{\hat{z}_j + \frac{\Delta}{2} - \mu_j}{\sigma_j}\right) - \operatorname{sig}\left(\frac{\hat{z}_j - \frac{\Delta}{2} - \mu_j}{\sigma_j}\right)
%with $\operatorname{sig}(\cdot)$ being the sigmoid function. 
Parameters $\mu_j$ and $\sigma_j$ at every position $j$ are computed by 
a parameterized neural network $\phi(\mathbf{h}_l)$. 
Given this probability model, the sampling path amounts to using 
$\tilde{z}_j = \mu_j$.
%\chris{wouldn't it be a more coherent notation to use a single letter (g) or whatever is still free?}
%$[\mathbf{\mu}_l, \mathbf{\sigma}_l] = \mathtt{NN}(\mathbf{h}_l)$.
%\textcolor{red}{Should we reference the integer flow paper here?}

\paragraph{Compression at different quality levels.}
In order to avoid retraining image compression models for different
rate-distortion levels \cite{DBLP:conf/iccv/ChoiEL19} 
use the Lagrange multiplier and the quantization bin size
as conditioning variables to the network. %as rate control parameters, which
It is unclear how such a strategy would fit in the normalizing flows model.
Instead, to get different quality levels on the rate-distortion curve, 
we train a single model using a fixed $\lambda$ value (see experiments for details),
and at test time, we reach different rate-distortion points by varying 
the quantization step size used in the latent representation.
In our case however, multiple levels are present and hence 
multiple step sizes must be set. 
To achieve best performance, we propose to fine-tune the quantization step values
by minimizing the rate-distortion loss for various values of $\lambda$:
\begin{equation}
L\left(\mathbf{x}; \mathbf{\Delta}\right) = - \operatorname{log}p_\theta\left(\hat{\mathbf{z}}\right) 
	+ \lambda~d\left(\mathbf{x}, \hat{\mathbf{x}}\right) \;,
\end{equation}
where $\mathbf{\Delta}$ is the set of quantization steps to be estimated:
%the pre-trained model and optimize the quantization steps. 
a single scalar for each one of $\hat{\mathbf{z}}_2$ and $\hat{\mathbf{z}}_1$, 
and a distinct value for each channel of $\hat{\mathbf{z}}_0$.

%We further observed % the quantization step leads 
%unnecessary transmission of information %in many cases.
%when quality increases. \chris{is it possible / does it make sense to explain this more. I.e. how we observe it and why this is the case (  or was that due to the probability model we currently use and maybe we want to try to not say it? ) } In this case, as the prediction %become more reliable % distributions become 
To further improve the bit-rate as the predictions %
reach higher probability values around the means (with respect to the quantization step),
it becomes more beneficial to simply sample this value. 
% sending to less possible but more probable symbols.
% Hence, if 
%as the quantized bin around the mean value becomes sufficiently large 
%and 
In our compression scheme, we use a threshold 
on the probability value around the mean,
$\operatorname{DLogistic}\left(\mu_j\,|\,\mu_j, \sigma_j \right) >  P_{\textrm{thresh}}$,
for which no information is transferred and the mean $\mu_j$ is used ($z_j = \mu_j$). 
In all our experiments $P_{\textrm{thresh}} = 0.9$.

\section{Related Work}
\label{chap:related}
% \paragraph{Neural Image Compression}
Recent works made significant progress in applying neural networks to image compression. 
\cite{toderici2015variable,toderici2017full} first introduced a general 
framework for variable-rate image compression using an LSTM based architecture.
Advantageously, the method directly outputs a binary representation 
and has no explicit limit on the amount of information that can be encoded,
however their empirical results suggest this solution does not
necessarily lead to rate–distortion optimality.

%Instead of using pre-defined transformations, 
Better performances have been achieved using autoencoders to learn 
in an end-to-end fashion an optimal transformation to a latent 
representation, as well as the probability distribution required 
for entropy coding to the final bitstream~\citep{balle2017endtoend,theis2017lossy,rippel2017real}.
For example~\cite{rippel2017real} use a pyramidal encoder to extract 
features of various scales. After quantizing the encoding, they use adaptive 
arithmetic coding and adaptive code length regularization to reduce 
the length of the bitstream. \cite{balle2017endtoend} propose 
a model that uses newly proposed nonlinear transformations (GDN) 
which implement a form of local gain control. 
%Through the introduced continuous proxy for the discontinuous 
%loss function, they are able to optimize the model end-to-end. 
%The used entropy model is fully factorized. 
To further improve performance, more focus was placed on the entropy 
model of the learned latent representation. In particular,
\cite{balle2018variational} use side information in the form
of a hyperprior to condition the Gaussian mixture model 
used to entropy code the latents. 
%The link with variational autoencoders is discussed in They also compare their model to optimizing a variational .
In a different direction, \cite{mentzer2018conditional} use a 3D-CNN  
to learn a conditional probability model of the latent distribution. 
%This context model learns the dependencies between the symbols in the latent representation.
State of the art results were achieved by combining these 
strategies~\cite{minnen2018full}.
%In general, there are two types of approaches to improve compression algorithms. The first approach to develop powerful encoders and decoders which are able to reconstruct the input images and the second type of approach is to learn a image or dataset dependent entropy model.

%The first works of neural network based image compression methods optimize fixed entropy models on a training set that is shared between the encoder and decoder. 
%The authors in \citep{DBLP:conf/cvpr/TodericiVJHMSC17} propose a set of full-resolution lossy image compression methods. 
%Their models consist of recurrent neural network (RNN)-based encoder and decoder, a binarizer and a neural network for entropy coding. 

%\paragraph{Image Compression}
At this stage we find it interesting to draw back attention to the traditional approaches 
in image compression. These methods use an invertible transformation
before a quantization step where information is explicitly discarded. 
This is the case for JPEG~\citep{DBLP:journals/cacm/Wallace91} where image patches are 
transformed with a discrete cosine transformation (DCT) then the obtained coefficients 
are scaled, quantized and entropy encoded to retrieve the bit stream.
This strategy ensures that it is always possible to increase the quality 
by sending more data. Recent image codecs have improved on some particular aspects, like
JPEG 2000~\citep{taubman2002jpeg2000} which replaces the DCT with a wavelet-based 
method or BPG~\citep{nxp:bpg} and Webp~\citep{nxp:webp} that explored directions 
such as intra prediction and in-loop filtering, but all are capable of leveraging
more data for better quality and even provide a lossless mode.

Some learned lossless compression approaches have been proposed recently:
\cite{mentzer2019practical} use a hierarchical probabilistic model;
\cite{DBLP:journals/corr/abs-1905-07376}
%demonstrated the interest of using normalizing flow for image compression and 
introduced integer coupling layers to learn a discrete mapping 
from the image space to a discrete base distribution
which can be used for lossless image compression.
We note however that given their probability model, the solution
is limited to a fixed size and encoding a large image 
requires patch-wise processing. This work is the first to explore 
using normalizing flows for \emph{lossy} image compression 
and demonstrating compression results on a wide range of bit-rates.

\section{Experiments}
\label{chap:experiments}

The model we consider in our experiments 
%is illustrated in Figure~$\ref{fig:overview}$. It is based on 
%a simpler variant of 
%the multi-scale architecture introduced by~\cite{realNVP_DinhSB17}
%and has $L = 3$ levels. 
uses $8$ additive layers at each level. 
%A step is defined as a random permutation of the channels 
%followed by an additive coupling layer.
%While there are various different types of coupling layers, we chose the additive coupling layer, because its $\operatorname{log}|\operatorname{det}(J)| = 0$. Optimizing $\operatorname{log}p(x)$ is therefore equivalent to optimizing $\operatorname{log}p(z)$.
Each of these coupling layers~(Figure~\ref{fig:coupling}), 
uses half of the channels. 
%to parameterize the transformation network that learns the shift of 
%the remaining of the channels. 
The transformation network $t(\cdot)$ is a ResNet with two blocks. 
During training, we use the AdaMax optimizer~\citep{DBLP:journals/corr/KingmaB14}
with a learning rate of $10^{-3}$ (with $\epsilon = 10^{-7}$) 
and the quantization step is set to $\Delta = 1$.
%\aziz{Does it make sense to give more details in suppl. material? (and mention it here)}\\
In addition to the traditional image compression codecs, JPEG and BPG, 
we compare our model with recent state of the art learning based 
models. %~\citep{balle2018variational}.
%Instead of train multiple models with various penalization of the rate-term, 
In order to properly compare the performance, we train on the same dataset (COCO with $\approx$ 285k images) and refer to the retrained versions of~\citep{balle2018variational} and ~\citep{balle2017endtoend} as \textit{AE + Hyperprior} and \textit{AE + Fully-Factorized} respectively.
%For each of the two models we follow the same training procedure and 
%For all the learning based solutions and with the different training setups, we 
Furthermore, we always train a single model and use different 
quantization steps at test time %~\cite{dumas2018autoencoder} 
to create the rate-distortion curves. 
%
% Differences we know:
% - training data set
% - batch size (8 (balle) vs. 32 (ours)
% - we quantize after training a single model.
%
%-------------------------------------------------------------------------------
% General Plot Style Settings
%-------------------------------------------------------------------------------
\pgfplotsset{compat=1.14}
\pgfplotsset{grid style={color=black!15, },} % dashed
\pgfplotsset{every axis x grid/.style={draw opacity=0}}
%\pgfplotsset{minor grid style={dashed,red}}
%\pgfplotsset{major grid style={dotted,green!50!black}}

\pgfplotsset{
	axis x line=bottom,
	axis y line=left,
	axis line style={draw=none},
	tick style={draw=none},
	grid,
	xmin=0, xmax=1,
	xlabel=Rate (bpp),
	legend style={at={(1.0,0.04)},anchor=south east},
	legend cell align={left},
	label style={font=\footnotesize},
	legend style={font=\tiny},
	legend style={draw=gray},
	tick label style={font=\footnotesize},
	width=0.49\columnwidth
}
\pgfplotsset{
	every axis plot/.append style={line width=1pt, cap=round, smooth} %tension=0.2},
	%	every axis plot post/.append style={
	%		every mark/.append style={line width=1.6pt,draw=green,fill=red}
	%	}
}

%-------------------------------------------------------------------------------
% Some colors
%-------------------------------------------------------------------------------

\definecolor{myBlue}{RGB}{41,77,165}
\definecolor{myBlueLight}{RGB}{164,185,226}
\definecolor{myOrange}{RGB}{231,116,46 }
\definecolor{myOrangeLight}{RGB}{239,177,105}
\definecolor{myGreen}{RGB}{78,145,41}
\definecolor{myGreenFlashy}{RGB}{0,219,0}
\definecolor{myGreenLight}{RGB}{160,219,207}
\definecolor{myRed}{RGB}{181,45,38}
\definecolor{myRedLight}{RGB}{233,138,133}
\definecolor{myBlack}{RGB}{0,0,0}
\definecolor{myViolet}{RGB}{231,41,138}
\definecolor{myVioletLight}{RGB}{100,0,100}
\definecolor{myDarkOrange}{RGB}{117,112,179}
\newcommand{\BPGTWO}{myBlue}
\newcommand{\BPGFOUR}{myRed}

\newcommand{\JPEG}{myOrangeLight}

\newcommand{\OURS}{myGreenFlashy}
\newcommand{\FF}{myViolet}
\newcommand{\HYPER}{myDarkOrange}
\newcommand{\AELIMIT}{myBlack}

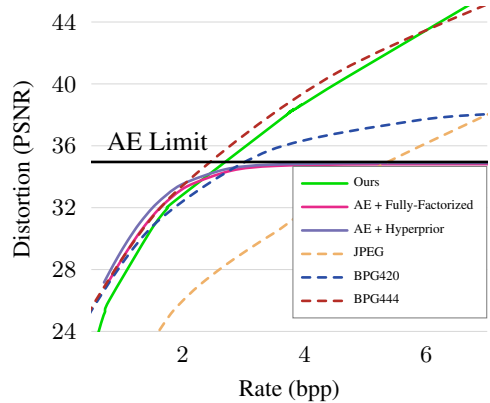
\begin{wrapfigure}{r}{0.45\textwidth}
\vspace{-0.2cm}
\begin{tikzpicture}[scale=1.0]
\begin{axis}[legend style={font=\fontsize{5}{5}\selectfont},
ymin=24, ymax=45, ytick={24,28, ..., 46}, xmin=0.5, xmax=7,ylabel=Distortion (PSNR),]
%		\addplot[draw=\OURS] table [x=bpp, y=psnr, col sep=comma] {data/csvs/imagenet/result_line_imagenet64_logistic_lv_3_fl_8_nc_64_lat_additive_tt_resnet_pt_ff_lot_nfrdl_qt_uninoise__2020-04-07_01_01_48.csv};
\addplot[draw=\OURS] table [x=bpp, y=psnr, col sep=comma] {data/csvs/imagenet/result_line_imagenet64_logistic_lv_3_fl_8_nc_64_lat_additive_tt_resnet_pt_ff_lot_nfrdl_qt_uninoise__2020-04-07_01_01_48_mix.csv};
%\addplot[draw=\OURS] table [x=bpp, y=psnr, col sep=comma] {data/csvs/imagenet/result_line_imagenet64_logistic_lv_3_fl_8_nc_64_lat_additive_tt_resnet_pt_ff_lot_nfrdl_qt_uninoise__2020-04-07_01_01_48_new.csv};
\addlegendentry{Ours}
\addplot[draw=\FF,line width=1.0pt] table [x=bpp, y=psnr_round, col sep=comma] {data/csvs/imagenet/result_ff.csv};
\addlegendentry{AE + Fully-Factorized}	
\addplot[draw=\HYPER,line width=1.0pt]table [x=bpp, y=psnr_round, col sep=comma] {data/csvs/imagenet/result_hp_gsm.csv};
\addlegendentry{AE + Hyperprior}
\addplot[draw=\JPEG, dashed]  table [x=bpp, y=psnr_round, col sep=comma] {data/csvs/imagenet/result_jpeg.csv};
\addlegendentry{JPEG}
\addplot[draw=\BPGTWO, dashed] table [x=bpp, y=psnr_round, col sep=comma] {data/csvs/imagenet/result_bpg420.csv};
\addlegendentry{BPG420}
\addplot[draw=\BPGFOUR, dashed] table [x=bpp, y=psnr_round, col sep=comma] {data/csvs/imagenet/result_bpg444.csv};
\addlegendentry{BPG444}	
\addplot[draw=\AELIMIT,postaction={decorate,decoration={text along path,
		text={~~~~~~ AE Limit}, raise=1ex, text align={left}}}] 
table [x=id, y=psnr_round, mark options={scale=1},smooth, col sep=comma] {data/csvs/imagenet/result_no.csv};
\end{axis}
\end{tikzpicture}
\caption{\label{fig:patch-results} Results on ImageNet64. %fixed resolution $64\times64$. 
%Both training and testing data have the same resolution, which is the regular setup used 
%for training and testing normalizing flow models. 
Our model is not limited 
on the reconstruction quality contrary to auto-encoder models, which have a hard 
limit indicated by the \emph{AE Limit} line.}
\vspace{-1.5cm}
\end{wrapfigure}
For completeness, we add the numbers published in the original paper where available. 
In addition to this, and to better understand the limits on autoencoder based models, 
we also train a version of the networks to achieve the highest reconstruction without any rate penalty.
This is done by dropping the rate penalty term during training.
Models trained this way are referred to as \textit{AE Limit}. 
%	While the psnr values in the lower bit-rates 
%	are comparable to the numbers the authors published, a deviation in higher bit-rates can be noticed.
%	The reason for this could be either a higher number of iterations
%	with smaller batch size \leo{(8 (balle) vs. 32 (ours)}
%	or the larger dataset ($\approx$ 1 million) which leads
%	to a better generalization of the auto encoder.
%	\leo{Note, while this would also lead to a higher \textit{AE Limit},
%		the best reconstruction an auto encoder without entropy model can achieve,
%		the reconstruction has due to the dimension reduction a hard limit.}	
%	We added for completeness, also the numbers published in \cite{balle2018variational}.

%and  Additionally, we compare our model with an learned based compression network 
%trained without rate-term, which could be seen as the best possible reconstruction the auto encoder can achieve.
%The learning based methods are implemented as proposed in \cite{balle2017endtoend} and \cite{balle2018variational}. 
%The plot shows, that our model outperforms the traditional image compression codecs. It can be also seen, that our model covers a wider range of reconstruction qualities and performs better on higher bit rates than previous state of the art models.
%Next we detail the different datasets and how there were used. 
%The following datasets: Sprites~\aziz{ref?}, the downscaled 
%ImageNet ILSVRC 2012 dataset~\citep{ILSVRC15}, Tencik~\aziz{ref?} and Kodak~\aziz{ref?}
%are used in this work. More details are provided along corresponding experiment.

\subsection{Image Compression - Fixed Resolution}
When using normalizing flows related work both in image synthesis
\citep{glow_KingmaD18,realNVP_DinhSB17} and recently in image 
compression~\citep{DBLP:journals/corr/abs-1905-07376} only considered fixed size
images. In the same spirit, and to demonstrate the potential of normalizing flows,
our first experiment is based on a fixed size of $64\times64$.
%We used the \emph{Sprites} and the \emph{ImageNet64} datasets which we first introduce before discussing rate-distortion performance.
%
%\paragraph{Sprites.} To create this dataset we used the publicly available sprite 
%sheets\footnote{\url{https://github.com/jrconway3/Universal-LPC-spritesheet}}. 
%We chose 4 attributes (skin color, shirt, legs and hair-color) to define a unique identity.
%For each of this attributes we selected 6 different appearances which 
%makes in total $6^4 = 1296$ different combinations of identities. 
%The background of each image is a random crop of an high resolution image. 
%The size of a single image is $64 \times 64$ pixels.
%Our model is trained for $40$ epochs. 
%To create the validation and test sets we randomly chose 4 identities
%not seen during training. 
%\paragraph{ImageNet64.} We resized the images from ImageNet 
For this, we resized the images from ImageNet 
to a resolution of $64\times64$ as described in~\citep{DBLP:journals/corr/ChrabaszczLH17}. 
In this case, the models is trained for $25$ epochs, where each epoch consists 
of $40$k batches of size $32$.

%Trained is done for \aziz{X epochs?} with a learning rate of \aziz{LR?}.
Image compression results are presented in Figure~\ref{fig:patch-results}. 
%Here the learned compression models~\citep{balle2017endtoend,balle2018variational}
%are trained on the same data as our model. 
The autoencoder based models have 
a hard limit on the image reconstruction quality contrary to our solution
that consistently increases in quality as more information is transferred. 
This limit is shown with a horizontal line and the
value is obtained by training autoencoder networks 
to achieve the highest reconstruction without any rate penalty. 

In the low bit-rate part of curve, auto-encoder based solutions
are the best performing. This is due to the more 
powerful transformations that are in place, in particular the use
of nonlinear transformations (GDN layers). For the moment we only
use additive coupling layers in our network and there is potential
for substantial improvements as better bijective layers are developed
for normalizing flows. 
%\paragraph{Patch Camelyon.} The Patch Camelyon dataset (\cite{Veeling2018-qh}) consists of 328k images (96x96) extracted from histology scans. We reduced the size of the test set to 512 by randomly sampling from the test images. (dataset with a high frequency details)

%-------------------------------------------------------------------------------
% General Plot Style Settings
%-------------------------------------------------------------------------------
\pgfplotsset{compat=1.14}
\pgfplotsset{grid style={color=black!15, },} % dashed
\pgfplotsset{every axis x grid/.style={draw opacity=0}}
%\pgfplotsset{minor grid style={dashed,red}}
%\pgfplotsset{major grid style={dotted,green!50!black}}

\pgfplotsset{
	axis x line=bottom,
	axis y line=left,
	axis line style={draw=none},
	tick style={draw=none},
	grid,
	xmin=0, xmax=1,
	xlabel=Rate (bpp),
	legend style={at={(1.0,0.04)},anchor=south east},
	legend cell align={left},
	label style={font=\footnotesize},
	legend style={font=\tiny},
	legend style={draw=gray},
	tick label style={font=\footnotesize},
	width=0.52\columnwidth
}
\pgfplotsset{
	every axis plot/.append style={line width=1pt, cap=round, smooth} %tension=0.2},
	%	every axis plot post/.append style={
	%		every mark/.append style={line width=1.6pt,draw=green,fill=red}
	%	}
}

%-------------------------------------------------------------------------------
% Some colors
%-------------------------------------------------------------------------------

\definecolor{myBlue}{RGB}{41,77,165}
\definecolor{myBlueLight}{RGB}{164,185,226}
\definecolor{myOrange}{RGB}{231,116,46 }
\definecolor{myOrangeLight}{RGB}{239,177,105}
\definecolor{myGreen}{RGB}{78,145,41}
\definecolor{myGreenFlashy}{RGB}{0,219,0}
\definecolor{myGreenLight}{RGB}{160,219,207}
\definecolor{myRed}{RGB}{181,45,38}
\definecolor{myRedLight}{RGB}{233,138,133}
\definecolor{myBlack}{RGB}{0,0,0}
\definecolor{myViolet}{RGB}{231,41,138}
\definecolor{myVioletLight}{RGB}{100,0,100}
\definecolor{myDarkOrange}{RGB}{117,112,179}
\definecolor{myGray}{RGB}{100,100,100}

\newcommand{\PAPERFF}{myViolet}
\newcommand{\PAPERHYPER}{myDarkOrange}
\newcommand{\PAPERTHEIS}{myGray}
\newcommand{\PAPERTODICI}{myBlack}

\begin{figure*}
	\hspace{-0.4cm}
	\subfloat{
		\begin{tikzpicture}[scale=1.0]
		\begin{axis}[title=Tecnick Dataset,
		ymin=32, ymax=50, ytick={32,36, ..., 48}, xmin=0, xmax=6, 
		ylabel=Distortion (PSNR)]
%		\addplot[draw=\OURS] table [x=bpp, y=psnr_round, col sep=comma] {data/csvs/tecnick/result_line_mscoco128_logistic_lv_3_fl_8_nc_128_lat_additive_tt_resnet3_pt_ff_lot_nfrdl_qt_uninoise__2020-04-04_13_42_35.csv};
		\addplot[draw=\OURS] table [x=bpp, y=psnr_round, col sep=comma] {data/csvs/tecnick/result_line_mscoco128_logistic_lv_3_fl_8_nc_128_lat_additive_tt_resnet3_pt_ff_lot_nfquantrdl_qt_uninoise__2020-05-11_18_33_52_mix.csv};
		%\addlegendentry{Ours}
		\addplot[draw=\JPEG, dashed] table [x=bpp, y=psnr_round, col sep=comma] {data/csvs/tecnick/result_jpeg.csv};
		%\addlegendentry{JPEG}
		\addplot[draw=\BPGTWO, dashed] table [x=bpp, y=psnr_round, col sep=comma] {data/csvs/tecnick/result_bpg420.csv};
		%\addlegendentry{BPG420}
		\addplot[draw=\BPGFOUR, dashed] table [x=bpp, y=psnr_round, col sep=comma] {data/csvs/tecnick/result_bpg444.csv};
		%\addlegendentry{BPG444}
		%\addplot[draw=\FF,line width=1.0pt] table [x=bpp, y=psnr_round, col sep=comma] {data/csvs/tecnick/result_ff.csv};
		%\addlegendentry{\cite{balle2017endtoend}}	
		\addplot[draw=\HYPER,line width=1.0pt] table [x=bpp, y=psnr_round, col sep=comma] {data/csvs/tecnick/result_hp_gsm.csv};
		
			%papers
		\addplot[draw=\PAPERFF,dashed, line width=0.8pt] table [x=bpp, y=psnr, col sep=comma] {data/manual/tecnick/balle_ff.csv};
		\addplot[draw=\PAPERHYPER,dashed, line width=0.8pt] table [x=bpp, y=psnr, col sep=comma] {data/manual/tecnick/balle_hp.csv};

		%\addlegendentry{\cite{balle2018variational}}
		\addplot[draw=\AELIMIT,postaction={decorate,decoration={text along path,
				text={ AE Limit}, raise=1ex, text align={left}}}] 
			table [x=id, y=psnr_round, col sep=comma] {data/csvs/tecnick/result_no.csv};
		\end{axis}
		\end{tikzpicture}
	}
	\subfloat{
		\hspace{0.5cm}
		\begin{tikzpicture}[scale=1.0]
		\begin{axis}[title=Kodak Dataset,
		legend style={font=\fontsize{5}{5}\selectfont},
		legend pos = south east, 
		ymin=32, ymax=50, ytick={32,36, ..., 48}, xmin=0, xmax=6]
%		\addplot[draw=\OURS] table [x=bpp, y=psnr_round, col sep=comma] {data/csvs/kodak/result_line_mscoco128_logistic_lv_3_fl_8_nc_128_lat_additive_tt_resnet3_pt_ff_lot_nfrdl_qt_uninoise__2020-04-04_13_42_35.csv};
		\addplot[draw=\OURS] table [x=bpp, y=psnr_round, col sep=comma] {data/csvs/kodak/result_line_mscoco128_logistic_lv_3_fl_8_nc_128_lat_additive_tt_resnet3_pt_ff_lot_nfquantrdl_qt_uninoise__2020-05-11_18_33_52_mix.csv};
		\addlegendentry{Ours}
		%\addplot[draw=\FF,line width=1.0pt] table [x=bpp, y=psnr_round, col sep=comma] {data/csvs/kodak/result_ff.csv};
		%\addlegendentry{\cite{balle2017endtoend}}	
		\addplot[draw=\HYPER,line width=1.0pt] table [x=bpp, y=psnr_round, col sep=comma] {data/csvs/kodak/result_hp_gsm.csv};
		\addlegendentry{AE + Hyperprior}

			%papers
		\addplot[draw=\PAPERFF,dashed, line width=0.8pt] table [x=bpp, y=psnr, col sep=comma] {data/manual/kodak/balle_ff.csv};
		\addlegendentry{\cite{balle2017endtoend}}	
		\addplot[draw=\PAPERHYPER,dashed, line width=0.8pt] table [x=bpp, y=psnr, col sep=comma] {data/manual/kodak/balle_hp.csv};
		\addlegendentry{\cite{balle2018variational}}	
%		\addplot[draw=\PAPERTHEIS,dashed, line width=0.8pt] table [x=bpp, y=psnr, col sep=comma] {data/manual/kodak/theis.csv};
%		\addlegendentry{\cite{theis2017lossy}}	
		\addplot[draw=\PAPERTODICI,dashed, line width=0.8pt] table [x=bpp, y=psnr, col sep=comma] {data/manual/kodak/toderici.csv};
		\addlegendentry{\cite{toderici2017full}}

		\addplot[draw=\JPEG, dashed] table [x=bpp, y=psnr_round, col sep=comma] {data/csvs/kodak/result_jpeg.csv};
		\addlegendentry{JPEG}
		\addplot[draw=\BPGTWO, dashed] table [x=bpp, y=psnr_round, col sep=comma] {data/csvs/kodak/result_bpg420.csv};
		\addlegendentry{BPG420}
		\addplot[draw=\BPGFOUR, dashed] table [x=bpp, y=psnr_round, col sep=comma] {data/csvs/kodak/result_bpg444.csv};
		\addlegendentry{BPG444}
		\addplot[draw=\AELIMIT,postaction={decorate,decoration={text along path,
				text={AE Limit}, raise=1ex, text align={left}}}] 
		table [x=id, y=psnr_round, col sep=comma] {data/csvs/tecnick/result_no.csv};

		%\addplot+[mark options={scale=1, opacity=0.6}] table [x=bpp, y=psnr_round, only marks, scatter, mark options={fill opacity=0.5}, col sep=comma] {data/csvs/kodak/result_mscoco128_logistic_lv_3_fl_8_nc_128_lat_additive_tt_resnet3_pt_ff_lot_nfrdl_qt_uninoise__2020-04-04_13_42_35.csv};
		\end{axis}
		\end{tikzpicture}
	}
	\caption{\label{fig:evaluation-results} Image compression results on high 
		resolution images. Our model is not limited on the reconstruction quality 
		contrary to auto-encoder models as indicated by the \emph{AE Limit} line.
		%\chris{Balle 2018 seems to be consistently better than 2017 now (not sure if we were always getting this on the sprites) but since this is very much as expected, could we just only plot Balle 2018?}
	}
\end{figure*}
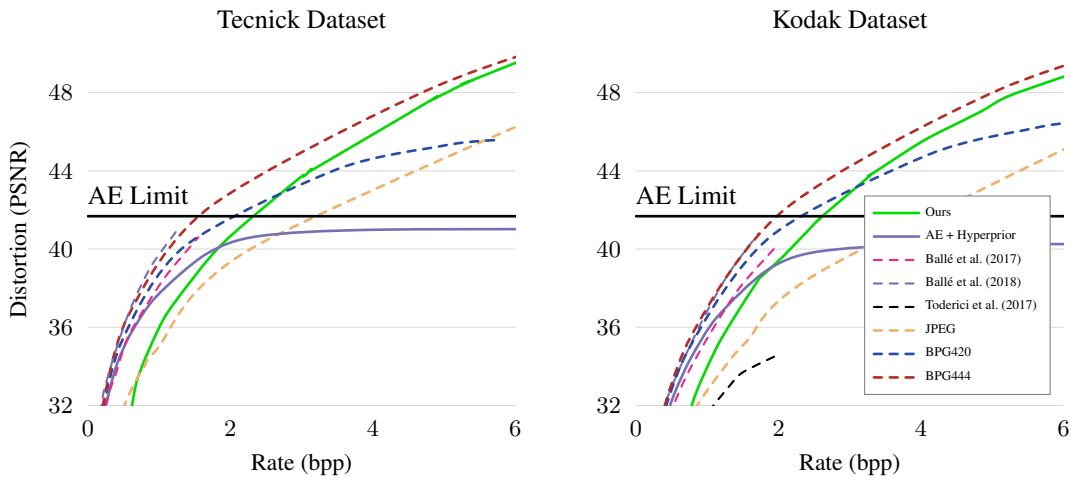

\subsection{Image Compression - Arbitrary High Resolution}

The proposed architecture is fully convolutional and the probability distribution of the latents
corresponding to the coarse level is modeled globally (see Equation~\ref{eq:ff_model}).
As a result, the model can be trained on a fixed resolution and applied to arbitrary 
sized images. We use the COCO 
dataset~\citep{DBLP:journals/corr/LinMBHPRDZ14} for training and extract 
image patches of size $128\times128$. 
We evaluate the trained models on the full images of the Kodak \footnote{Downloaded from \url{http://r0k.us/graphics/kodak/}}
and the Tecnick~\citep{DBLP:journals/jgtools/AsuniG13} data sets. 
%In this more complex \leo{(Simone: complex in terms of image content 
%not because of larger size)} setting, we use a ResNet with 3 blocks and 128 hidden channels. 
The results are visualized in Figure~\ref{fig:evaluation-results}.

For completeness, we also show the PSNR values available from the author \cite{balle2018variational} for the Kodak dataset. Here only a smaller range of bitrates is covered as multiple individual models were trained. Furthermore, the training was relying on a significantly larger dataset. As the PSNR values are higher than our retrained \textit{AE + Hyperprior} model, this may indicate that further gains for our flow based model might also be possible when training on a larger dataset.

In addition to this, the clear benefit of our normalizing flow based model is the capacity to target
a large range of compression rates from low bit-rate to high quality results. This allows to better match the behavior of traditional codecs and thus largely closes an important gap between learning based and traditional approaches.

\subsection{Quasi Lossless Reencoding}
Besides the capacity to handle a larger range of quality levels
than existing learned image compression techniques, 
our proposed method offers additional advantages.
%By using normalizing flows, our model learns bijective a mapping 
%from image space to latent space. 
%a distribution over images and their compressed version.
%The compressed images are therefore part of the learned distribution.
%This allow us, to get a bijective mapping from the compressed image to its latent encoding.
%-------------------------------------------------------------------------------
% General Plot Style Settings
%-------------------------------------------------------------------------------
\pgfplotsset{compat=1.14}
\pgfplotsset{grid style={color=black!15, },} % dashed
\pgfplotsset{every axis x grid/.style={draw opacity=0}}
%\pgfplotsset{minor grid style={dashed,red}}
%\pgfplotsset{major grid style={dotted,green!50!black}}

\pgfplotsset{
	axis x line=bottom,
	axis y line=left,
	axis line style={draw=none},
	%tick style={draw=none},
	grid,
	xmin=0, xmax=1,
	xlabel=Rate (bpp),
	legend style={at={(0.72,.04)},anchor=south east},
	legend cell align={left},
	label style={font=\footnotesize},
	legend style={font=\footnotesize},
	legend style={draw=gray},
	tick label style={font=\footnotesize},
	width=0.32\columnwidth,
	height=0.35\textwidth
}
\pgfplotsset{
	every axis plot/.append style={line width=1pt, cap=round, smooth} %tension=0.2},
	%	every axis plot post/.append style={
	%		every mark/.append style={line width=1.6pt,draw=green,fill=red}
	%	}
}

%-------------------------------------------------------------------------------
% Some colors
%-------------------------------------------------------------------------------

\definecolor{myOrange}{RGB}{217,95,2}
\definecolor{myGreen}{RGB}{27,158,119}

\renewcommand{\OURS}{myGreen}
\renewcommand{\FF}{myOrange}

\begin{wrapfigure}{r}{0.42\textwidth}
\centering
\vspace{-0.cm}
	\begin{tikzpicture}[scale=1.0]
	\definecolor{acolorbrewer1}{RGB}{27,158,119}
	\definecolor{acolorbrewer2}{RGB}{217,95,2}
	\pgfplotsset{scale only axis, xmin = 1, xmax = 17}
	%%------------------------------
	% PSNR values
	%------------------------------
	\begin{axis}[%title=Multi-compression,
	ymin=28, ymax=40, ytick={28,30, ..., 40}, 
	axis y line*=left,
	xlabel = Iterations,
	ylabel=Distortion (PSNR),
	mark options={scale=1.0}]
	\addplot+[mark options={scale=0.5}, acolorbrewer1] table [x=id, y=psnr_round, col sep=comma] {data/csvs/multicompression/kodak/qflow/result_mc_qflow1_kodim19_qsz0_None_z1_None_z2_1.45.png.csv};
	\addlegendentry{Ours}
	\addplot+[mark=*, mark options={scale=0.5}, acolorbrewer2] table [x=id, y=psnr_round, col sep=comma] {data/csvs/multicompression/kodak/ae/result_mc_ae_kodim19_qs1.0.png.csv};
	%\addlegendentry{\cite{balle2017endtoend}}
	\addlegendentry{AE + Hyperprior}
	\end{axis}
	\end{tikzpicture}
	\caption{\label{fig:multicompression-plots}Successive re-encodings}
	\vspace{-0.25cm}
\end{wrapfigure}
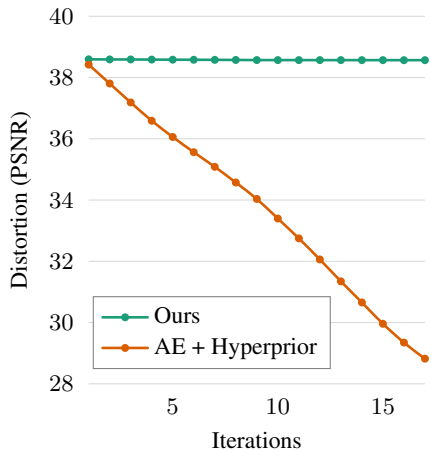
Since we use normalizing flows, there is a one-to-one mapping between quantized latent values 
and the corresponding reconstructed image. This means that compressing 
the decoded image again will result in the same latent values. Once the image quality is fixed,
chaining multiple compression and decompression steps does not change the quality 
of the image and retains rate-distortion performance. This is in contrast to existing autoencoder based solutions that can quickly deteriorate 
and produce visible artifacts as content is reencoded several times (see Figure~\ref{fig:teaser}).
Being able to properly reencode can be important in video production pipelines where 
%retaining a high quality is required, and 
the image and video content might be composited and edited by different parties requiring reencodings in the process.
An evaluation is shown in Figure~\ref{fig:multicompression-plots},
where both our model and the autoencoder based approach of~\cite{balle2018variational}
start from the same image quality. Both approaches achieve a bit-rate of $1.7$bpp
for all the steps, but the autoencoder reconstruction quality solution 
drops by almost $1$db after only $1$ step and up to $10$db after $15$ steps.

%------------------------------------------------
% Figure : Low to high 
%------------------------------------------------
\begin{figure*}
	\hspace{-0.5cm}
	\includegraphics[width=1.05\textwidth]{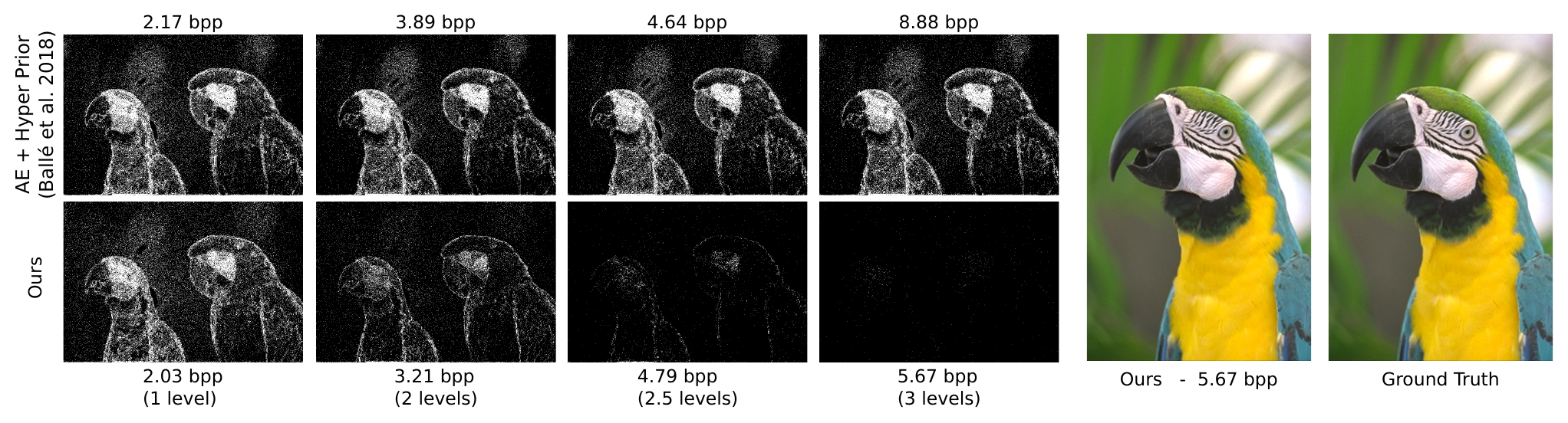} 
	\caption{\label{fig:low-to-high} From lower bit-rate values to near lossless quality.
	On the left side we highlight every pixel with an error on the color
	value. As levels are encoded and the bit-rate increases, our solution reaches
	near lossless quality levels. Visual comparison with the ground truth is shown on the right side.}
\end{figure*}
%------------------------------------------------

%[1]\url{https://en.wikipedia.org/wiki/JPEG_2000#Progressive_transmission_by_pixel_and_resolution_accuracy}
%[2]\url{https://de.wikipedia.org/wiki/JPEG#Progressives_JPEG}
%[3]\url{https://en.wikipedia.org/wiki/Scalable_Video_Coding}
\subsection{Towards Scalable Coding and Progressive Transmission}
Scalable video coding (SVC)~\citep{DBLP:journals/tcsv/SchwarzMW07} allows to transmit only parts of a bit stream with the goal of retaining rate-distortion performance w.r.t. the partial stream that could represent a lower spatial resolution, a lower temporal resolution, or a lower fidelity content. 
As such SVC provides interesting opportunities for graceful degradation and flexible bit-rate adaptation in the context of video streaming. However, due to added complexity, it has not found widespread industry adoption. 
Aside from SVC, image standards such as JPEG and JPEG2000 also allow progressive transmission. 
This can be seen as a more fine grained way of scaling image fidelity. 

Our solution uses a hierarchical model with factor out distributions and thus is designed to inherently allow us to store details 
in the top layers and coarse information in the bottom layers.
However, contrary to \citep{DBLP:journals/corr/abs-1905-07376}
these intermediate layers are optimized to have best rate-distortion performance.
When only using the information stored in the bottom layer $\hat{\mathbf{z}}_0$, we can automatically set the values on the remaining layers to their estimated means without requiring to transmit any further information. In this way, we obtain a decoded image that has competitive rate-distortion performance w.r.t this partial stream of data. I.e. the results are 
comparable with those produced by autoencoders in terms of rate \emph{and} distortion~(see Figure.~\ref{fig:low-to-high}, $1$~level).
%The latents $\hat{z}_0$ can be therefore seen as the bottleneck of the flow.
To incrementally scale to a higher quality image, it is sufficient to only send the remaining 
latents $\hat{\mathbf{z}}_1$ (Figure.~\ref{fig:low-to-high}, $2$~levels),
%a fraction of $\hat{\mathbf{z}}_2$ (e.g. \textit{50\%} in Figure.~\ref{fig:low-to-high}, $2.5$~levels)
or the full encodings $\hat{\mathbf{z}}_2$ (Figure.~\ref{fig:low-to-high}, $3$~levels). 
It is also possible to partially send data and achieve 
progressive scaling, following a predefined pattern.
This is illustrated for the latent $\hat{\mathbf{z}}_2$ (in Figure.~\ref{fig:low-to-high}, $2.5$~levels)
where only a fraction (\textit{50\%}) of the values are sent. 

\section{Conclusion}
\label{chap:conclusion}
In this paper, we explored an approach for lossy image compression based on normalizing flows. To the best of our knowledge, we are the first to evaluate the potential of normalizing flows in the task of lossy compression. Furthermore, we have showcased a number beneficial properties inherent to our solution: 
1) We are able to cover the widest range of quality levels with a single trained model ranging from low bitrates to almost lossless quality thereby bridging an important gap that had remained between currently used autoencoder based methods and traditional transform codecs.
2) We show that we can reencode images multiple times in a quasi lossless way closely retaining rate distortion performance. This has been problematic with autoencoder based methods.
3) We describe different ways of achieving scalable coding or progressive reconstruction with our model.

These properties make normalizing flows a very exciting candidate for 
lossy image compression models even though we do not yet outperform 
existing autoencoder based approaches over the full range of bit rates. 
We believe the results obtained in this work are encouraging
as there is still a lot of potential for improvement 
and would push new research works to explore new bijective layers,
architecture designs and refined training procedures.

\bibliography{imcodec_nf}
\bibliographystyle{icml2020}

\newpage
\appendix
\section{Supplementary Material: Lossy Image Compression with Normalizing Flows}

In this section we describe the datasets and the details of the model architecture. 
While some hyperparameters differ between datasets, the common architecture is the same for all experiments. 
We further provide a pseudo code for the entropy coding used in our compression method.

\begin{figure}[h]
	\centering
	\includegraphics[width=0.8\textwidth]{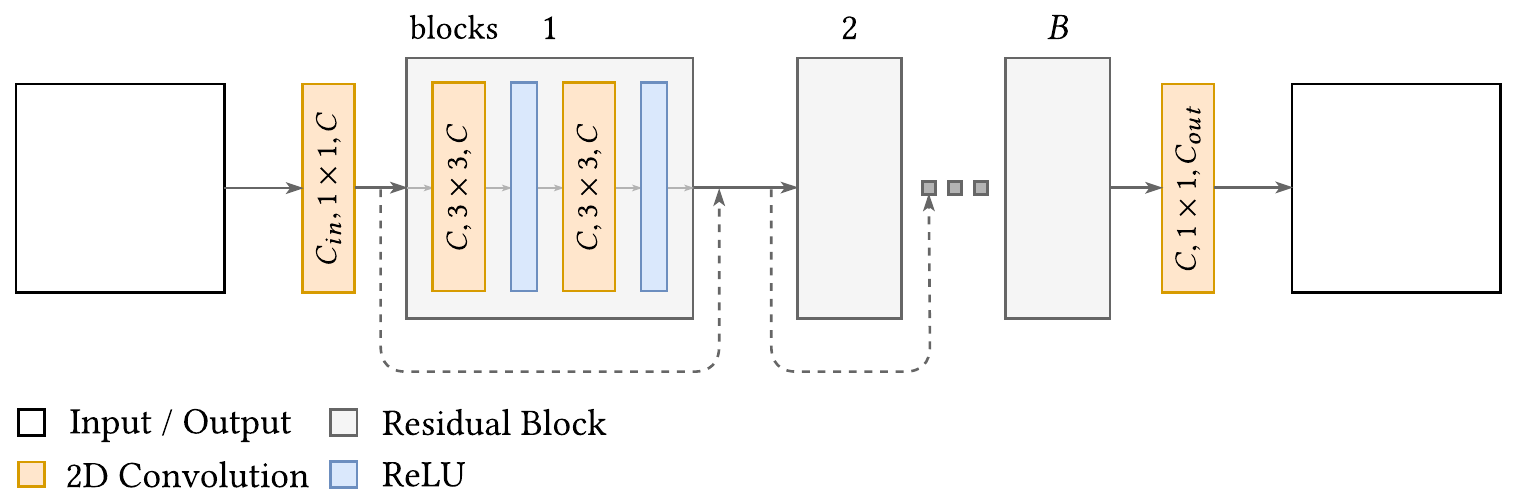}
	\caption{Overview of the architecture used as $t\left(.\right)$ and $\phi\left(.\right)$}
	\label{fig:resnet}
\end{figure}

We describe the model illustrated in Figure 3 in the main paper.
Independent of the dataset, the models have $L = 3$ levels with $K$ steps each. 
Each levels starts with a \textit{Squeeze}-layer. 
A single step consists of a random permutation and 
an \textit{additive transformation}. 
A ResNet (Figure~\ref{fig:resnet}) with $B$ 
blocks and $C$ hidden channels is used as transformer network $t\left(.\right)$ 
(see also Figure 2 in the main paper).

Each factor-out layer divides the latents into two equal sized partitions, 
%where one part is considered as finished and the other is passed to the next level. 
and only one of them is passed to the next level. 
We use a fully factorized model as prior $p\left(\mathbf{z}_0\right)$ 
and discrete logistic distributions as conditional distributions
$p\left(\mathbf{z}_1 | \mathbf{z}_0\right)$ and $p\left(\mathbf{z}_2 | \mathbf{z}_1\right)$. 
The network $\phi\left(.\right)$ used to predict the parameters of the 
conditional distributions is also a ResNet with $B$ blocks and $C$ channels.

During training, we use the AdaMax optimizer~\citep{DBLP:journals/corr/KingmaB14}
with a learning rate of $10^{-3}$ (with $\epsilon = 10^{-7}$) 
and the quantization step is set to $\Delta = 1$.

After training, to optimize the quantization steps $\mathbf{\Delta}$ 
for a fixed $\lambda \in \left[1, 10^6\right]$ we use Adam optimizer
~\citep{DBLP:journals/corr/KingmaB14} 
with learning rate $10^{-1}$. 
%Further, a threshold of $P_{\textrm{thresh}} = 0.9$ was used.

%We compared our models with state-of-the-art 
%autoencoder based models~\citep{balle2017endtoend, balle2018variational}. 

\subsection{ImageNet64}
We resized the images from ImageNet 
to a resolution of $64\times64$ as described in~\citep{DBLP:journals/corr/ChrabaszczLH17}. 
For this dataset, the model is trained for $25$ epochs, 
where each epoch consists of $40$k batches of size $32$.

\begin{figure}[h]
	\centering
	\begin{tabular}{l r}
		\midrule
		\textbf{Parameter} & \textbf{Value} \\ 
		\midrule
		Levels $L$ & 3 \\
		Steps per level $K$ & 8 \\
		$\phi\left(.\right)$ & ResNet ($B = 2$, $C = 64$) \\
		$t\left(.\right)$ & ResNet ($B = 2$, $C = 64$) \\
		$\lambda$ & 500 \\
		batch size & 32 \\
		learning rate & $10^{-3}$ \\
		\midrule
	\end{tabular}
	\caption{Model parameters used for ImageNet64 dataset}
\end{figure}

\paragraph{Baselines} We used the same architecture proposed 
in~\citep{balle2018variational} with $M = 192$ and $N = 192$. 
We optimized for each model (\textit{AE + Fully-Factorized} 
and \textit{AE + Hyperprior}) the rate-distortion-loss for 
a fixed $\lambda = 0.3$ with Adam optimizer %~\citep{DBLP:journals/corr/KingmaB14} 
and learning rate $lr = 2\cdot10^{-4}$ for $\sim 5.5$~million iterations with batch size $32$.

\subsection{COCO}
The COCO dataset consists of $\approx 285k$ images. 
We used the original train/validation/test - split. 
During training, we randomly cropped patches of size $128\times128$ due to the limitation of memory. 
The model is trained for $50$ epochs. 

\begin{figure}[h]
	\centering
	\begin{tabular}{l r}
		\midrule
		\textbf{Parameter} & \textbf{Value} \\ 
		\midrule
		Levels $L$ & 3 \\
		Steps per level $K$ & 8 \\
		$\phi\left(.\right)$ & ResNet ($B = 3$, $C = 128$) \\
		$t\left(.\right)$ & ResNet ($B = 3$, $C = 128$) \\
		$\lambda$ & 500 \\
		batch size & 8 \\
		learning rate & $10^{-3}$ \\
		\midrule
	\end{tabular}
	\caption{Model parameters used for COCO dataset}
\end{figure}

\paragraph{Baselines} We used the same architecture proposed in~\citep{balle2018variational} 
with $M = 192$ and $N = 192$. We optimized for each model (\textit{AE + Fully-Factorized} 
and \textit{AE + Hyperprior}) the rate-distortion-loss 
for a fixed $\lambda = 0.15$ with Adam optimizer %~\citep{DBLP:journals/corr/KingmaB14} 
and learning rate $lr = 2\cdot10^{-4}$ for $\sim 1.2$ Million iterations 
with batch size $32$. During training, we randomly cropped patched of size $256\times256$.

\subsection{Entropy Coding: Pseudo Algorithm}

The pseudo code in Algorithm 1 
%\ref{alg:compression_algorithm} 
describes how our thresholding is considered in entropy encoding and decoding. The thresholding is performed based on the probability of the predictions (see end of Chapter 3 in the main paper). As the threshold can be arbitrarily set in principle, it offers another degree of freedom when realizing progressive transmission. 

\label{sec:pseudo_algo}

\begin{algorithm}[th]
	\caption{Pseudo code for entropy coding the latents}
	
	%\algsetup{\tiny}
	\scriptsize
		%\SetAlgoNoLine
	\begin{multicols}{2}
	 \SetKwFunction{FEntropyEncode}{entropy\_encode}
	  \SetKwFunction{FEntropyDecode}{entropy\_decode}
	 	\DontPrintSemicolon
 
 	\SetKwProg{Fn}{Function}{:}{}
	 \Fn{\FEntropyEncode{\textit{Latents} $[\hat{\mathbf{z}}^{*}_0, \hat{\mathbf{z}}^{*}_1, \hat{\mathbf{z}}^{*}_2]$}}{
 	
 	$\mathbf{b}$ = $\texttt{encode}\left(\hat{\mathbf{z}}^{*}_0, P[\hat{\mathbf{z}}^{*}_0]\right)$
 	
 	\For{$l = 1 \rightarrow 2$}{
 		\tcc{for each position $j$ in the latent of level $l$}
 		\ForEach{$z_j \in \hat{\mathbf{z}}_l^{*}$}{
 			%\tcc{retrieve the mean of the learned conditional distribution}
 			$\mu_j$ = $\texttt{mean}\left(P[z_j\;|\;\hat{\mathbf{z}}^{*}_{l-1}, \dots, \hat{\mathbf{z}}^{*}_{0}]\right)$
 			
 			%\tcc{if the probability of the mean is below a threshold, encode the latent}
 			
 			\If{$P[\mu_j\:|\:\hat{\mathbf{z}}^{*}_{l-1}, \dots, \hat{\mathbf{z}}^{*}_{0}] \leq  P_{\textrm{thresh}}$}{$\mathbf{b}$ += $\texttt{encode}\left(z_j,P[\hat{\mathbf{z}}^{*}_j\:|\:\hat{\mathbf{z}}^{*}_{l-1}, \dots, \hat{\mathbf{z}}^{*}_{0}]\right)$}
 		}	
 	}
 	
 	\KwRet $\mathbf{b}$
 }
	    \columnbreak
	
	\SetKwProg{Fn}{Function}{:}{}
		\Fn{\FEntropyDecode{\textit{bitstream} $\mathbf{b}$}}{
		%# compute the probability of the prior distribution
		
		$\hat{\mathbf{z}}^{*}_0$ = $\texttt{decode}\left(\mathbf{b}, P[\hat{\mathbf{z}}^{*}_0]\right)$
		
		\For{$l = 1 \rightarrow 2$}{
			\tcc{for each position $j$ in the latent of level $l$}
			\ForEach{$z_j \in \hat{\mathbf{z}}_l^{*}$}{
				%\tcc{retrieve the mean of the learned conditional distribution}
				$\mu_j$ = $\texttt{mean}\left(P[z_j\;|\;\hat{\mathbf{z}}^{*}_{l-1}, \dots, \hat{\mathbf{z}}^{*}_{0}]\right)$
				
				%\tcc{if the probability of the mean is below a threshold, encode the latent}
				
				\If{$P[\mu_j\:|\:\hat{\mathbf{z}}^{*}_{l-1}, \dots, \hat{\mathbf{z}}^{*}_{0}] \leq  P_{\textrm{thresh}}$}{
					$z^{*}_j$ = $\texttt{decode}\left(\mathbf{b},P[z_j\:|\:\hat{\mathbf{z}}^{*}_{l-1}, \dots, \hat{\mathbf{z}}^{*}_{0}]\right)$}
				\Else{
					$z^{*}_j$ = $\mu_j$ 
				}
			}	
		}
		
    	\KwRet $[\hat{\mathbf{z}}^{*}_0, \hat{\mathbf{z}}^{*}_1, \hat{\mathbf{z}}^{*}_2]$
	}

	\label{alg:compression_algorithm} %https://www.overleaf.com/project/5d82453d84607c00014e48cb
\end{multicols}
\end{algorithm}

\end{document}